%% file: robust-saa.tex
\ifdefined\usemorstyle

\documentclass[moor]{informs1}              

\usepackage{natbib}
 \NatBibNumeric
 \bibpunct[, ]{[}{]}{,}{n}{}{,}%

\usepackage{statistics-macros-mor}
\usepackage[top=2.54cm,bottom=2.54cm,left=2.54cm,right=2.54cm]{geometry}
\usepackage{color}
\usepackage{bbm}
\usepackage{pgfplotstable}
\usepackage{subcaption}
\usepackage{epstopdf}
\newcommand{\red}[1]{\textcolor{red}{#1}}

\definecolor{innerboxcolor}{rgb}{.9,.95,1}
\definecolor{outerlinecolor}{rgb}{.6,0,.2}

\newcommand{\hncomment}[1]{\fcolorbox{outerlinecolor}{innerboxcolor}{
    \begin{minipage}{.9\textwidth}
      \red{\bf HN Comment:} {#1}
  \end{minipage}} \\
}

\newcommand{\jcdcomment}[1]{\fcolorbox{outerlinecolor}{innerboxcolor}{
    \begin{minipage}{.9\textwidth}
      \red{\bf JCD Comment:} {#1}
  \end{minipage}} \\
}

\usepackage[colorlinks=true,breaklinks=true,bookmarks=true,urlcolor=blue,
     citecolor=blue,linkcolor=blue,bookmarksopen=false,draft=false]{hyperref}

\def\EMAIL#1{\href{mailto:#1}{#1}}
\def\URL#1{\href{#1}{#1}}         

\TheoremsNumberedThrough     

\EquationsNumberedThrough    

\begin{document}


\RUNAUTHOR{Duchi, Glynn, and Namkoong}

\RUNTITLE{Statistics of Robust Optimization: A Generalized
  Empirical Likelihood Approach}

\TITLE{Statistics of Robust Optimization:
    \\ A Generalized Empirical Likelihood Approach}

\ARTICLEAUTHORS{%
  \AUTHOR{John C. Duchi} \AFF{Department of Electrical Engineering and
    Statistics, Stanford University, \EMAIL{jduchi@stanford.edu},
    \URL{}}

  \AUTHOR{Peter W. Glynn, Hongseok Namkoong} \AFF{Department of
    Management Science and Engineering, Stanford University,
    \EMAIL{glynn, hnamk@stanford.edu}, \URL{}}
} 

\ABSTRACT{%
  \input{abstract}
}%


\MSCCLASS{}
\ORMSCLASS{Primary: ; secondary: }
\HISTORY{}

\maketitle
%

\else

\documentclass[11pt, letterpaper]{article}
\usepackage{statistics-macros}
\usepackage{xcolor}

\definecolor{darkblue}{rgb}{0,0, .5}

\usepackage[colorlinks=true,linkcolor=darkblue,citecolor=darkblue,urlcolor=blue]{hyperref}
\usepackage[numbers,square]{natbib}
\usepackage[top=2.54cm,bottom=2.54cm,left=2.54cm,right=2.54cm]{geometry}
\usepackage{bbm}
\usepackage{pgfplotstable}
\usepackage{graphicx}
\usepackage{subcaption}
\usepackage{epstopdf}
\epstopdfsetup{outdir=./Simulations/}

\newcommand{\red}[1]{\textcolor{red}{#1}}

\definecolor{innerboxcolor}{rgb}{.9,.95,1}
\definecolor{outerlinecolor}{rgb}{.6,0,.2}

\newcommand{\hncomment}[1]{\fcolorbox{outerlinecolor}{innerboxcolor}{
    \begin{minipage}{.9\textwidth}
      \red{\bf HN Comment:} {#1}
  \end{minipage}} \\
}

\newcommand{\jcdcomment}[1]{\fcolorbox{outerlinecolor}{innerboxcolor}{
    \begin{minipage}{.9\textwidth}
      \red{\bf JCD Comment:} {#1}
  \end{minipage}} \\
}

\newcommand{\pwgcomment}[1]{\fcolorbox{outerlinecolor}{innerboxcolor}{
    \begin{minipage}{.9\textwidth}
      \red{\bf PWG Comment:} {#1}
  \end{minipage}} \\
}

\renewcommand{\theassumption}{\Alph{assumption}}




\author{John C. Duchi, Peter W. Glynn, Hongseok Namkoong}

\begin{document}


\vspace{1cm}
\begin{center}
  {\huge {Statistics of Robust Optimization:
    \\ A Generalized Empirical Likelihood Approach}} \\
\vspace{.5cm} {\large John C.\ Duchi$^{1}$
  ~~~~ Peter W.\ Glynn$^{2}$
  ~~~~ Hongseok Namkoong$^2$} \\
\vspace{.2cm} {\large
  $^1$Departments of Electrical Engineering and Statistics \\
  \vspace{.1cm}   $^2$Department of Management Science and Engineering
  \\ \vspace{.1cm} Stanford
  University }
\vspace{.2cm}
{\tt \{jduchi,glynn,hnamk\}@stanford.edu} \\
\vspace{.2cm} {Stanford
  University }
\end{center}

\begin{abstract}
  \input{abstract}
\end{abstract}

\fi

\newcommand{\zlow}{Z_\star}
\newcommand{\littleprob}{q_\star}
\newcommand{\fconp}{{f^*}'}
\newcommand{\image}{\mathop{\rm Im}}
\newcommand{\taylrem}[2]{\mathsf{r}_{#1,#2}}

\input{introduction}

\input{generalized-el}

\input{inference-stoch-opt}

\input{connections-ro}

\input{consistency}

\input{simulations}

\input{generalization}

\input{conclusion}

\bibliographystyle{abbrvnat}
\bibliography{bib}

\ifdefined\usemorstyle
\begin{APPENDICES}
\input{pointwise-expansion-proof-short}
\input{uniform-expansion-proof}
\input{danskin-proof}
\input{ergodic}
\input{consistency-proof}

\end{APPENDICES}

\else

\newpage 
\appendix

\input{pointwise-expansion-proof-short}
\input{uniform-expansion-proof}
\input{danskin-proof}
\input{ergodic}
\input{consistency-proof}
\fi

\end{document}

%% file: abstract.tex

We study statistical inference and distributionally robust solution methods
for stochastic optimization problems, focusing on confidence intervals for
optimal values and solutions that achieve exact coverage asymptotically. We
develop a generalized empirical likelihood framework---based on distributional
uncertainty sets constructed from nonparametric $f$-divergence balls---for
Hadamard differentiable functionals, and in particular, stochastic
optimization problems.  As consequences of this theory, we provide a
principled method for choosing the size of distributional uncertainty regions
to provide one- and two-sided confidence intervals that achieve exact
coverage. We also give an asymptotic expansion for our distributionally robust
formulation, showing how robustification regularizes problems by their
variance. Finally, we show that optimizers of the distributionally robust
formulations we study enjoy (essentially) the same consistency properties as
those in classical sample average approximations.  Our general approach
applies to quickly mixing stationary sequences, including geometrically
ergodic Harris recurrent Markov chains.

%% file: introduction.tex

\section{Introduction}

We study statistical properties of distributionally robust solution methods
for the stochastic optimization problem
\begin{equation}
  \label{eqn:pop}
  \minimize_{x\in\mathcal{X}} ~ \E_{P_0}[\obj]
  = \int_{\statdomain} \obj dP_0(\statval).
\end{equation}
In the formulation~\eqref{eqn:pop}, the feasible region
$\mathcal{X} \subset \R^d$ is a nonempty closed set, $\statval$ is a random
vector on the probability space $(\statdomain, \sigalg, P_0)$, where the
domain $\statdomain$ is a (subset of) a separable metric space, and the
function $\loss : \mathcal{X} \times \statdomain \to \R$ is a lower
semi-continuous (loss) function. In most data-based decision making scenarios,
the underlying distribution $P_0$ is unknown, and even in scenarios, such as
simulation optimization, where $P_0$ is known, the integral $\E_{P_0}[\obj]$
may be high-dimensional and intractable to compute. Consequently, one
typically~\cite{ShapiroDeRu09} approximates the population
objective~\eqref{eqn:pop} using the sample average approximation (SAA) based on
a sample $\statrv_1, \ldots, \statrv_n \simiid P_0$
\begin{equation}
  \label{eqn:saa}
  \minimize_{x \in \mathcal{X}} ~ \E_{\emp}[\obj]
  = \frac{1}{n} \sum_{i=1}^n \obji,
\end{equation}
where $\emp$ denotes the usual empirical measure over the sample
$\{\statval_i\}_{i = 1}^n$.

We study approaches to constructing confidence intervals for
problem~\eqref{eqn:pop} and demonstrating consistency of its approximate
solutions. We develop a family of convex optimization programs, based on the
distributionally robust optimization framework~\cite{DelageYe10,
  Ben-TalGhNe09, BertsimasGuKa18, Ben-TalHeWaMeRe13}, which allow us to
provide confidence intervals with asymptotically exact coverage for optimal
values of the problem~\eqref{eqn:pop}. Our approach further yields approximate
solutions $\what{x}_n$ that achieve an asymptotically guaranteed level of
performance as measured by the population objective
$\E_{P_0}[\loss(x; \statrv)]$. More concretely, define the optimal value
functional $T_{\rm opt}$ that acts on probability distributions on $\statdomain$ by
\begin{equation*}
  T_{\rm opt}(P) \defeq \inf_{x \in \mathcal{X}}~ \E_P[\obj].
\end{equation*}
For a fixed confidence level $\alpha$, we show
how to construct an interval $[l_n, u_n]$ based on the sample
$\statrv_1, \ldots, \statrv_n$  with (asymptotically) \emph{exact
  coverage}
\begin{equation}
  \label{eqn:asymptotic-convergence}
  \lim_{n \to \infty} \P\left(T_{\rm opt}(P_0) \in [l_n, u_n]\right) = 1 - \alpha.
\end{equation}
This exact coverage
indicates the interval $[l_n, u_n]$ has correct size as the sample
size $n$ tends to infinity. We also give sharper statements than the
asymptotic guarantee~\eqref{eqn:asymptotic-convergence}, providing expansions
for $l_n$ and $u_n$ and giving rates at which $u_n - l_n \to 0$.

Before summarizing our main contributions, we describe our approach and
discuss related methods. We begin by recalling divergence measures for
probability distributions~\cite{AliSi66,Csiszar67}. For a lower
semi-continuous convex function $f : \R_+ \to \R \cup \{+\infty\}$
satisfying $f(1) = 0$, the \emph{$f$-divergence} between probability
distributions $P$ and $Q$ on $\statdomain$ is
\begin{equation*}
  \fdiv{P}{Q} = \int f\left(\frac{dP}{dQ}\right) dQ
  = \int_\statdomain f\left(\frac{p(\statval)}{q(\statval)}\right)
  q(\statval) d\mu(\statval),
\end{equation*}
where $\mu$ is a $\sigma$-finite measure with $P, Q \ll \mu$, and
$p \defeq dP / d\mu$ and $q \defeq dQ / d\mu$. With this
definition, we will show that for $f \in \mathcal{C}^3$ near
$1$ with $f''(1) = 2$, the upper and lower confidence bounds
\begin{subequations}
  \label{eqn:upper-lower-def}
  \begin{align}
    u_n & \defeq \inf_{x \in \mathcal{X}}
    \sup_{P \ll \emp}\left\{ \E_{P}[\loss(x; \statrv)]
    : \fdivs{P}{\emp} \leq \frac{\tol}{n} \right\} \label{eqn:upper} \\
    l_n & \defeq \inf_{x \in \mathcal{X}}
    \inf_{P \ll \emp}\left\{ \E_{P}[\loss(x; \statrv)]
    : \fdivs{P}{\emp} \leq \frac{\tol}{n} \right\} \label{eqn:lower}
  \end{align}
\end{subequations}
yield asymptotically exact coverage~\eqref{eqn:asymptotic-convergence}.
In the formulation~\eqref{eqn:upper-lower-def}, the parameter
$\rho = \chi^2_{1, 1 - \alpha}$ is chosen as the $(1 - \alpha)$-quantile of
the $\chi_1^2$ distribution.


The upper endpoint~\eqref{eqn:upper} is a natural distributionally robust
formulation for the sample average approximation~\eqref{eqn:saa}, proposed by
\citet{Ben-TalHeWaMeRe13} for distributions $P$ with finite support.  The
approach in the current paper applies to arbitrary distributions, and we are
therefore able to explicitly link (typically dichotomous~\cite{Ben-TalGhNe09})
robust optimization formulations with stochastic optimization. We show how a
robust optimization approach for dealing with parameter uncertainy yields
solutions with a number of desirable statistical properties, even in
situations with dependent sequences $\{\statval_i\}$. The exact coverage
guarantees~\eqref{eqn:asymptotic-convergence} give a
principled method for choosing the size $\tol$ of distributional uncertainty
regions to provide one- and two-sided confidence intervals.

We now summarize our contributions, unifying the approach to uncertainty based
on robust optimization with classical statistical goals.
\begin{enumerate}[(i)]
\item \label{item:contribution-el} We develop an empirical likelihood
  framework for general smooth functionals $T(P)$, applying it in particular
  to the optimization functional $T_{\rm opt}(P) = \inf_{x\in\mathcal{X}}
  \E_{P}[\obj]$. We show how the
  construction~\eqref{eqn:upper}--\eqref{eqn:lower} of $[l_n, u_n]$ gives a
  confidence interval with exact coverage~\eqref{eqn:asymptotic-convergence}
  for $T_{\rm opt}(P_0)$ when the minimizer of $\E_{P_0}[\obj]$ is
  unique. To do so, we extend Owen's empirical likelihood
  theory~\cite{Owen88, Owen90} to suitably smooth (Hadamard differentiable)
  nonparametric functionals $T(P)$ with general $f$-divergence measures (the
  most general that we know in the literature); our proof is different from
  Owen's classical result even when $T(P) = \E_P[X]$ and extends
  to stationary sequences $\{\statval_i\}$.
\item We show that the upper confidence set $\openleft{-\infty}{u_n}$ is a
  one-sided confidence interval with exact coverage when
  $\tol = \chi^2_{1, 1-2\alpha} = \inf\{\tol' : \P(Z^2 \le \tol') \ge 1 -
  2\alpha, Z \sim \normal(0, 1)\}$.  That is, under suitable conditions on
  $\loss$ and $P_0$,
  \begin{equation*}
    \lim_{n \to \infty}
    \P\left(\inf_{x\in\mathcal{X}} \E_{P_0}[\obj] \in (-\infty, u_n]\right)
    = 1-\alpha.
  \end{equation*}
  This shows that the robust optimization problem~\eqref{eqn:upper}, which is
  efficiently computable when $\loss$ is convex, provides an upper confidence
  bound for the optimal population objective~\eqref{eqn:pop}.
\item \label{item:contribution-expansion} We show that the robust
  formulation~\eqref{eqn:upper} has the (almost sure) asymptotic
  expansion
  \begin{align}
    \label{eqn:variance-expansion-intro}
    \sup_{P \ll \emp} \left\{ \E_P[\obj]: \fdivs{P}{\emp} \le
    \frac{\tol}{n}\right\}
    = \E_{\emp}[\obj] + (1 + o(1)) \sqrt{\frac{\tol}{n} \var_P (\obj)},
  \end{align}
  and that this expansion is uniform in $x$ under mild restrictions.
  Viewing the second term in the expansion as a regularizer for the SAA
  problem~\eqref{eqn:saa} makes concrete the intuition that robust
  optimization provides regularization; the regularizer accounts for the
  variance of the objective function (which is generally non-convex in $x$ even
  if $\loss$ is convex), reducing uncertainty. We give weak conditions
  under which the expansion is uniform in $x$, showing that the
  regularization interpretation is valid when we choose $\what{x}_n$ to
  minimize the robust formulation~\eqref{eqn:upper}.
\item Lastly, we prove consistency of estimators $\what{x}_n$ attaining the
  infimum in the problem~\eqref{eqn:upper} under essentially the same
  conditions for consistency of SAA (see
  Assumption~\ref{assumption:moment}). More precisely, for the sets of optima
  defined by
  \begin{align*}
    S\opt \defeq \argmin_{x \in \xdomain} \E_{P_0}[\obj]
    ~~\mbox{and} ~~
    S_n\opt \defeq \argmin_{x \in \mathcal{X}}
    \sup_{P \ll \emp}\left\{ \E_{P}[\loss(x; \statrv)]:
    \fdivs{P}{\emp} \leq \frac{\tol}{n} \right\},
  \end{align*}
  the distance from any point in $S_n\opt$ to $S\opt$
  tends to zero so long as $\loss$ has more than one moment under $P_0$
  and is lower semi-continuous.
\end{enumerate}

\subsection*{Background and prior work}

The nonparametric inference framework for stochastic optimization we
develop in this paper is the empirical likelihood counterpart of the normality
theory that Shapiro develops~\cite{Shapiro89, Shapiro91}. While
an extensive literature exists on statistical inference for stochastic
optimization problems (see, for example, the line of work~\cite{DupacovaWe88,
  Shapiro89, King89, Shapiro90, KingWe91, Shapiro91, KingRo93, Shapiro93,
  ShapiroDeRu09}), Owen's empirical likelihood framework~\cite{Owen01} has
received little attention in the stochastic optimization literature save for
notable recent exceptions \cite{LamZh17, Lam18}. In its classical form,
empirical likelihood provides a confidence set for a $d$-dimensional mean
$\E_{P_0}[Y]$ (with a full-rank covariance) by using the set
$C_{\tol, n} \defeq \{\E_P[Y] : \fdivs{P}{\emp} \le \frac{\tol}{n} \}$ where
$f(t) = -2\log t$.
Empirical likelihood theory shows that if we set
$\tol = \chi_{d,1 - \alpha}^2 \defeq \inf\left\{\tol' : \P(\ltwo{Z}^2 \le
  \tol') \ge 1 - \alpha ~~ \mbox{for} ~ Z \sim \normal(0, I_{d \times
    d})\right\}$, then $C_{\tol, n}$ is an asymptotically exact
$(1-\alpha)$-confidence region, i.e.
$\P(\E_{P_0}[Y] \in C_{\tol,n}) \to 1 - \alpha$. Through a self-normalization
property, empirical likelihood requires no knowledge or estimation of unknown
quantities, such as variance. We show such
asymptotically pivotal results also apply for the robust optimization
formulation~\eqref{eqn:upper-lower-def}. The empirical likelihood confidence
interval $[l_n, u_n]$ has the desirable characteristic that when
$\loss(x; \statval) \ge 0$, then $l_n \ge 0$ (and similarly for $u_n$), which
is not necessarily true for confidence intervals based on the normal
distribution.

Using confidence sets to robustify optimization problems involving randomness
is common (see \citet[Chapter 2]{Ben-TalGhNe09}).  A number of researchers
extend such techniques to situations in which one observes a sample
$\statval_1, \ldots, \statval_n$ and constructs an uncertainty set over the
data directly, including the papers~\cite{DelageYe10, WangGlYe15,
  Ben-TalHeWaMeRe13, BertsimasGuKa18, BertsimasGuKa14}.  The duality of
confidence regions and hypothesis tests~\cite{LehmannRo05} gives a natural
connection between robust optimization, uncertainty sets, and statistical
tests. \citet{DelageYe10} made initial progress in this direction by
constructing confidence regions based on mean and covariance matrices from the
data, and \citet{JiangGu13} expand this line of research to other moment
constraints.  \citet*{BertsimasGuKa18, BertsimasGuKa14} develop uncertainty
sets based on various linear and higher-order moment conditions. They also
propose a robust SAA formulation based on goodness of fit tests, showing
tractability as well as some consistency results based on Scarsini's linear
convex orderings~\cite{Scarsini99} so long as the underlying distribution is
bounded; they further give confidence regions that do not have exact coverage.
The formulation~\eqref{eqn:upper-lower-def} has similar motivation to the
preceding works, as the uncertainty set
\begin{equation*}
  \left\{\E_{P}[\obj]: \fdivs{P}{\emp} \le \frac{\tol}{n} \right\}
\end{equation*}
is a confidence region for $\E_{P_0}[\obj]$ for each fixed $x\in\mathcal{X}$
(as we show in the sequel).  Our results extend this
by showing that, under mild conditions, the values~\eqref{eqn:upper}
and~\eqref{eqn:lower} provide upper and lower confidence bounds
for $T(P) = \inf_{x \in \mathcal{X}} \E_{P}[\obj]$ with (asymptotically) exact
coverage.

\citet{Ben-TalHeWaMeRe13} explore a similar scenario to ours, focusing on the
robust formulation~\eqref{eqn:upper}, and they show that when $P_0$ is
finitely supported, the robust program~\eqref{eqn:upper} gives a one-sided
confidence interval with (asymptotically) inexact coverage (that is, they only
give a bound on the coverage probability). In the unconstrained setting
$\fr = \R^d$, \citet{LamZh17} used estimating equations to show that standard
empirical likelihood theory gives confidence bounds for stochastic
optimization problems. Their confidence bounds have asymptotically inexact
confidence regions, although they do not require unique solutions of the
optimization problem as our results sometimes do.  The
result~\eqref{item:contribution-el} generalizes these works, as we show how
the robust formulation~\eqref{eqn:upper-lower-def} yields asymptotically exact
confidence intervals for general distributions $P_0$, and general constrained
$(\fr \subset \R^d)$ stochastic optimization problems.

Ben-Tal et al.'s robust sample approximation~\cite{Ben-TalHeWaMeRe13} and
Bertsimas et al.'s goodness of fit testing-based
procedures~\cite{BertsimasGuKa14} provide natural motivation for
formulations similar to ours~\eqref{eqn:upper-lower-def}.  However, by
considering completely nonparametric measures of fit we can depart from
assumptions on the structure of $\statdomain$ (i.e.\ that it is finite or a
compact subset of $\R^m$). The $f$-divergence
formulation~\eqref{eqn:upper-lower-def} allows for a more nuanced
understanding of the underlying structure of the population
problem~\eqref{eqn:pop}, and it also allows the precise confidence
statements, expansions, and consistency guarantees outlined
in~\eqref{item:contribution-el}--\eqref{item:contribution-expansion}.
Concurrent with the initial \texttt{arXiv} version of this work,
Lam~\cite{Lam16,Lam18} develops variance expansions similar to
ours~\eqref{eqn:variance-expansion-intro}, focusing on the KL-divergence and
empirical likelihood cases (i.e.\ $f(t) = -2 \log t$ with i.i.d.\ data). Our
methods of proof are different, and our expansions hold almost-surely (as
opposed to in probability), apply to general $f$-divergences, and generalize
to dependent sequences under standard ergodicity conditions.


The recent line of work on distributionally robust optimization using
Wasserstein distances~\cite{PflugWo07, Wozabal12, EsfahaniKu17,
  Shafieezadeh-AbadehEsKu15, BlanchetMu16, SinhaNaDu17} is similar in spirit
to the formulation considered here. Unlike $f$-divergences, uncertainty
regions formed by Wasserstein distances contain distributions that have
support different to that of the empirical distribution. Using concentration
results for Wasserstein distances with light-tailed random
variables~\cite{FournierGu15},~\citet{EsfahaniKu17} gave finite sample
guarantees with nonparametric rates $O(n^{-1/d})$. The $f$-divergence
formulation we consider yields different statistical guarantees; for random
variables with only second moments, we give confidence bounds that achieve
(asymptotically) \emph{exact} coverage at the parametric rate
$O(n^{-1/2})$. 
Further, the robustification approach via Wasserstein distances is often
computationally challenging (with current techology), as tractable convex
formulations are available~\cite{Shafieezadeh-AbadehEsKu15, EsfahaniKu17} only
under stringent conditions on the functional
$\statval \mapsto \loss(x; \statval)$.  On the other hand, efficient solution
methods~\cite{Ben-TalHeWaMeRe13, NamkoongDu16} for the robust
problem~\eqref{eqn:upper} are obtainable without restriction on the objective
function $\obj$ other than convexity in $x$.

\paragraph{Notation}

We collect our mostly standard notation here.  For a sequence of random
variables $X_1, X_2, \ldots$ in a metric space $\mathcal{X}$, we say
$X_n \cd X$ if $\E[f(X_n)] \to \E[f(X)]$ for all bounded continuous functions
$f$.  We write $X_n \cpstar X$ for random variables $X_n$ converging to a
random variable $X$ in outer probability~\cite[Section 1.2]{VanDerVaartWe96}.
Given a set $A \subset \R^d$, norm $\norm{\cdot}$, and point $x$, the distance
$\dist(x, A) = \inf_y \{\norm{x - y} : y \in A\}$.  The \emph{inclusion
  distance}, or the \emph{deviation}, from a set $A$ to $B$ is
\begin{equation}
  \dinclude(A, B) \defeq \sup_{x \in A} \dist(x, B)
  = \inf \left\{\epsilon \ge 0
    : A \subset \{y : \dist(y, B) \le \epsilon \} \right\}.
  \label{eqn:deviation-distance}
\end{equation}
For a measure $\mu$ on a measurable space $(\statdomain, \sigalg)$ and $p
\ge 1$, we let $L^p(\mu)$ be the usual $L^p$ space, that is, $L^p(\mu)
\defeq \{f : \statdomain \to \R \mid \int |f|^p d\mu < \infty\}$. For a
deterministic or random sequence $a_n \in \R$, we say that a sequence of
random variables $X_n$ is $O_P(a_n)$ if $\lim_{c \to \infty} \limsup_n
P(|X_n| \ge c \cdot a_n) = 0$.  Similarly, we say that $X_n = o_P(a_n)$ if
$\limsup P(|X_n| \ge c \cdot a_n) = 0$ for all $c > 0$.  For $\alpha \in [0,
  1]$, we define $\chi_{d, \alpha}^2$ to be the $\alpha$-quantile of a
$\chi_d^2$ random variable, that is, the value such that $\P(\ltwo{Z}^2 \le
\chi_{d,\alpha}^2) = \alpha$ for $Z \sim \normal(0, I_{d \times d})$.  The
Fenchel conjugate of a function $f$ is $f^*(y) = \sup_x \{y^Tx -
f(x)\}$. For a convex function $f : \R \to \R$, we define the right
derivative $f'_+(x) = \lim_{\delta \downarrow 0} \frac{f(x + \delta) -
  f(x)}{\delta}$, which must exist~\cite{HiriartUrrutyLe93ab}. We let
$\bigindic{A}(x)$ be the $\{0,\infty\}$-valued membership function, so
$\bigindic{A}(x) = \infty$ if $x \not \in A$, 0 otherwise. To address
measurability issues, we use outer measures and corresponding convergence
notions~\cite[Section 1.2-5]{VanDerVaartWe96}. Throughout the paper, the
sequence $\{\statrv_i\}$ is i.i.d.\ unless explicitly stated.

\subsection*{Outline}

In order to highlight applications of our general results to stochastic
optimization problems, we first present results for the optimal value
functional $T_{\rm opt}(P) \defeq \inf_{x \in \mathcal{X}}~ \E_P[\obj]$,
before presenting their most general forms.  In Section~\ref{section:stat}, we
first describe the necessary background on generalized empirical likelihood
and establish our basic variance expansions. We apply these results in
Section~\ref{section:el-stoch-opt} to stochastic optimization problems,
including those involving dependent data, and give computationally tractable
procedures for solving the robust formulation~\eqref{eqn:upper}.  In
Section~\ref{section:robust}, we develop the connections between
distributional robustness and principled choices of the size $\tol$ in the
uncertainty sets $\{P : \fdivs{P}{\emp} \le \tol/n\}$, choosing $\tol$ to
obtain asymptotically exact bounds on the population optimal
value~\eqref{eqn:pop}. To understand that the cost of the types of robustness
we consider is reasonably small, in Section~\ref{section:consistency} we show
consistency of the empirical robust optimizers under (essentially) the same
conditions guaranteeing consistency of SAA.  We conclude the ``applications''
of the paper to optimization and modeling with numerical investigation in
Section~\ref{section:sim}, demonstrating benefits and drawbacks of the
robustness approach over classical stochastic approximations.  To conclude the
paper, we present the full generalization of empirical likelihood theory to
$f$-divergences, Hadamard differentiable functionals, and uniform (Donsker)
classes of random variables in Section~\ref{section:general}.

%% file: generalized-el.tex
\newcommand{\hclass}{\mathcal{H}}
\newcommand{\dualh}{\mathcal{L}(\hclass)}
\newcommand{\hnorm}[1]{\norm{#1}_{\hclass}}

\section{Generalized Empirical Likelihood and Asymptotic Expansions}
\label{section:stat}

We begin by briefly reviewing generalized empirical likelihood
theory~\cite{Owen01,NeweySm04,Imbens02}, showing classical
results in Section~\ref{subsection:el} and then turning to
our new expansions in Section~\ref{section:asymptotic-expansion}.
Let $Z_1, \ldots, Z_n$ be
independent random vectors---formally, measurable functions $Z : \statdomain
\to \mathbb{B}$ for some Banach space $\mathbb{B}$---with common
distribution $P_0$. Let $\mathcal{P}$ be the set of probability
distributions on $\statdomain$ and let $T: \mathcal{P} \to \R$ be the
statistical quantity of interest. We typically consider $T_{\rm opt}(P) =
\inf_{ x\in \mathcal{X}} \E_{P}[\obj]$ with $Z(\statval) \defeq \loss(\cdot;
\statval)$, although our theory applies in more generality (see
Section~\ref{section:general}). The \emph{generalized empirical
  likelihood confidence region} for $T(P_0)$ is
\begin{equation*}
  C_{n,\tol} \defeq \left\{ T(P): \fdivs{P}{\emp} \le \frac{\tol}{n} \right\},
\end{equation*}
where $\emp$ is the empirical distribution of $Z_1, \ldots, Z_n$.  The set
$C_{n,\tol}$ is the image of $T$ on an $f$-divergence neighborhood of the
empirical distribution $\emp$.  We may define a dual quantity, the profile
divergence $R_n : \R \to \R_+$ (called the profile likelihood~\cite{Owen01}
when $f(t) = -2 \log t$), by
\begin{equation*}
  R_n(\theta) \defeq \inf_{P \ll \emp}
  \left\{ \fdivs{P}{\emp}: T(P) = \theta \right\}.
\end{equation*}
Then for any $P \in \mathcal{P}$, we have $T(P) \in C_{n,\tol}$ if and only
if $R_n(T(P)) \le \frac{\tol}{n}$.  Classical empirical
likelihood~\cite{Owen88,Owen90,Owen01} considers $f(t) = -2\log t$ so that
$\fdivs{P}{\emp} = 2 \dkls{\emp}{P}$, in which case the divergence is the
nonparametric log-likelihood ratio.  To
show that $C_{n,\tol}$ is actually a meaningful confidence set,
the goal is then to demonstrate that
(for appropriately smooth functionals $T$)
\begin{equation*}
  \P(T(P_0) \in C_{n,\tol})
  = \P\left(R_n(T(P_0)) \le \frac{\tol}{n}\right)
  \to 1 - \alpha(\tol)
  ~~ \mbox{as} ~~
  n \to \infty,
\end{equation*}
where $\alpha(\tol)$ is a desired confidence level (based on $\tol$) for the
inclusion $T(P_0) \in C_{n,\tol}$.


\subsection{Generalized Empirical Likelihood for Means}
\label{subsection:el}

In the classical case in which the vectors $Z_i \in \R^d$ and are i.i.d.,
\citet{Owen90} shows that empirical likelihood applied to the mean $T(P_0)
\defeq \E_{P_0}[Z]$ guarantees elegant asymptotic properties: when $\cov(Z)$
has rank $d_0 \le d$, as $n \to \infty$ one has $R_n(\E_{P_0}[Z]) \cd
\chi_{d_0}^2$, where $\chi_{d_0}^2$ denotes the $\chi^2$-distribution with
$d_0$ degrees of freedom. Then $C_{n,\tol(\alpha)}$ is an asymptotically
exact $(1-\alpha)$-confidence interval for $T(P_0) = \E_{P_0}[Z]$
if we set $\tol(\alpha) = \inf\{\tol' : \P(\chi_{d_0}^2 \le \tol')
\ge 1 - \alpha\}$.  We extend these results to more general functions $T$
and to a variety of $f$-divergences satisfying the following condition,
which we henceforth assume without mention (each of our theorems requires
this assumption).
\begin{assumption}[Smoothness of $f$-divergence]
  \label{assumption:fdiv}
  The function $f: \R_+ \to \overline{\R}$ is convex, three times
  differentiable in a neighborhood of $1$, and satisfies
  $f(1) = f'(1) = 0$ and $f''(1) = 2$.
\end{assumption}
\noindent The assumption that $f(1) = f'(1) = 0$ is no loss of
generality, as the function $t \mapsto f(t) + c (t - 1)$ yields
identical divergence measures to $f$, and the assumption that
$f''(1) = 2$ is a normalization for easier calculation. We make
no restrictions on the behavior of $f$ at $0$, as a number of
divergence measures, such as KL with
$f(t) = -2 \log t + 2t - 2$, approach infinity as $t \downarrow
0$.

The following proposition is a generalization of Owen's
results~\cite{Owen90} to smooth $f$-divergences. While the result is
essentially known~\cite{Baggerly98, Corcoran98, BertailGaHa14}, it is also
an immediate consequence of our uniform variance expansions to come.
\begin{proposition}
  \label{proposition:conf}
  Let Assumption~\ref{assumption:fdiv} hold. Let $Z_i \in \R^d$ be
  drawn i.i.d.\ $P_0$ with finite covariance of rank $d_0 \le d$. Then
  \begin{equation}
    \label{eqn:pointwise-empirical-divergence}
    \lim_{n \to \infty}
    \P\left(\E_{P_0}[Z] \in \left\{\E_P[Z] : \fdivs{P}{\emp} \le \frac{\tol}{n}
    \right\}\right) = \P\left(\chi_{d_0}^2 \le \tol\right).
  \end{equation}
\end{proposition}
\noindent
When $d = 1$, the proposition is a direct consequence of
Lemma~\ref{lemma:key-expansion} to come; for more general dimensions $d$, we
present the proof in Appendix~\ref{subsection:el-proof}.  If we consider the
random variable $Z_x(\statval) \defeq \obj$, defined for each
$x \in \mathcal{X}$, Proposition~\ref{proposition:conf} allows us to construct
pointwise confidence intervals for the distributionally robust
problems~\eqref{eqn:upper-lower-def}. 


\subsection{Asymptotic Expansions}
\label{section:asymptotic-expansion}

To obtain inferential guarantees on
$T(P) = \inf_{x \in \mathcal{X}} \E_{P}[\obj]$, we require stronger results
than the pointwise guarantee~\eqref{eqn:pointwise-empirical-divergence}.  We
now develop an asymptotic expansion that essentially gives all of the major
distributional convergence results in this paper. Our results on convergence
and exact coverage build on two asymptotic expansions, which we now present.
In the statement of the lemma, we recall that a sequence $\{Z_i\}$ of
random variables is ergodic and stationary
if for all bounded functions $f : \R^k \to \R$ and $g : \R^m \to \R$ we have
\begin{equation*}
  \lim_{n \to \infty}
  \E[f(Z_t, \ldots, Z_{t + k - 1}) g(Z_{t + n}, \ldots, Z_{t + n + m - 1})]
  = \E[f(Z_1, \ldots, Z_k)] \E[g(Z_1, \ldots, Z_m)].
\end{equation*}
We then have the following lemma.
\begin{lemma}
  \label{lemma:key-expansion}
  Let $Z_1, Z_2, \ldots$ be a strictly stationary ergodic sequence of random
  variables with $\E[Z_1^2] < \infty$, and let
  Assumption~\ref{assumption:fdiv} hold. Let $s_n^2 = \E_{\emp}[Z^2] -
  \E_{\emp}[Z]^2$ denote the sample variance of $Z$. Then
  \begin{equation}
    \label{eqn:simple-expansion}
    \bigg|\sup_{P: \fdivs{P}{\emp} \le \frac{\tol}{n}}
    \E_P[Z] - \E_{\emp}[Z]
    - \sqrt{\frac{\tol}{n} s_n^2}\bigg|
    \le \frac{\varepsilon_n}{\sqrt{n}}
  \end{equation}
  where $\varepsilon_n \cas 0$.
\end{lemma}

\noindent See Appendix~\ref{appendix:expansion-proof} for the proof.
For intuition, we may rewrite the expansion~\eqref{eqn:simple-expansion} as
\begin{subequations}
  \label{eqn:simple-expansion-ip}
  \begin{align}
    \sup_{P : \fdivs{P}{\emp} \le \frac{\tol}{n}}
    \E_P[Z] & = \E_{\emp}[Z] + \sqrt{\frac{\tol}{n} \var_{\emp}(Z)}
    + \frac{\varepsilon_n^+}{\sqrt{n}}
    \label{eqn:simple-expansion-ip-sup} \\
    \inf_{P : \fdivs{P}{\emp} \le \frac{\tol}{n}}
    \E_P[Z] & = \E_{\emp}[Z] - \sqrt{\frac{\tol}{n} \var_{\emp}(Z)}
    + \frac{\varepsilon_n^-}{\sqrt{n}}
  \end{align}
\end{subequations}
with $\varepsilon_n^{\pm} \cas 0$,
where the second equality follows from symmetry. Applying the classical
central limit theorem and Slutsky's lemma, we then obtain
\begin{align*}
  \P\left(\sqrt{n} \left|\E_{P_0}[Z] - \E_{\emp}[Z]\right|
  \le \sqrt{\tol \var_{\emp}(Z)}
  \right)
  & \mathop{\to}_{n \uparrow \infty}
  \P( | N(0, 1) |\le \sqrt{\tol}) = \P(\chi^2_1 \le \tol),
\end{align*}
yielding Proposition~\ref{proposition:conf} in the case that $d = 1$.
Concurrently with the original version of this paper, \citet{Lam18} gives
an in-probability version of the result~\eqref{eqn:simple-expansion-ip}
(rather than almost sure) for the case $f(t) = -2 \log t$, corresponding to
empirical likelihood. Our proof is new, gives a probability 1 result,
and generalizes to ergodic stationary sequences.

Next, we show a uniform variant of the asymptotic
expansion~\eqref{eqn:simple-expansion-ip} which relies on local
Lipschitzness of our loss. While our results apply in significantly more
generality (see Section~\ref{section:general}), the following assumption
covers many practical instances of stochastic optimization problems.
\begin{assumption}
  \label{assumption:lipschitz}
  The set $\mathcal{X} \subset \R^d$ is compact, and there exists a measurable
  function $M: \statdomain \to \R_+$ such that for all
  $\statval \in \statdomain$, $\loss(\cdot; \statval)$ is
  $M(\statval)$-Lipschitz with respect to some norm $\norm{\cdot}$ on
  $\mathcal{X}$.
\end{assumption}

\begin{theorem}
  \label{theorem:real-asymptotic-expansion-stoch-opt}
  Let Assumption~\ref{assumption:lipschitz} hold with
  $\E_{P_0}[M(\statval)^2] < \infty$ and assume that $\E_{P_0}[|\loss(x_0;
    \statval)|^2] < \infty$ for some $x_0 \in \mathcal{X}$. If $\statval_i
  \simiid P_0$, then
  \begin{equation}
    \label{eqn:real-asymptotic-expansion-stoch-opt}
    \sup_{P : \fdivs{P}{\emp} \le \frac{\tol}{n}}
    \E_P[\loss(x, \statrv)]
    = \E_{\emp}[\loss(x, \statrv)]
    + \sqrt{\frac{\tol}{n} \var_{\emp}(\loss(x, \statrv))}
    + \varepsilon_n(x),
  \end{equation}
  where $\sup_{x \in \mathcal{X}} \sqrt{n} |\varepsilon_n(x)| \cpstar 0$.
\end{theorem}
\noindent This theorem is a consequence of the more general uniform expansions
we develop in Section~\ref{section:general}, in particular
Theorem~\ref{theorem:real-asymptotic-expansion}.  In addition to generalizing
classical empirical likelihood theory, these results also allow a novel proof
of the classical empirical likelihood result for means
(Proposition~\ref{proposition:conf}).


%% file: inference-stoch-opt.tex
\section{Statistical Inference for Stochastic Optimization}
\label{section:el-stoch-opt}


With our asymptotic expansion and convergence results in place, we now
consider application of our results to stochastic optimimization problems and
study the mapping
\begin{equation*}
  T_{\rm opt}: \mathcal{P} \to \R, ~~~
  P \mapsto T_{\rm opt}(P) \defeq \inf_{x\in\mathcal{X}}\E_P[\obj].
\end{equation*}
Although the functional $T_{\rm opt}(P)$ is nonlinear, we can provide
regularity conditions guaranteeing its smoothness (Hadamard
differentiability), so that the generalized empirical likelihood approach
provides asymptotically exact confidence bounds on $T_{\rm opt}(P)$.
Throughout this section, we make a standard assumption guaranteeing
existence of minimizers~\cite[e.g.][Theorem 1.9]{RockafellarWe98}.
\begin{assumption}
  \label{assumption:lsc}
  The function $\loss(\cdot; \statval)$ is proper and lower semi-continuous
  for $P_0$-almost all $\statval \in \statdomain$. Either $\xdomain$ is
  compact or $x \mapsto \E_{P_0}[\loss(x; \statrv)]$ is coercive, meaning
  $\E_{P_0}[\loss(x; \statval)] \to \infty$ as $\norm{x} \to \infty$.
\end{assumption}

In the remainder of this section, we explore the generalized empirical
likelihood approach to confidence intervals on the optimal value for both
i.i.d.\ data and dependent sequences (Sections~\ref{subsection:el-stoch-opt}
and~\ref{subsection:el-stoch-opt}, respectively), returning
in Section~\ref{subsection:computing-conf} to discuss a few
computational issues, examples, and generalizations.

\subsection{Generalized Empirical Likelihood for Stochastic Optimization}
\label{subsection:el-stoch-opt}

The first result we present applies in the case that the data
is i.i.d.
\begin{theorem}
  \label{theorem:el-stoch-opt}
  Let Assumptions~\ref{assumption:fdiv},~\ref{assumption:lipschitz}
  hold with $\E_{P_0}[M(\statval)^2] < \infty$ and
  $\E_{P_0}[|\loss(x_0; \statval)|^2] < \infty$ for some
  $x_0 \in \mathcal{X}$. If $\statval_i \simiid P_0$ and the optimizer
  $x\opt \defeq \argmin_{x\in\mathcal{X}}\E_{P_0}[\obj]$ is unique, then
  \begin{equation*}
    \lim_{n \to \infty}
    \P\left(T_{\rm opt}(P_0) \in \left\{T_{\rm opt}(P) : \fdiv{P}{\emp} \le \frac{\rho}{n}
      \right\}\right)
    = \P\left( \chi_1^2 \le \tol \right).
  \end{equation*}
\end{theorem}
\noindent
This result follows from a combination of two steps: the generalized
empirical likelihood theory for smooth functionals we give in
Section~\ref{section:general}, and a proof that the conditions of the
theorem are sufficient to guarantee smoothness of $T_{\rm opt}$.  See
Appendix~\ref{appendix:danskin-proof} for the full derivation.


Setting $\xdomain = \R^d$, meaning that the problem is unconstrained, and
assuming that the loss $x \mapsto \loss(x; \statval)$ is differentiable for
all $\statval \in \statdomain$, \citet{LamZh17} give a similar (but different)
result to Theorem~\ref{theorem:el-stoch-opt} for the special case that
$f(t) = -2 \log t$, which is the classical empirical likelihood setting. They
use first order optimality conditions as an estimating equation and apply
standard empirical likelihood theory~\cite{Owen01}. This approach gives a
non-pivotal asymptotic distribution; the limiting law is a
$\chi_r^2$-distribution with
$r = \mbox{rank}(\cov_{P_0}(\nabla \loss(x\opt; \statrv))$ degrees of freedom,
though $x\opt$ need not be unique in this approach. The resulting confidence
intervals are too conservative and fail to have (asymptotically) exact
coverage. The estimating equations approach also requires the loss
$\loss(\cdot; \statval)$ to be differentiable and the covariance matrix of
$(\loss(x\opt; \statval), \nabla_x \loss(x\opt; \statval))$ to be positive
definite for some
$x^* \in \argmin_{x \in \mathcal{X}} \E_{P_0}[\loss(x; \statrv)]$. In
contrast, Theorem~\ref{theorem:el-stoch-opt} applies to problems with general
constraints, as well as more general objective functions $\loss$ and
$f$-divergences, by leveraging smoothness properties (over the space of
probability measures) of the functional
$T_{\rm opt}(P) \defeq \inf_{x \in \mathcal{X}} \E_{P}[\obj]$. A consequence
of the more general losses, divergences, and exact coverage is that the
theorem requires the minimizer of $\E_{P_0}[\loss(x; \statrv)]$ to be unique.

Shapiro~\cite{Shapiro89,Shapiro91} develops a number of normal
approximations and asymptotic normality theory for stochastic
optimization problems. The normal
analogue of Theorem~\ref{theorem:el-stoch-opt} is that
\begin{equation}
  \label{eqn:normal-approximation}
  \sqrt{n} \left( \inf_{x \in \mathcal{X}} \E_{\emp}[\obj] -
    \inf_{x\in\mathcal{X}} \E_{P_0}[\obj] \right) \cd N\left(0, \var_{P_0}(
  \loss(x\opt; \statrv))\right)
\end{equation}
which holds under the conditions of Theorem~\ref{theorem:el-stoch-opt}. The
normal approximation~\eqref{eqn:normal-approximation} depends on the unknown
parameter $\var_{P_0}(\loss(x\opt; \statrv))$ and is not asymptotically
pivotal. The generalized empirical likelihood approach, however, is pivotal,
meaning that there are no hidden quantities we must estimate; generally
speaking, the normal approximation~\eqref{eqn:normal-approximation} requires
estimation of $\var_{P_0}(\loss(x\opt; \statrv))$, for which one usually uses
$\var_{\emp}(\loss(\what{x}_n; \statrv))$ where $\what{x}_n$ minimizes the
sample average~\eqref{eqn:saa}.

When the optimum is not unique, we can still provide an exact asymptotic
characterization of the limiting probabilities that
$l_n \le T_{\rm opt}(P_0) \le u_n$, where we recall the
definitions~\eqref{eqn:upper-lower-def} of
$l_n = \inf_P \{T_{\rm opt}(P) : \fdivs{P}{\emp} \le \tol/n\}$ and
$u_n = \sup_P\{T_{\rm opt}(P) : \fdivs{P}{\emp} \le \tol/n\}$, which also
shows a useful symmetry between the upper and lower bounds. The
characterization depends on the excursions of a non-centered Gaussian process
when $x\opt$ is non-unique, which unfortunately makes it hard to evaluate.  To
state the result, we require the definition of a few additional processes. Let
$G$ be the mean-zero Gaussian process with covariance
\begin{equation*}
  \cov(x_1, x_2) =
  \E[G(x_1) G(x_2)] = \cov(\loss(x_1; \statrv), \loss(x_2; \statrv))
\end{equation*}
for $x_1, x_2 \in \xdomain$, and define the non-centered processes
$H_+$ and $H_-$ by
\begin{equation}
  H_+(x) \defeq G(x) + \sqrt{\tol \var_{P_0}(\loss(x; \statrv))}
  ~~ \mbox{and} ~~
  H_-(x) \defeq G(x) -\sqrt{\tol \var_{P_0}(\loss(x; \statrv))}.
  \label{eqn:upper-lower-GPs}
\end{equation}
Letting
  $S_{P_0}\opt \defeq \argmin_{x \in \xdomain} \E_{P_0}[\loss(x; \statrv)]$
be the set of optimal solutions for the population problem~\eqref{eqn:pop},
we obtain the following theorem. (It is possible to extend this result
to mixing sequences, but we focus for simplicity on the i.i.d.\ case.)
\begin{theorem}
  \label{theorem:ucb}
  Let Assumptions~\ref{assumption:fdiv}, \ref{assumption:lipschitz},
  and \ref{assumption:lsc} hold, where 
  the Lipschitz constant $M$ satisfies
  $\E_{P_0}[M(\statval)^2] < \infty$. Assume there exists
  $x_0 \in \mathcal{X}$ such that
  $\E_{P_0}[|\loss(x_0; \statval)|^2] < \infty$. If $\statval_i \simiid P_0$,
  then
  \begin{equation*}
    \lim_{n\to \infty} \P\left( \inf_{x\in\mathcal{X}} \E_{P_0}[\obj] \le
      u_n \right)
    = \P\left( \inf_{x\in S_{P_0}\opt} H_+(x) \ge 0 \right)
  \end{equation*}
  and
  \begin{equation*}
    \lim_{n \to \infty} \P\left(\inf_{x \in \xdomain} \E_{P_0}[\obj] \ge l_n\right)
    = \P\left(\inf_{x \in S_{P_0}\opt} H_-(x) \le 0 \right).
  \end{equation*}
  If $S_{P_0}\opt$ is
  a singleton, both limits are equal to
  $1 - \frac{1}{2} P\left(\chi^2_1 \ge \tol\right)$.
\end{theorem}
\noindent
We defer the proof of the theorem to Appendix~\ref{subsection:ucb-proof},
noting that it is essentially an immediate consequence of the uniform
results in Section~\ref{section:general} (in particular, the uniform
variance expansion of Theorem~\ref{theorem:real-asymptotic-expansion} and
the Hadamard differentiability result of Theorem~\ref{theorem:el-smooth}).

Theorem~\ref{theorem:ucb} provides us with a few benefits. First, if all one
requires is a one-sided confidence interval (say an upper interval), we may
shorten the confidence set via a simple correction to the threshold
$\tol$. Indeed, for a given desired confidence level $1-\alpha$, setting
$\tol = \chi^2_{1, 1-2\alpha}$ (which is smaller than $\chi^2_{1,1 -
  \alpha}$) gives a one-sided confidence interval $(-\infty, u_n]$ with
  asymptotic coverage $1-\alpha$.

\subsection{Extensions to Dependent Sequences}
\label{subsection:mixing}

We now give an extension of Theorem~\ref{theorem:el-stoch-opt} to dependent
sequences, including Harris recurrent Markov
chains mixing suitably quickly.
To present our results, we
first recall $\beta$-mixing sequences~\cite[Chs.~7.2--3]{Bradley05,EthierKu09}
(also called absolute regularity in the literature).
\begin{definition}
  \label{def:beta}
  The $\beta$-mixing coefficient between two sigma algbras $\mathcal{B}_1$ and
  $\mathcal{B}_2$ on $\statdomain$ is
  \begin{equation*}
    \beta(\mathcal{B}_1, \mathcal{B}_2) = \frac{1}{2}
    \sup \sum_{\mathcal{I} \times \mathcal{J}} \left|
      \P(A_i \cap B_j ) - \P(A_i) \P(B_j) \right|
  \end{equation*}
  where the supremum is over finite partitions
  $\{ A_i \}_{i \in \mathcal{I}}$, $\{ B_j \}_{j \in \mathcal{J}}$ of
  $\statdomain$ such that $A_i \in \mathcal{B}_1$ and $B_j \in \mathcal{B}_2$.
\end{definition}

Let $\{ \statrv \}_{i \in \Z}$ be a sequence of strictly stationary random
vectors. Given the $\sigma$-algebras
\begin{equation*}
  \mathcal{G}_0 \defeq \sigma(\statval_i: i \le 0)
  ~~~\mbox{and}~~~
  \mathcal{G}_n \defeq \sigma(\statval_i: i \ge n)
  ~~\mbox{for}~~ n \in \N,
\end{equation*}
the $\beta$-mixing coefficients of $\{\statval_i\}_{i \in \Z}$ are
defined via Definition~\ref{def:beta} by
\begin{equation}
  \label{eqn:beta-mixing-coefficient}
  \beta_n \defeq \beta(\mathcal{G}_0, \mathcal{G}_n).
\end{equation}
A stationary sequence $\{\statval_i\}_{i \in \Z}$ is $\beta$-mixing if
$\beta_n \to 0$ as $n \to \infty$.
For Markov chains,
$\beta$-mixing has a particularly nice interpretation:
a strictly stationary Markov chain
is $\beta$-mixing if and only if it is Harris recurrent and
aperiodic~\cite[Thm.~3.5]{Bradley05}.

With these preliminaries, we may state our asymptotic convergence
result, which is based on a uniform central limit
theorem that requires fast enough mixing~\cite{DoukhanMaRi95}.
\begin{theorem}
  \label{theorem:el-stoch-opt-ergodic}
  Let $\{\statrv_n\}_{n=0}^{\infty}$ be an aperiodic, positive Harris
  recurrent Markov chain taking values on $\statdomain$ with stationary
  distribution $\pi$. Let Assumptions~\ref{assumption:fdiv}
  and~\ref{assumption:lipschitz} hold and assume that there exists $r > 1$ and
  $x_0 \in \mathcal{X}$ satisfying
  $\sum_{n=1}^\infty n^{\frac{1}{r-1}} \beta_n < \infty$, the Lipschitz moment
  bound $\E_{\pi}[|M(\statval)|^{2r}] < \infty$, and
  $\E_{\pi}[|\loss(x_0; \statval)|^{2r}] < \infty$.  If the optimizer
  $x\opt \defeq \argmin_{x\in\mathcal{X}}\E_{\pi}[\obj]$ is unique then for
  any $\statval_0 \sim \nu$
  \begin{equation}
    \label{eqn:el-stoch-opt-ergodic}
    \lim_{n \to \infty}
    \P_{\nu}\left(T_{\rm opt}(\pi) \in \left\{T_{\rm opt}(P)
    : \fdiv{P}{\emp} \le \frac{\rho}{n} \right\}\right)
    = \P\left( \chi_1^2 \le \frac{\tol
      \var_{\pi} \hspace{1pt}
      \loss(x\opt; \statval)}{\sigma_{\pi}^2(x\opt)} \right)
  \end{equation}
  where
  $\sigma_{\pi}^2(x\opt) = \var_{\pi}\hspace{1pt} \loss(x\opt; \statval) +
  2\sum_{n=1}^\infty \cov_{\pi}(\loss(x\opt; \statval_0), \loss(x\opt;
  \statval_n))$.
\end{theorem}
\noindent
Theorem~\ref{theorem:el-stoch-opt-ergodic} is more or less a consequence
of the general results we prove in Section~\ref{section:ergodic}
on ergodic sequences, and we show how it follows from
these results in Appendix~\ref{sec:proof-el-stoch-opt-ergodic}.

We give a few examples of Markov chains satisfying the mixing condition
$\sum_{n=1}^\infty n^{\frac{1}{r-1}} \beta_n < \infty$ for some $r > 1$.
\ifdefined\usemorstyle
\vspace{5pt}
\else

\fi
\begin{example}[Uniform Ergodicity]
  If an aperiodic, positive Harris recurrent Markov chain is uniformly ergodic
  then it is geometrically $\beta$-mixing~\cite[Theorem
    16.0.2]{MeynTw09}, meaning that
  $\beta_n = O(c^n)$ for some constant $c \in (0, 1)$ In this case,
  the Lipschitzian assumption in Theorem~\ref{theorem:el-stoch-opt-ergodic}
  holds whenever $\E_\pi[M(\statrv)^2 \log_+ M(\statrv)] < \infty$.
\end{example}
\ifdefined\usemorstyle
\vspace{5pt}
\else
\\
\fi
As our next example, we consider geometrically $\beta$-mixing processses that
are not necessarily uniformly mixing. The following result is due
to~\citet{Mokkadem90}.
\ifdefined\usemorstyle
\vspace{5pt}
\else

\fi
\begin{example}[Geometric $\beta$-mixing]
  Let $\statdomain = \R^p$ and consider the affine auto-regressive process
  \begin{equation*}
    \statval_{n+1} = A(\epsilon_{n+1}) \statval_n + b(\epsilon_{n+1})
  \end{equation*}
  where $A$ is a polynomial $p \times p$ matrix-valued function and $b$ is a
  $\R^p$-valued polynomial function. We assume that the noise sequence
  $\{\epsilon_n\}_{n \ge 1} \simiid F$ where $F$ has a density with respect to
  the Lebesgue measure. If $(i)$ eigenvalues of $A(0)$ are inside the open
  unit disk and $(ii)$ there exists $a > 0$ such that
  $\E \norm{A(\epsilon_n)}^a + \E\norm{b(\epsilon_n)}^a < \infty$, then
  $\{\statval_n\}_{n \ge 0}$ is geometrically $\beta$-mixing. That is, there
  exists $c \in (0, 1)$ such that $\beta_n = O(s^n)$.
\end{example}
\ifdefined\usemorstyle
\vspace{5pt}
\else
\\
\fi
See~\citet[Section 2.4.1]{Doukhan94} for more
examples of $\beta$-mixing processes.

Using the equivalence of geometric $\beta$-mixing and geometric ergodicity
for Markov chains~\cite[Chapter 15]{NummelinTw78,MeynTw09},
we can give a Lyapunov criterion.
\ifdefined\usemorstyle
\vspace{5pt}
\else

\fi
\begin{example}[Lyapunov Criterion]
  \label{example:lyapunov}
  Let $\{ \statrv_n\}_{ n \ge 0}$ be an aperiodic Markov chain.  For
  shorthand, denote the regular conditional distribution of $\statrv_m$
  given $\statrv_0 = z$ by $P^m(z, \cdot) \defeq \P_{z}(\statrv_m \in \cdot)
  = \P(\statrv_m \in \cdot | \statrv_0 = z)$. Assume that there exists a
  measurable set $C \in \sigalg$, a probability measure $\nu$ on
  $(\statdomain, \sigalg)$, a potential function $w: \statdomain \to
  \openright{1}{\infty}$, and constants $m \ge 1, \lambda > 0, \gamma \in
  (0, 1)$ such that (i) $P^m(z, B) \ge \lambda \nu(B)$ for all $z \in C, B
  \in \sigalg$, (ii) $\E_z w(\statrv_1) \le \gamma w(z)$ for all $z \in
  C^c$, and (iii) $\sup_{z \in C} \E_z w(\statrv_1) < \infty$.  (The set $C$
  is a \emph{small set}~\cite[Chapter 5.2]{MeynTw09}.)  Then
  $\{\statval_n\}_{n \ge 0}$ is aperiodic, positive Harris recurrent, and
  geometrically ergodic~\cite[Theorem 15.0.1]{MeynTw09}. Further, we can
  show that $\{\statrv_n\}_{n \ge 0}$ is geometrically $\beta$-mixing: there
  exists $c \in (0, 1)$ with $\beta_n = O(c^n)$. For
  completeness, we include a proof of this in
  Appendix~\ref{section:example-lyapunov}.
\end{example}
\ifdefined\usemorstyle
\vspace{5pt}
\fi

For dependent sequences, the asymptotic distribution in the
limit~\eqref{eqn:el-stoch-opt-ergodic} contains unknown terms such as
$\sigma_\pi^2$ and $\var_{\pi}(\loss(x\opt; \statval))$; such quantities
need to be estimated to obtain exact coverage. This loss of a pivotal limit
occurs because
$\sqrt{n}(\emp - P_0)$ converges to a Gaussian process $G$ on
$\mathcal{X}$ with covariance function
\begin{equation*}
  \cov (x_1, x_2) \defeq \cov (G(x_1), G(x_2))
  =  \sum_{n \ge 1}
  \cov_{\pi}\left( \loss(x_1; \statrv_0), \loss(x_2; \statrv_n) \right),
\end{equation*}
while empirical likelihood self-normalizes based on $\cov_{\pi} \left(
\loss(x_1; \statrv_0), \loss(x_2; \statrv_0)\right)$. (These covariances are
identical if $\statrv_i$ are i.i.d.) As a result, in
Theorem~\ref{theorem:el-stoch-opt-ergodic}, we no longer have the
self-normalizing behavior of Theorem~\ref{theorem:el-stoch-opt} for
i.i.d.\ sequences. To remedy this, we now give a sectioning method that
yields pivotal asymptotics, even for dependent sequences.

Let $m \in \N$ be a fixed integer. Letting $b \defeq \floor{n / m}$, partition
the $n$ samples into $m$ sections
\begin{equation*}
  \{ \statval_1, \ldots, \statval_{b} \},~
  \{ \statval_{b+1}, \ldots, \statval_{2b} \},~
  \cdots,~
  \{ \statval_{(m-1)b + 1}, \ldots, \statval_{m b} \}
\end{equation*}
and denote by $\what{P}^{j}_b$ the empirical distribution on each of the
blocks for $j = 1,\ldots, m$. Let
\begin{equation*}
  U^{i}_b \defeq \sup_{P \ll \what{P}^j_b} \left\{ T_{\rm opt}(P):
    \fdiv{P}{\what{P}^j_b} \le \frac{\tol}{n} \right\}
\end{equation*}
and define
\begin{equation*}
  \wb{U}_b \defeq \frac{1}{m} \sum_{j=1}^m U^j_b
  ~~~\mbox{and}~~~
  s_m^2(U_b) \defeq
  \frac{1}{m} \sum_{j=1}^m \left( U^{j}_b - \wb{U}_b \right)^2.
\end{equation*}
Letting $\what{x}_{n}^* \in \argmin_{x \in \mathcal{X}} \E_{\emp}[\obj]$, we
obtain the following result. 
\begin{proposition}
  \label{proposition:sectioning}
  Under the conditions of Theorem~\ref{theorem:el-stoch-opt-ergodic}, for any
  initial distribution $\statval_0 \sim \nu$
  \begin{equation*}
    \lim_{n \to \infty}
    \P_{\nu} \left( T_{\rm opt}(\pi) \le \wb{U}_b - \sqrt{\frac{\tol}{b}
        \var_{\emp} \loss(\what{x}_{n}^*; \statval)}
      + s_m(U_b) t \right)
    = \P(T_{m-1} \ge -t)
  \end{equation*}
  where $T_{m-1}$ is the Student-t distribution with $(m-1)$-degress of
  freedom.
\end{proposition}
\noindent See Section~\ref{section:proof-sectioning} for the proof of
Proposition~\ref{proposition:sectioning}.
Thus, we recover an estimable quantity guaranteeing an exact
confidence limit.

\subsection{Computing the Confidence Interval and its Properties}
\label{subsection:computing-conf}

For convex objectives, we can provide efficient procedures for computing our
desired confidence intervals on the optimal value $T_{\rm opt}(P_0)$.  We
begin by making the following assumption.
\begin{assumption}
  \label{assumption:convex}
  The set $\mathcal{X} \subset \R^d$ is convex and
  $\loss(\cdot; \statrv): \mathcal{X} \to \R$ is a proper closed convex
  function for $P_0$-almost all $\statval \in \statdomain$.
\end{assumption}
For $P$ finitely supported on $n$ points, the functional
$P \mapsto T_{\rm opt}(P) = \inf_{x\in \mathcal{X}} \E_{P}[\obj]$ is
continuous (on $\R^n$) because it is concave and finite-valued; as a
consequence, the set
\begin{align}
  \label{eqn:el-conf-crude}
  \left\{T_{\rm opt}(P) : \fdivs{P}{\emp} \le \tol/n\right\}
  = \left\{ \inf_{x\in\mathcal{X}} \sum_{i=1}^n p_i\loss(x; \statval_i) :
  p^\top \onevec = 1,~p \ge 0,~
  \sum_{i=1}^n f(np_i) \le \tol \right\},
\end{align}
is an interval, and in this section we discuss a few methods for computing it.
To compute the interval~\eqref{eqn:el-conf-crude}, we solve for the two
endpoints $u_n$ and $l_n$ of
expressions~\eqref{eqn:upper}--\eqref{eqn:lower}. 

The upper bound is computable using convex optimization methods under
Assumption~\ref{assumption:convex}, which follows from the coming results.
The first is a minimax theorem~\cite[Theorem VII.4.3.1]{HiriartUrrutyLe93}.
\begin{lemma}
  \label{lemma:saddle}
  Let Assumptions~\ref{assumption:lsc} and~\ref{assumption:convex} hold. Then
  \begin{equation}
    \label{eqn:saddle}
    u_n = \inf_{x\in\mathcal{X}} \sup_{p \in \R^n}
    \left\{ \sum_{i=1}^n p_i\loss(x; \statval_i):
      p^\top \onevec = 1,~p\ge 0,~
      \sum_{i=1}^n f(np_i) \le \tol \right\}.
  \end{equation}
\end{lemma}
\noindent
By strong duality, we can write the minimax problem~\eqref{eqn:saddle}
as a joint minimization problem.
\begin{lemma}[\citet{Ben-TalHeWaMeRe13}]
  \label{lemma:dual-problem}
  The following duality holds:
  \begin{equation}
    \label{eqn:robust-dual}
    \sup_{P \ll \emp} \left\{ \E_P[\loss(x; \statrv)]
    : \fdiv{P}{\emp} \leq \frac{\tol}{n} \right\}
    = \inf_{\lambda \geq 0, \eta \in \R}
    \left\{
    \E_{\emp}\left[\lambda f^* \left(\frac{\loss(x; \statrv) -
        \eta}{\lambda}\right)\right] + \frac{\tol}{n}\lambda + \eta
    \right\}.
  \end{equation}
\end{lemma}
\noindent When $x \mapsto \loss(x; \statrv)$ is convex in $x$, the
minimization~\eqref{eqn:saddle} is a convex problem because it is the supremum
of convex functions. The reformulation~\eqref{eqn:robust-dual} shows that we
can compute $u_n$ by solving a problem jointly convex in $\eta$, $\lambda$,
and $x$.

Finding the lower confidence bound~\eqref{eqn:lower} is in general not a
convex problem even when the loss $\loss(\cdot; \statval)$ is convex in its
first argument. With that said, the conditions of Theorem~\ref{theorem:ucb},
coupled with convexity, allow us to give an efficient two-step minimization
procedure that yields an estimated lower confidence bound $\what{l}_n$ that
achieves the asymptotic pivotal behavior of $l_n$ whenever the population
optimizer for problem~\eqref{eqn:pop} is unique. Indeed, let us assume the
conditions of Theorem~\ref{theorem:ucb} and
Assumption~\ref{assumption:convex}, additionally assuming that the set
$S_{P_0}\opt$ is a singleton. Then standard consistency
results~\cite[Chapter 5]{ShapiroDeRu09} guarantee that under our conditions,
the empirical minimizer $\what{x}_n = \argmin_{x \in \xdomain}
\E_{\emp}[\loss(x; \statrv)]$ satisfies $\what{x}_n \cas x\opt$, where
$x\opt = \argmin_{x \in \xdomain} \E_{P_0}[\loss(x; \statrv)]$. Now,
consider the one-step estimator
\begin{equation}
  \what{l}_n \defeq \inf_{P : \fdivs{P}{\emp} \le \tol/n}
  \E_P[\loss(\what{x}_n; \statrv)].
  \label{eqn:one-step-lower}
\end{equation}
Then by Theorem~\ref{theorem:real-asymptotic-expansion-stoch-opt},
we have
\begin{equation*}
  \what{l}_n = \frac{1}{n} \sum_{i = 1}^n \loss(\what{x}_n; \statrv_i)
  - \sqrt{\frac{\tol}{n} \var_{\emp}(\loss(\what{x}_n; \statrv))}
  + o_{P_0}(n^{-\half})
\end{equation*}
because $\what{x}_n$ is eventually in any open set (or set open relative to
$\xdomain$) containing $x\opt$.  Standard limit
results~\cite{VanDerVaartWe96} guarantee that $\var_{\emp}(\loss(\what{x}_n;
\statrv)) \cas \var_{P_0}(\loss(x\opt; \statrv))$, because $x \mapsto
\loss(x; \statval)$ is Lipschitzian by
Assumption~\ref{assumption:lipschitz}.  Noting that
$\E_{\emp}[\loss(\what{x}_n; \statrv)] \le \E_{\emp}[\loss(x\opt;
  \statrv)]$, we thus obtain
\begin{equation*}
  \inf_{P : \fdivs{P}{\emp} \le \tol/n}
  \E_P[\loss(\what{x}_n; \statrv)]
  \le \E_{\emp}[\loss(x\opt; \statrv)]
  - \sqrt{\frac{\tol}{n} \var_{P_0}(\loss(x\opt; \statrv))}
  + o_{P_0}(n^{-\half})
\end{equation*}
Defining $\sigma^2(x\opt) = \var_{P_0}(\loss(x\opt; \statrv))$ for
notational convenience and rescaling by $\sqrt{n}$, we have
\begin{equation*}
  \sqrt{n} \left(\E_{\emp}[\loss(x\opt; \statrv)]
    - \E_{P_0}[\loss(x\opt; \statrv)]
    - \sqrt{\frac{\tol}{n} \sigma^2(x\opt)}
    + o_{P_0}(n^{-\half})
  \right)
  \cd \normal\left(
    -\sqrt{\tol \sigma^2(x\opt)}, \sigma^2(x\opt)\right).
\end{equation*}
Combining these results, we have that that $\sqrt{n} (l_n -
\E_{P_0}[\loss(x\opt; \statrv)]) \cd \normal(-\sqrt{\tol \sigma^2(x\opt)},
\sigma^2(x\opt))$ (looking forward to
Theorem~\ref{theorem:real-asymptotic-expansion} and using
Theorem~\ref{theorem:el-stoch-opt}), and
\begin{equation*}
  l_n \le \what{l}_n
  \le \E_{\emp}[\loss(x\opt; \statrv)] - \sqrt{\frac{\tol}{n} \sigma^2(x\opt)}
  + o_{P_0}(n^{-\half}).
\end{equation*}

Summarizing, the one-step estimator~\eqref{eqn:one-step-lower} is upper and
lower bounded by quantities that, when shifted by
$-\E_{P_0}[\loss(x\opt; \statrv)]$ and rescaled by $\sqrt{n}$, are
asymptotically $\normal(-\sqrt{\tol \sigma^2(x\opt)}, \sigma^2(x\opt))$.
Thus under the conditions of Theorem~\ref{theorem:el-stoch-opt}
and Assumption~\ref{assumption:lipschitz},
the one-step estimator $\what{l}_n$ defined by
expression~\eqref{eqn:one-step-lower}
guarantees that
\begin{equation*}
  \lim_{n \to \infty}
  \P\left(\what{l}_n \le \E_{P_0}[\loss(x\opt; \statrv)]
    \le u_n \right)
  = \P\left(\chi_1^2 \le \tol \right),
\end{equation*}
giving a computationally feasible and asymptotically pivotal statistic.
We remark in passing that alternating by minimizing over $P : \fdivs{P}{\emp}
\le \tol/n$ and $x$ (i.e.\ more than the single-step minimizer) simply
gives a lower bound $\wt{l}_n$ satisfying $l_n \le \wt{l}_n \le \what{l}_n$,
which will evidently have the same convergence properties.


%% file: connections-ro.tex
\section{Connections to Robust Optimization and Examples}
\label{section:robust}

To this point, we have studied the statistical properties of generalized
empirical likelihood estimators, with particular application to estimating the
population objective $\inf_{x \in \xdomain} \E_{P_0}[\obj]$.  We now make
connections between our approach of minimizing worst-case risk over
$f$-divergence balls and approaches from the robust
optimization and risk minimization literatures. We first relate our approach
to classical work on coherent risk measures for optimization problems, after
which we briefly discuss regularization properties of the formulation.


\subsection{Upper Confidence Bounds as a Risk Measure}

Sample average approximation is optimistic~\cite{ShapiroDeRu09}, because
$\inf_{x \in \xdomain} \E[\obj] \ge \E[\inf_{x \in \xdomain} \frac{1}{n}
\sum_{i =1}^n \loss(x; \statrv_i)]$.  The robust formulation~\eqref{eqn:upper}
addresses this optimism by looking at a worst case objective based on the
confidence region $\{P : \fdivs{P}{\emp} \le \tol/n\}$. It is clear that the
robust formulation~\eqref{eqn:upper} is a coherent risk
measure~\cite[Ch.~6.3]{ShapiroDeRu09} of $\obj$: it is convex, monotonic in
the loss $\loss$, equivariant to translations $\loss \mapsto \loss + a$, and
positively homogeneous in $\loss$. A number of authors have studied coherent
risk measures measures~\cite{ArtznerDeEbHe99,RockafellarUr00, Krokhmal07,
  ShapiroDeRu09}, and we study their connections to statistical confidence
regions for the optimal population objective~\eqref{eqn:pop} below.

As a concrete example, we consider Krokhmal's higher-order
generalizations~\cite{Krokhmal07} of conditional value at risk, where for
$k_* \ge 1$ and a constant $c > 0$ the risk functional has the form
\begin{equation*}
  R_{k_*}(x; P)
  \defeq \inf_{\eta \in \R}
  \left\{(1 + c) \E_P\left[\hinge{\loss(x; \statrv) - \eta}^{k_*}\right]^\frac{1}{k_*}
  + \eta \right\}.
\end{equation*}
The risk $R_{k_*}$ penalizes upward deviations of the objective $\obj$ from
a fixed value $\eta$, where the parameter $k_* \ge 1$ determines the degree
of penalization (so $k_* \uparrow \infty$ implies substantial penalties for
upward deviations). These risk measures capture a natural type of risk
aversion~\cite{Krokhmal07}.

We can recover such formulations, thus providing asymptotic guarantees for
their empirical minimizers, via the robust formulation~\eqref{eqn:upper}.  To
see this, we consider the classical Cressie-Read family~\cite{CressieRe84} of
$f$-divergences. Recalling that $f^*$ denotes the Fenchel conjugate
$f^*(s) \defeq \sup_t \{st - f(t) \}$, for $k \in (-\infty, \infty)$ with
$k \not\in \{0, 1\}$, one defines
\begin{equation}
  \label{eqn:cressie-read}
  f_k(t) = \frac{t^{k} - kt + k - 1}{2k(k-1)}
  ~~~ \mbox{so} ~~~
  f_k^*(s) = \frac{2}{k}\left[ \hinge{\frac{k-1}{2}s + 1}^{k_*} - 1\right]
\end{equation}
where we define $f_k(t) = +\infty$ for $t < 0$, and $k_*$ is given by
$1/k + 1/k_* = 1$. We define $f_1$ and $f_0$ as their respective limits as
$k \to 0, 1$. (We provide the dual calculation $f_k^*$ in the proof of
Lemma~\ref{lemma:cressie-read-risk}.)  The family~\eqref{eqn:cressie-read}
includes divergences such as the $\chi^2$-divergence ($k=2$), empirical
likelihood $f_0(t) = -2\log t + 2t - 2$, and KL-divergence
$f_1(t) = 2 t \log t - 2t + 2$. All such $f_k$ satisfy
Assumption~\ref{assumption:fdiv}.

For the Cressie-Read family, we may compute the dual~\eqref{eqn:robust-dual}
more carefully by infimizing over $\lambda \ge 0$, which yields the following
duality result.  As the lemma is a straightforward consequence of
Lemma~\ref{lemma:dual-problem}, we defer its proof to
Appendix~\ref{sec:proof-cressie-read-risk}.
\begin{lemma}
  \label{lemma:cressie-read-risk}
  Let $k \in (1, \infty)$ and define
  $\mathcal{P}_n \defeq \{P : D_{f_k}(P |\!| \emp) \le \tol/n\}$. Then
  \begin{equation}
    \sup_{P \in \mathcal{P}_n}
    \E_P[\obj]
    = \inf_{\eta \in \R}
    \left\{
      \left(1 + \frac{k (k - 1) \tol}{2n}\right)^\frac{1}{k}
      \E_{\emp}\left[\hinge{\loss(x; \statrv) - \eta}^{k_*}\right]^\frac{1}{k_*}
      + \eta \right\}
    \label{eqn:cressie-read-risk}
  \end{equation}
\end{lemma}
\noindent
The lemma shows that we indeed recover a variant of the risk $R_{k_*}$, where
taking $\tol \uparrow \infty$ and $k \downarrow 1$ (so that
$k_* \uparrow \infty$) increases robustness---penalties for upward deviations
of the loss $\loss(x; \statrv)$---in a natural way.  The confidence guarantees
of Theorem~\ref{theorem:ucb}, on the other hand, show how (to within first
order) the asymptotic behavior of the risk~\eqref{eqn:cressie-read-risk}
depends only on $\tol$, as each value of $k$ allows upper confidence bounds on
the optimal population objective~\eqref{eqn:pop} with asymptotically exact
coverage.

\subsection{Variance Regularization}
\label{subsection:variance-regularization}

We now consider the asymptotic variance expansions of
Theorem~\ref{theorem:real-asymptotic-expansion-stoch-opt},
which is that
\begin{equation}
  \label{eqn:asymptotic-expansion}
  \sup_{P : \fdiv{P}{P_n} \le \frac{\tol}{n}}
  \E_P[\loss(x; \statrv)]
  = \E_{P_n}[\loss(x; \statrv)]
  + \sqrt{\frac{\tol}{n} \var_{P_n}(\loss(x; \statrv))}
  + \varepsilon_n(x)
\end{equation}
where $\sqrt{n} \sup_{ x\in \fr} |\varepsilon_n(x)| \cpstar 0$.  In a
companion to this paper, Duchi and Namkoong~\cite{DuchiNa16,NamkoongDu17}
explore the expansion~\eqref{eqn:asymptotic-expansion} in substantial depth
for the special case $f(t) = \half (t -
1)^2$. Eq.~\eqref{eqn:asymptotic-expansion} shows that in an asymptotic sense,
we expect similar results to theirs to extend to general $f$-divergences, and
we discuss this idea briefly.

The expansion~\eqref{eqn:asymptotic-expansion} shows that the robust
formulation~\eqref{eqn:upper} ameliorates the optimism bias of standard
sample average approximation by penalizing the variance of the loss.
Researchers have investigated connections between regularization and
robustness, including \citet{XuCaMa09} for standard supervised machine
learning tasks (see also~\cite[Chapter 12]{Ben-TalGhNe09}), though these
results consider uncertainty on the data vectors $\statrv$ themselves,
rather than the distribution. Our approach yields a qualitatively different
type of (approximate) regularization by variance. In our follow-up
work~\cite{DuchiNa16,NamkoongDu17}, we analyze finite-sample performance of
the robust solutions. The naive variance-regularized
objective
\begin{equation}
  \label{eqn:normal-ucb}
  \E_{\emp}[\obj] + \sqrt{\frac{\tol}{n}\var_{\emp} \obj}
\end{equation}
is neither convex (in general) nor coherent, so that the
expansion~\eqref{eqn:asymptotic-expansion} allows us to solve a convex
optimization problem that approximately regularizes variance.

In some restricted situations, the variance-penalized
objective~\eqref{eqn:normal-ucb} is convex---namely, when $\obj$ is linear in
$x$. A classical example of this is the sample version of the Markowitz
portfolio problem~\cite{Markowitz52}.
\ifdefined\usemorstyle
\vspace{5pt}
\else

\fi
\begin{example}[Portfolio Optimization]
  \label{example:portfolio}
  Let $x \in \R^d$ denote
  investment allocations and $\statrv \in \R^d$ returns on investiment,
  and consider the optimization problem
 \begin{equation*}
   \maximize_{x\in\R^d}   ~ \E_{P_0}\left[\statrv^\top x \right]
    ~~\subjectto  ~~ x^\top \onevec = 1, x \in [a, b]^d.
  \end{equation*}
  Given a sample $\{\statrv_1, \ldots, \statrv_n\}$ of returns,
  we define $\mu_n \defeq \E_{\emp}[\statrv]$ and
  $\Sigma_n \defeq \cov_{\emp}(\statrv)$ to be the sample mean
  and covariance. Then the Lagrangian form of the Markowitz problem is to
  solve
 \begin{equation*}
   \maximize_{x\in\R^d}   ~ \mu_n^\top x
   - \sqrt{\frac{\tol}{n} x^\top \Sigma_n x}
    ~~\subjectto  ~~ x^\top \onevec = 1, x \in [a, b]^d.
  \end{equation*}
  The robust approximation of Theorem~\ref{theorem:real-asymptotic-expansion}
  (and Eq.~\eqref{eqn:asymptotic-expansion}) shows that
  \begin{align*}
    \inf \left\{\E_P[\statrv^\top x] ~ : ~ \fdivs{P}{\emp} \le \tol/n\right\}
    = \mu_n^\top x
    - \sqrt{\frac{\tol}{n} x^\top \Sigma_n x} +
    o_p(n^{-\frac{1}{2}}),
  \end{align*}
  so that distributionally robust formulation approximates the Markowitz
  objective to $o_p(n^{-\half})$. There are minor differences, however, in
  that the Markowitz problem penalizes both upward deviations (via the
  variance) as well as the downside counterpart. The robust formulation, on
  the other hand, penalizes downside risk only and is a coherent risk
  measure.
\end{example}


%% file: consistency.tex
\section{Consistency}
\label{section:consistency}

In addition to the inferential guarantees---confidence intervals and variance
expansions---we have thus far discussed, we can also give a number of
guarantees on the asymptotic consistency of minimizers of the robust upper
bound~\eqref{eqn:upper}.  We show that robust solutions are consistent under
(essentially) the same conditions required for that of sample average
approximation, which are more general than that required for the uniform
variance expansions of
Theorem~\ref{theorem:real-asymptotic-expansion-stoch-opt}.  We show this in
two ways: first, by considering uniform convergence of the robust
objective~\eqref{eqn:upper} to the population risk
$\E_{P_0}[\loss(x; \statrv)]$ over compacta
(Section~\ref{sec:uniform-convergence}), and second by leveraging epigraphical
convergence~\cite{RockafellarWe98} to allow unbounded feasible region
$\mathcal{X}$ when $\loss(\cdot; \statval)$ is convex
(Section~\ref{sec:epi-convergence}).  In the latter case, we require no
assumptions on the magnitude of the noise in estimating
$\E_{P_0}[\loss(x; \statrv)]$ as a function of $x \in \xdomain$; convexity
forces the objective to be large far from the minimizers, so the noise cannot
create minimizers far from the solution set.

\citet{BertsimasGuKa14} also provide consistency results for robust variants
of sample average approximation based on goodness-of-fit tests, though they
require a number of conditions on the domain $\statdomain$ of the random
variables for their results (in addition to certain continuity conditions on
$\loss$). 
In our context, we abstract away from this by parameterizing our problems by
the $n$-vectors $\{P : \fdivs{P}{\emp} \le \tol/n\}$ and give more direct
consistency results that generalize to mixing sequences.

\subsection{Uniform Convergence}
\label{sec:uniform-convergence}

For our first set of consistency results, we focus on uniform convergence of
the robust objective to the population~\eqref{eqn:pop}. We begin by
recapitulating a few standard statistical results abstractly. Let $\hclass$ be a
collection of functions $h : \statdomain \to \R$.
We have the following definition on uniform strong laws of large
numbers.
\begin{definition}
  A collection of functions $\hclass$, $h : \statdomain \to \R$ for
  $h \in \hclass$, is \emph{Glivenko-Cantelli} if
  \begin{equation*}
    \sup_{h \in \hclass} \left| \E_{\emp}[h] - \E_{P_0}[h] \right| \casstar 0.
  \end{equation*}
\end{definition}
\noindent
There are many conditions sufficient to guarantee Glivenko-Cantelli
properties. Typical approaches include covering number bounds on
$\hclass$~\cite[Chapter 2.4]{VanDerVaartWe96}; for example, Lipschitz
functions form a Glivenko-Cantelli class, as do continuous functions that
are uniformly dominated by an integrable function in the
next example.

\begin{example}[Pointwise compact class~\cite{VanDerVaart98},
  Example 19.8]
  \label{example:gc-conti}
  Let $\mathcal{X}$ be compact and $\loss(\cdot; \statval)$ be continuous in
  $x$ for $P_0$-almost all $\statval \in \statdomain$. Then
  $\hclass = \{\loss(x; \cdot) : x \in \xdomain\}$ is Glivenko-Cantelli if
  there exists a measurable envelope function
  $Z : \statdomain \to \R_+$ such that
  $|\loss(x; \statval)| \le Z(\statval)$ for all $x \in \xdomain$ and
  $\E_{P_0}[Z] < \infty$.
\end{example}

\noindent If $\mathcal{H}$ is Glivenko-Cantelli for $\statval \simiid P_0$,
then it is Glivenko-Cantelli for $\beta$-mixing sequences~\cite{NobelDe93}
(those with coefficients~\eqref{eqn:beta-mixing-coefficient} $\beta_n \to 0$).
Our subsequent results thus apply to $\beta$-mixing sequences
$\{\statval_i\}$.

If there is an envelope function for objective $\obj$ that has more than one
moment under $P_0$, we can show that the robust risk converges uniformly to
the population risk (compared to just the first moment for SAA).
\begin{assumption}
  \label{assumption:moment}
  There exists $Z : \statdomain \to \R_+$ with
  $|\loss(x; \statval)| \le Z(\statval)$ for all $x \in \mathcal{X}$
  and $\epsilon > 0$ such that
  $ \E_{P_0}[Z(\statrv)^{1+\epsilon}]<\infty$.
\end{assumption}
\noindent
Under this assumption, we have the following theorem.
\begin{theorem}
  \label{theorem:uniform-convergence}
  Let Assumptions~\ref{assumption:fdiv} and~\ref{assumption:moment} hold,
  and assume the class $\{ \loss(x; \cdot): x \in \mathcal{X} \}$ is
  Glivenko-Cantelli.  Then
  \begin{equation*}
    \sup_{x\in\mathcal{X}} \sup_{P \ll \emp}\left\{ \left|\E_{P}[\loss(x; \statrv)]
        - \E_{P_0}[\loss(x; \statrv)] \right|:
      \fdiv{P}{\emp} \leq \frac{\tol}{n} \right\}
       \casstar 0
  \end{equation*}
\end{theorem}
\noindent
See Appendix~\ref{appendix:uniform-convergence-proof} for a proof of the
result. 


When uniform convergence holds, the consistency of robust solutions
follows. As in the previous section, we define the sets of optima
\begin{equation}
  \label{eqn:optimal-sets}
  S_{P_0}\opt \defeq \argmin_{x \in \xdomain} \E_{P_0}[\obj]
  ~~ \mbox{and} ~~
  S_{\emp}\opt \defeq \argmin_{x \in \mathcal{X}}
    \sup_{P \ll \emp}\left\{ \E_{P}[\loss(x; \statrv)]:
    \fdivs{P}{\emp} \leq \frac{\tol}{n} \right\}.
\end{equation}
Then we immediately attain the following corollary to
Theorem~\ref{theorem:uniform-convergence}. In the corollary,
we recall the definition of the inclusion distance, or deviation, between
sets~\eqref{eqn:deviation-distance}.
\begin{corollary}
  \label{corollary:uniform-consistency}
  Let Assumptions~\ref{assumption:fdiv} and~\ref{assumption:moment} hold, let
  $\xdomain$ be compact, and assume $\loss(\cdot; \statval)$ is continuous on
  $\xdomain$.  Then
  \begin{equation*}
    \inf_{x \in \mathcal{X}} \sup_{P \ll \emp}
    \left\{ \E_{P}[\loss(x; \statrv)]:
        \fdivs{P}{\emp} \leq \frac{\tol}{n} \right\}
      - \inf_{x \in \mathcal{X}} \E_{P_0}[\loss(x; \statrv)] \cpstar 0
  \end{equation*}
  and $\dinclude(S_{\emp}\opt, S_{P_0}\opt) \cpstar 0$.
\end{corollary}
\ifdefined\usemorstyle
\proof{Proof.}
\else
\begin{proof}
  \fi
  The first conclusion is immediate by
  Theorem~\ref{theorem:uniform-convergence} and
  Example~\ref{example:gc-conti}.
  To show convergence of the optimal sets, we denote by
  $A^{\epsilon} = \{ x : \dist(x, A) \le \epsilon \}$ the $\epsilon$-enlargement
  of $A$. By the uniform convergence given in
  Theorem~\ref{theorem:uniform-convergence} and continuous mapping
  theorem~\cite[Theorem 1.3.6]{VanDerVaartWe96}, for all $\epsilon > 0$
  \begin{align*}
    \limsup_{n \to \infty}
    \P^*\left(\dinclude(S_{\emp}\opt, S_{P_0}\opt) \ge \epsilon\right)
    & \le \limsup_{n \to \infty}
    \P^*\left( \inf_{x \in S_{P_0}^{\star \epsilon}} \what{F}_n(x)
    > \inf_{x \in \mathcal{X}} \what{F}_n(x)\right) \\
    & = \P^*\left( \inf_{x \in S_{P_0}^{\star \epsilon}} F(x)
    > \inf_{x \in \mathcal{X}} F(x)\right) = 0
  \end{align*}
  where
  $\what{F}_n(x) \defeq \sup_{P \ll \emp} \{ \E_{P}[\loss(x; \statrv)]:
  \fdivs{P}{\emp} \leq \frac{\tol}{n} \}$ and
  $F(x) := \E_{P_0}[\loss(x; \statrv)]$.
\ifdefined\usemorstyle
\endproof
\else
\end{proof}
\fi

\subsection{Consistency for convex problems}
\label{sec:epi-convergence}

When the function $\loss(\cdot; \statval)$ is convex, we can give
consistency guarantees for minimizers of the robust upper
bound~\eqref{eqn:upper} by leveraging epigraphical convergence
theory~\cite{KingWe91,RockafellarWe98}, bypassing the uniform convergence
and compactness conditions above.  Analogous results hold for sample average
approximation~\cite[Chapter 5.1.1]{ShapiroDeRu09}.

In the theorem, we let $S_{P_0}\opt$ and $S_{\emp}\opt$ be the solution
sets~\eqref{eqn:optimal-sets} as before. We require a much weaker variant of
Assumption~\ref{assumption:moment}: we assume that for some $\epsilon > 0$,
we have $\E[|\loss(x; \statrv)|^{1 + \epsilon}] < \infty$ for all $x \in
\mathcal{X}$.  We also assume there exists a function $g : \xdomain \times
\statdomain \to \R$ such that for each $x \in \xdomain$, there is a
neighborhood $U$ of $x$ with $\inf_{z \in U} \loss(z; \statval) \ge g(x,
\statval)$ and $\E[|g(x, \statval)|] < \infty$.  Then we have the following
result, whose proof we provide in Appendix~\ref{appendix:consistency-proof}.
\begin{theorem}
  \label{theorem:epi-consistency}
  In addition to the conditions of the previous paragraph, let
  Assumptions~\ref{assumption:fdiv},~\ref{assumption:lsc},
  and~\ref{assumption:convex} hold. Assume that
  $\E_{\emp}[|\loss(x; \statrv)|^{1+\epsilon}] \cas \E_{P_0}[|\loss(x;
  \statrv)|^{1+\epsilon}]$ for $x \in \mathcal{X}$. Then
  \begin{equation*}
    \inf_{x \in \xdomain} \sup_{P \ll \emp}
    \left\{\E_P[\obj] : \fdivs{P}{\emp} \le \frac{\tol}{n} \right\}
    \cpstar \inf_{x \in \xdomain} \E_{P_0}[\loss(x; \statrv)]
  \end{equation*}
  and $\dinclude(S_{\emp}\opt, S_{P_0}\opt) \cpstar 0$.
\end{theorem}
\noindent
By comparison with Theorem~\ref{theorem:uniform-convergence} and
Corollary~\ref{corollary:uniform-consistency}, we see that
Theorem~\ref{theorem:epi-consistency} requires weaker conditions on the
boundedness of the domain $\xdomain$, instead relying on the compactness of
the solution set $S_{P_0}\opt$ and the growth of $\E_{P_0}[\loss(x; \statrv)]$
off of this set, which means that eventually $S_{\emp}\opt$ is nearly
contained in $S_{P_0}\opt$. When $\{\statval_i\}$ are not i.i.d., the
pointwise strong law for $|\loss(x; \statrv)|^{1+\epsilon}$ holds
if $\{\statrv_i\}$ is strongly mixing
($\alpha$-mixing)~\cite{Ibragimov62}, so the theorem immediately
generalizes to dependent sequences.


%% file: simulations.tex
\section{Simulations}
\label{section:sim}
We present three simulation experiments in this section: portfolio
optimization, conditional value-at-risk optimization,
and optimization in the multi-item newsvendor model.
In each of our three simulations, we compute and compare the following
approaches to estimation and inference:
\begin{enumerate}[(i)]
\item
  We compute the generalized empirical likelihood confidence
  interval $[l_n, u_n]$ as in expression~\eqref{eqn:upper-lower-def},
  but we use the (computable) estimate $\what{l}_n$ of
  Eq.~\eqref{eqn:one-step-lower} in Section~\ref{subsection:computing-conf}.
  With these, we simulate the true coverage probability
  $\P(\inf_{x \in \mathcal{X}} \E_{P_0}[\obj] \in [\what{l}_n, u_n])$ (because
  we control the distribution $P_0$ and $\obj$) of our
  confidence intervals, and we compare it to the nominal $\chi^2$-confidence
  level $\P(\chi_1^2 \le \tol)$ our asymptotic theory in
  Section~\ref{section:el-stoch-opt}
  suggests.
\item We compute the coverage rates of the normal confidence
  intervals~\eqref{eqn:normal-approximation} at the same level as our $\chi^2$
  confidence level.
\end{enumerate}
Throughout our simulations (and for both the normal and generalized empirical
likelihood/robust approximations), we use the nominal 95\% confidence level,
setting $\tol = \chi^2_{1, 0.95}$, so that we attain the asymptotic coverage
$\P\left( \chi_1^2 \le \tol \right) = 0.95$. We focus on i.i.d. sequences and
assume that $\statrv_i \simiid P_0$ in the rest of the section.

To solve the convex optimization problems~\eqref{eqn:robust-dual}
and~\eqref{eqn:one-step-lower} to compute $u_n$ and $\what{l}_n$,
respectively, we use the \texttt{Julia} package
\texttt{convex.jl}~\cite{UdellMoZeHoDiBo14}. Each experiment consists of
$1250$ independent replications for each of the sample sizes $n$ we report,
and we vary the sample size $n$ to explore its effects on coverage
probabilities. In all of our experiments, because of its computational
advantages, we use the $\chi^2$-squared divergence
$f_2(t) = \frac{1}{2} (t-1)^2$. We summarize our numerical results in
Table~\ref{table:coverages}, where we simulate runs of sample size up to
$n = 10,000$ for light-tailed distributions, and $n = 100,000$ for
heavy-tailed distributions; in both cases, we see that actual coverage very
closely approximates the nominal coverage $95\%$ at the largest value of
sample size ($n$) reported. In Figure~\ref{fig:conf}, we plot upper/lower
confidence bounds and mean estimates, all of which are averaged over the
$1250$ independent runs.


\begin{table}[ht!]
  \centering
  \caption{Coverage Rates (nominal = 95\%)}
  \label{table:coverages}
    \begin{filecontents*}{./Simulations/coverage.csv}
sample size,pfo_el,pfo_normal,news_el,news_normal,cvar_el,cvar_normal,cvar_el_4,cvar_normal_4,cvar_el_6,cvar_normal_6
20,75.16,89.2,30.1,91.38,91.78,95.02,29,100,35.4,100
40,86.96,93.02,55.24,90.32,93.3,94.62,48.4,100,59.73,100
60,89.4,93.58,69.5,88.26,93.8,94.56,42.67,100,51.13,100
80,90.46,93.38,74.44,86.74,93.48,93.94,47.73,100,57.73,100
100,91,93.8,77.74,85.64,94.22,94.38,46.33,100,55.67,99.87
200,92.96,93.68,86.73,95.27,94.64,95.26,48.4,99.8,56.73,98.93
400,94.28,94.52,91,94.49,94.92,95.06,48.67,98.93,55.27,97.93
600,94.48,94.7,92.73,94.29,94.8,94.78,51.13,98.53,56.73,97.67
800,94.36,94.36,93.02,93.73,94.64,94.64,51.67,97.93,57.47,97.6
1000,95.25,95.15,92.84,94.31,94.62,94.7,53.07,98.47,58.6,97.33
2000,95.48,95.25,93.73,95.25,94.92,95.04,54.07,96.8,59.07,96.53
4000,96.36,95.81,95.1,95.78,95.3,95.3,58.6,96,62.07,96.6
6000,96.33,95.87,94.61,95,94.43,94.51,61.8,95.8,66.07,95.73
8000,96.46,95.9,94.56,94.71,94.85,94.85,64.67,95.67,69,95.33
10000,96.43,95.51,94.71,94.85,94.43,94.43,66.87,94.73,69.4,96.13
20000,,,,,,,74.27,95.8,76.8,96.13
40000,,,,,,,81.8,94.2,84.87,94.87
60000,,,,,,,86.87,93.93,89.47,94.47
80000,,,,,,,91.4,93.67,92.33,95
100000,,,,,,,94.2,94.33,95.07,95.2
\end{filecontents*}
\pgfplotstabletypeset[
    col sep=comma,
    string type,
    every head row/.style={%
      before row={\hline \%
        & \multicolumn{2}{c}{Portfolio} 
        & \multicolumn{2}{c}{Newsvendor} 
        & \multicolumn{2}{c}{CVaR Normal} 
        & \multicolumn{2}{c}{CVaR $\mbox{tail}~a=3$} 
        & \multicolumn{2}{c}{CVaR $\mbox{tail}~a= 5$} \\
      },
      after row=\hline
    },
    every last row/.style={after row=\hline},
    columns/sample/.style={column name=Sample, column type=l},
    columns/pfo_el/.style={column name=EL, column type=l},
    columns/pfo_normal/.style={column name=Normal, column type=c},
    columns/news_el/.style={column name=EL, column type=l},
    columns/news_normal/.style={column name=Normal, column type=c},
    columns/cvar_el/.style={column name=EL, column type=l},
    columns/cvar_normal/.style={column name=Normal, column type=c},
    columns/cvar_el_4/.style={column name=EL, column type=l},
    columns/cvar_normal_4/.style={column name=Normal, column type=c},
    columns/cvar_el_6/.style={column name=EL, column type=l},
    columns/cvar_normal_6/.style={column name=Normal, column type=c}
    ]{./Simulations/coverage.csv}
\end{table}

\subsection{Portfolio Optimization}

\begin{figure}[ht!] 
  \begin{subfigure}[1]{0.5\linewidth}
    \centering
    \includegraphics[width=1\linewidth]{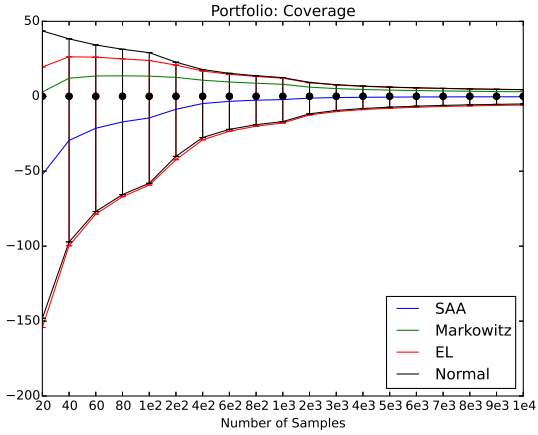} 
    \caption{Portfolio Optimization} 
    \label{fig:pfo} 
    \vspace{0ex}
  \end{subfigure}
  \begin{subfigure}[2]{0.5\linewidth}
    \centering
    \includegraphics[width=1\linewidth]{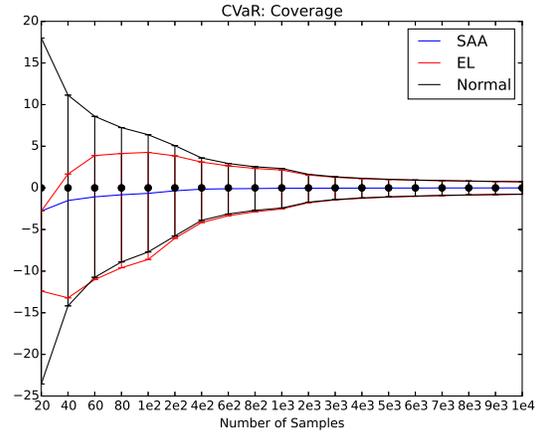} 
    \caption{Conditional Value-at-Risk} 
    \label{fig:cvar} 
    \vspace{0ex}
  \end{subfigure}
  \centering
  \begin{subfigure}[3]{0.5\linewidth}
    \centering
    \includegraphics[width=1\linewidth]{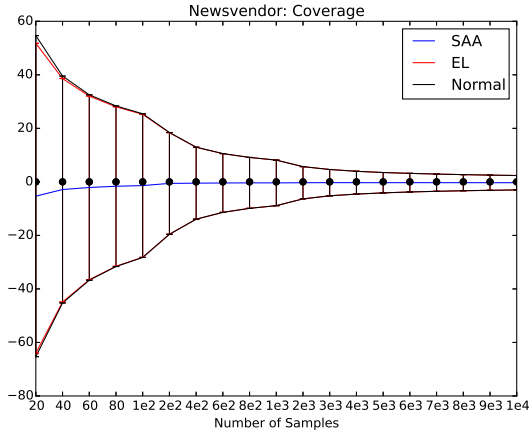} 
    \caption{Multi-item Newsvendor} 
    \label{fig:news} 
  \end{subfigure}
  \caption{\label{fig:conf} Average confidence sets for
    $\inf_{x \in \xdomain} \E_{P_0}[\loss(x; \statrv)]$ for normal
    approximations~\eqref{eqn:normal-approximation} (``Normal'') and the
    generalized empirical likelihood confidence
    set~\eqref{eqn:upper-lower-def} (``EL'').  The true value being
    approximated in each plot is centered at $0$.  The optimal objective
    computed from the sample average approximation (``SAA'') has a negative
    bias.}
\end{figure}

Our first simulation investigates the standard portfolio optimization problem
(recall Example~\ref{example:portfolio}, though we \emph{minimize} to be
consistent with our development). We consider problems in dimension $d = 20$
(i.e.\ there are 20 assets). For this problem, the objective is
$\loss(x; \statrv) = x^\top \statrv$, we set
$\mathcal{X} = \{x \in \R^d \mid \ones^\top x = 1, -10 \le x \le 10\}$ as our
feasible region (allowing leveraged investments), and we simulate returns
$\statrv \simiid N(\mu, \Sigma)$.  Within each simulation, the vector $\mu$
and covariance $\Sigma$ are chosen as $\mu \sim N(0, I_d)$ and $\Sigma$ is
standard Wishart distributed with $d$ degrees of freedom. The population
optimal value is $\inf_{x \in \xdomain} \mu^\top x$. As $\mu \in \R^d$ has
distinct entries, the conditions of Theorem~\ref{theorem:el-stoch-opt} are
satisfied because the population optimizer is unique. We also consider the
(negative) Markowitz problem
\begin{equation*}
  \minimize_{x \in \mathcal{X}} ~ \E_{\emp}[x^\top \statrv]
  + \sqrt{\frac{\tol}{n}\var_{\emp} (x^\top \statrv)},
\end{equation*}
as the variance-regularized expression is efficiently minimizable (it is
convex) in the special case of linear objectives. In Figure~\ref{fig:pfo}, we
plot the results of our simulations. The vertical axis is the estimated
confidence interval for the optimal solution value for each of the methods,
shifted so that $0 = \mu^\top x\opt$, while the horizontal axis is the sample
size $n$. We also plot the estimated value of the objective returned by the
Markowitz optimization (which is somewhat conservative) and the estimated
value given by sample average approximation (which is optimistic), averaging
the confidence intervals over 1250 independent simulations. Concretely, we see
that the robust/empirical likelihood-based confidence interval at $n = 20$ is
approximately $[-150, 40]$, and the Markowitz portfolio is the line slightly
above 0, but below each of the other estimates of expected returns.  In
Table~\ref{table:coverages}, we give the actual coverage rates--- the fraction
of time the estimated confidence interval contains the true value
$\mu^\top x\opt$.  In comparison with the normal confidence interval,
generalized empirical likelihood undercovers in small sample settings, which
is consistent with previous observations for empirical likelihood (\eg,
\cite[Sec 2.8]{Owen01}).

\subsection{Conditional Value-at-Risk}

For a real-valued random variable $\statrv$, the \emph{conditional
  value-at-risk} $\alpha$ (CVaR) is the expectation of $\statrv$ conditional
on it taking values above its $1 - \alpha$ quantile~\cite{RockafellarUr00}.
More concisely, the CVaR (at $\alpha$) is
\begin{equation*}
  \E[\statrv \mid \statrv \ge q_{1-\alpha}]
  \stackrel{(i)}{=}
  \inf_x \left\{ \frac{1}{1-\alpha} \E[\hinge{\statrv-x}] +
    x \right\}
  ~~ \mbox{where} ~
  q_{1 - \alpha} = \inf\{x \in \R : 1 - \alpha \le \P(\statrv \le x)\}.
\end{equation*}
Conditional Value-at-Risk is of interest in many
financial applications~\cite{RockafellarUr00, ShapiroDeRu09}.

For our second simulation experiment, we investigate three different
distributions: a mixture of normal distributions and two different mixtures of
heavy-tailed distributions.  For our normal experiments, we draw $\statrv$
from an equal weight mixture of normal distributions with means
$\mu \in \{-6, -4, -2, 0, 2, 4, 6\}$ and variances
$\sigma^2 \in\{2, 4, 6, 8, 10, 12, 14\}$, respectively. In keeping with our
financial motivation, we interpret $\mu$ as negative returns, where $\sigma^2$
increases as $\mu$ increases, reminiscent of the oft-observed tendency in bear
markets (the leverage effect)~\cite{Black76, Christie82}.  For the
heavy-tailed experiments, we take $\statrv = \mu + Z$ for $Z$ symmetric with
$\P(|Z| \ge t) \propto \min\{1, t^{-a}\}$, and we we take an equal weight
mixture of distributions centered at $\mu \in \{-6, -4, -2, 0, 2, 4, 6\}$.

Our optimization problem is thus to minimize the loss
$\loss(x; \statval) = \frac{1}{1 - \alpha} \hinge{\statval - x} + x$, and we
compare the performance of the generalized empirical likelihood confidence
regions we describe and normal approximations.  For all three mixture
distributions, the cumulative distribution function is increasing, so there is
a unique population minimizer.  To approximate the population optimal value,
we take $n = 1,\!000,\!000$ to obtain a close sample-based approximation to
the CVaR $ \E_{P_0}[\statval \mid \statval \ge q_{1-\alpha}]$. Although the
feasible region $\mathcal{X} = \R$ is not compact, we compute the generalized
empirical likelihood interval~\eqref{eqn:upper-lower-def} and compare coverage
rates for confidence regions that asymptotically have the nominal level
$95\%$. In Table~\ref{table:coverages}, we see that the empirical likelihood
coverage rates are generally smaller than the normal coverage rates, which is
evidently (see Figure~\ref{fig:cvar}) a consequence of still remaining
negative bias (optimism) in the robust estimator~\eqref{eqn:upper}.  In
addition, the true coverage rate converges to the nominal level (95\%) more
slowly for heavy-tailed data (with $\beta \in \{3, 5\}$).

\subsection{Multi-item Newsvendor}

Our final simulation investigates the performance of the generalized empirical
likelihood integral~\eqref{eqn:upper-lower-def} for the multi-item newsvendor
problem. In this problem, the random variables $\statrv \in \R^d$ denote
demands for items $j = 1, \ldots, d$, and for each item $j$, there
is a backorder cost $b_j$ per unit and inventory cost $h_j$ per unit.
For a given allocation $x \in \R^d$ of items, then, the loss upon receiving
demand $\statval$ is $\loss(x; \statval)
= b^\top \hinge{x - \statval} + h^\top \hinge{\statval - x}$,
where $\hinge{\cdot}$ denotes the elementwise positive-part of its argument.

For this experiment, we take $d = 20$ and set
$\xdomain = \{ x \in \R^d : \lone{x} \le 10 \}$, letting
$\statrv \simiid \normal(0, \Sigma)$ (there may be negative demand), where
$\Sigma$ is again standard Wishart distributed with $d$ degrees of freedom.
We choose $b, h$ to have i.i.d.\ entries distributed as
$\expdist(\frac{1}{10})$.  For each individual simulation, we approximate the
population optimum using a sample average approximation based on a sample of
size $n = 10^5$.  As Table~\ref{table:coverages} shows, the proportion of
simulations in which $[\what{l}_n, u_n]$ covers the true optimal value is
still lower than the nominal 95\%, though it is less pronounced than other
cases. Figure~\ref{fig:news} shows average confidence intervals for the
optimal value for both generalized empirical likelihood-based and normal-based
confidence sets.


%% file: generalization.tex
\section{General Results}
\label{section:general}

In this section, we abstract away from the stochastic optimization setting
that motivates us. By leveraging empirical process theory, we give general
results that apply to suitably smooth functionals (Hadamard differentiable)
and classes of functions $\{\loss(x; \cdot): x \in \mathcal{X} \}$ for which
a uniform central limit theorem holds ($P_0$-Donsker). Our subsequent
development implies the results presented in previous sections as
corollaries. We begin by showing results for i.i.d.\ sequences and defer
extensions to dependent sequences to Section~\ref{section:ergodic}.  Let
$Z_1, \ldots, Z_n$ be independent random vectors with common distribution
$P_0$. Let $\mathcal{P}$ be the set of probability distributions on
$\statdomain$ and let $T: \mathcal{P} \to \R$ be a functional of interest.

First, we show a general version of the uniform asymptotic
expansion~\eqref{eqn:real-asymptotic-expansion-stoch-opt} that applies to
$P_0$-Donsker classes in
Section~\ref{subsection:uniform-asymptotic-expansion}. In
Section~\ref{subsection:el-smooth} we give a generalized empirical likelihood
theory for Hadamard differentiable functionals $T(P)$, which in particular
applies to $T_{\rm opt}(P) = \inf_{x \in \mathcal{X}} \E_P[\obj]$ (cf.\
Theorem~\ref{theorem:el-stoch-opt}). The general treatment for Hadamard
differentiable functionals is necessary as Frech\'{e}t differentiability is
too stringent for studying constrained stochastic
optimization~\cite{Shapiro91}. Finally, we present extensions of the above
results to (quickly-mixing) dependent sequences in
Section~\ref{section:ergodic}.

\subsection{Uniform Asymptotic Expansion}
\label{subsection:uniform-asymptotic-expansion}

A more general story requires some background on empirical
processes, which we now briefly summarize (see \citet{VanDerVaartWe96} for a
full treatment).  Let $P_0$ be a fixed probability distribution on the
measurable space $(\statdomain, \mathcal{A})$, and recall the space $L^2(P_0)$
of functions square integrable with respect to $P_0$, where we equip functions
with the $L^2(P_0)$ norm $\ltwop{h}{P_0} = \E_{P_0}[h(\statrv)^2]^\half$. For
any signed measure $\mu$ on $\statdomain$ and $h : \statdomain \to \R$, we use
the functional shorthand $\mu h \defeq \int h(\statval) d\mu(\statval)$ so
that for any probability measure we have $Ph = \E_P[h(\statrv)]$.  Now, for a
set $\hclass \subset L^2(P_0)$, let $\dualbdd$ be the space of bounded linear
functionals on $\hclass$ equipped with the uniform norm
$\hnorm{L_1 - L_2} = \sup_{h \in \hclass} |L_1h - L_2h|$ for
$L_1, L_2 \in \dualbdd$.  To avoid measurability issues, we use outer
probability and expectation with the corresponding convergence notions as
necessary~\cite[e.g.][Section 1.2]{VanDerVaartWe96}. We then have the
following definition~\cite[Eq.~(2.1.1)]{VanDerVaartWe96} that describes
sets of functions on which the central limit theorem holds uniformly.
\begin{definition}
  \label{definition:donsker}
  A class of functions $\hclass$ is \emph{$P_0$-Donsker} if
  $\sqrt{n} (\what{P}_n - P_0) \cd G$ in the space $\dualbdd$,
  where $G$ is a tight Borel
  measurable element of $\dualbdd$, and
  $\what{P}_n$ is the empirical distribution
  of $\statrv_i \simiid P_0$.
\end{definition}
\noindent
In Definition~\ref{definition:donsker}, the measures $\emp$, $P_0$ are
considered as elements in $\dualbdd$ with $\emp f = \E_{\emp}f$,
$P_0 f = \E_{P_0} f$ for $f \in \hclass$.

With these preliminaries in place, we can state a general form of
Theorem~\ref{theorem:real-asymptotic-expansion-stoch-opt}.  We let $\mathcal{H}$
be a $P_0$-Donsker collection of functions $h : \statdomain \to \R$ with
$L^2$-integrable envelope, that is, $M_2 : \statdomain \to \R_+$ with
$h(\statval) \le M_2(\statval)$ for all $h \in \mathcal{H}$ with
$\E_{P_0}[M_2(\statrv)^2] < \infty$. Assume
the data $\statrv_i \simiid P_0$. Then we have
\begin{theorem}
  \label{theorem:real-asymptotic-expansion}
  Let the conditions of the preceding paragraph hold. Then
  \begin{equation*}
    \sup_{P : \fdivs{P}{\emp} \le \frac{\tol}{n}}
    \E_P[h(\statrv)]
    = \E_{\emp}[h(\statrv)]
    + \sqrt{\frac{\tol}{n} \var_{\emp}(h(\statrv))}
    + \varepsilon_n(h),
  \end{equation*}
  where $\sup_{h \in \mathcal{H}} \sqrt{n} |\varepsilon_n(h)| \cpstar 0$.
\end{theorem}
\noindent
See Appendix~\ref{section:proof-of-uniform-convergence}, in particular
Appendix~\ref{appendix:uniform-expansion-proof}, for the proof.

Theorem~\ref{theorem:real-asymptotic-expansion} is useful,
and in particular, we can derive
Theorem~\ref{theorem:real-asymptotic-expansion-stoch-opt} as a corollary:

\begin{example}[Functions Lipschitz in $x$]
  \label{example:lipschitz}
  Suppose that for each $\statval \in \statdomain$, the function
  $x \mapsto \loss(x; \statval)$ is $L(\statval)$-Lipschitz, where
  $\E[L(\statval)^2] < \infty$. If in addition the set $\xdomain$ is compact,
  then functions $\mathcal{H} \defeq \{\loss(x; \cdot)\}_{x \in \xdomain}$
  satisfy all the conditions of
  Theorem~\ref{theorem:real-asymptotic-expansion}. (See
  also \cite[Chs.\ 2.7.4 and 3.2]{VanDerVaartWe96}.)
\end{example}



\subsection{Hadamard Differentiable Functionals}
\label{subsection:el-smooth}

\newcommand{\measures}{\mathcal{M}}

In this section, we present an analogue of the asymptotic calibration in
Proposition~\ref{proposition:conf} for smooth functionals of probability
distributions, which when specialized to the optimization context yield the
results in Section~\ref{section:el-stoch-opt}.  Let $(\statdomain,
\mathcal{A})$ be a measurable space, and $\hclass$ be a collection of
functions $h : \statdomain \to \R$, where we assume that $\hclass$ is
$P_0$-Donsker with envelope $M_2 \in L^2(P_0)$
(Definition~\ref{definition:donsker}).  Let $\mathcal{P}$ be the space of
probability measures on $(\statdomain, \mathcal{A})$ bounded with respect to
the supremum norm $\norm{\cdot}_{\mathcal{H}}$ where we view measures as
functionals on $\mathcal{H}$. Then, for $T : \mathcal{P} \to \R$, the
following definition captures a form of differentiability sufficient for
applying the delta method to show that $T$ is asymptotically
normal~\cite[Chapter 3.9]{VanDerVaartWe96}. In the definition, we let
$\measures$ denote the space of signed measures on $\statdomain$ bounded
with respect to $\norm{\cdot}_{\mathcal{H}}$, noting that $\measures \subset
\dualbdd$ via the mapping $\mu h = \int h(\statval) d\mu(\statval)$.
\begin{definition}
  \label{definition:hadamard-directional}
  The functional $T: \mathcal{P} \to \R$ is \emph{Hadamard directionally
    differentiable at $P \in \mathcal{P}$ tangentially to
    $B \subset \measures$} if for all $H \in B$, there exists $dT_P(H) \in \R$
  such that for all convergent sequences $t_n \to 0$ and $H_n \to H$ in
  $\dualbdd$ (i.e. $\hnorm{H_n - H} \to 0$) for which
  $P + t_n H_n \in \mathcal{P}$, and
  \begin{equation*}
    \frac{T(P + t_n H_n) - T(P)}{t_n} \to dT_P(H)
    ~~ \mbox{as~} n \to \infty.
  \end{equation*}
  Equivalently, \emph{$T$ is Hadamard directionally differentiable at $P \in
  \mathcal{P}$ tangentially to $B \subset \measures$} if for every compact $K
  \subset B$,
  \begin{equation}
    \label{eqn:compact-directional-hadamard}
    \lim_{t \to 0}
    \sup_{H \in K, P + t H\in \mathcal{P}} \left| \frac{T(P + t H) - T(P)}{t}
    - dT_P(H) \right| = 0.
  \end{equation}
  Moreover, $T: \mathcal{P} \to \R$ is \emph{Hadamard differentiable at
    $P \in \mathcal{P}$ tangentially to $B \subset \dualbdd$} if
  $dT_P : B \to \R$ is linear and continuous on $B$.
\end{definition}

By restricting ourselves very slightly to a nicer class of Hadamard
differentiable functionals, we may present a result on asymptotically
pivotal confidence sets provided by $f$-divergences. To that end, we
say that $T : \mathcal{P} \to \R$ has \emph{influence function}
$T^{(1)} : \statdomain \times \mathcal{P} \to \R$ if
\begin{equation}
  \label{eqn:grad}
  dT_P(Q - P) = \int_\statdomain T^{(1)}(\statval; P) d(Q - P)(\statval)
\end{equation}
and $T^{(1)}$ satisfies $\E_P[T^{(1)}(\statrv; P)] = 0$.\footnote{A
  sufficient condition for $T^{(1)}(\cdot; P)$ to exist is that $T$ be
  Hadamard differentiable at $P$ tangentially to any set $B$ including the
  measures $\onevec_\statval - P$ for each $\statval \in P$: indeed, let
  $H_\statval \defeq \onevec_\statval - P$, then the $\int H_\statval
  dP(\statval) = 0$, and the linearity of $dT_P : B \to \R$ guarantees that
  $\int dT_P(H_\statval) dP(\statval) = \int dT_P(\onevec_\statval - P)
  dP(\statval) = dT_P(P - P) = 0$, and we define $T^{(1)}(\statval; P) =
  dT_P(\onevec_\statval - P)$.}
If we let $B = B(\hclass, P) \subset \dualbdd$ be the set of linear
functionals on $\mathcal{H}$ that are $\ltwop{\cdot}{P}$-uniformly
continuous and bounded, then this is sufficient for the existence of the
canonical derivative $T^{(1)}$, by the Riesz Representation Theorem
for $L^2$ spaces (see \cite[Chapter 25.5]{VanDerVaart98}
or \cite[Chapter 18]{Kosorok08}).




We now extend Proposition~\ref{proposition:conf} to Hadamard differentiable
functionals $T : \mathcal{P} \to \R$. \citet{Owen88} shows a similar result
for empirical likelihood (i.e.\ with $f(t) = -2\log t + 2t - 2$) for the
smaller class of Frech\'et differentiable functionals. Bertail et
al.~\cite{Bertail06, BertailGaHa14} also claim a similar result under certain
uniform entropy conditions, but their proofs are incomplete.\footnote{Their
  proofs~\cite[pg.\ 308]{Bertail06} show that confidence sets converge to one
  another in Hausdorff distance, which is not sufficient for their claim. The
  sets $A_n \defeq \{v / n : v \in \Z^d\}$ and $B = \R^d$ have Hausdorff
  distance $\frac{1}{2n}$, but for any random variable $Z$ with Lebesgue
  density, we certainly have $\P(Z \in A_n) = 0$ while $\P(Z \in B) = 1$.}
Recall that $\mathcal{M}$ is the (vector) space of signed measures in
$\dualbdd$.

\begin{theorem}
  \label{theorem:el-smooth}
  Let Assumption~\ref{assumption:fdiv} hold and let $\mathcal{H}$ be a
  $P_0$-Donsker class of functions with an $L^2$-envelope $M$.  Let
  $\statval_i \simiid P_0$ and let $B \subset \mathcal{M}$ be such that $G$
  takes values in $B$ where $G$ is the limit $\sqrt{n}(\emp - P_0) \cd G$ in
  $\dualbdd$ given in Definition~\ref{definition:donsker}. Assume that
  $T: \mathcal{P} \subset \mathcal{M} \to \R$ is Hadamard differentiable at
  $P_0$ tangentially to $B$ with infludence function $T^{(1)}(\cdot; P_0)$
  and that $dT_P$ is defined and continuous on the
  whole of $\mathcal{M}$. If $0 < \var(T^{(1)}(\statrv; P_0)) < \infty$,
  then
  \begin{equation}
    \label{eqn:el}
    \lim_{n \to \infty}
    \P\left(T(P_0) \in \left\{T(P) : \fdiv{P}{P_n} \le \frac{\rho}{n}
      \right\}\right)
    = \P\left( \chi_1^2 \le \tol \right),
  \end{equation}
\end{theorem}
\noindent
We use Theorem~\ref{theorem:real-asymptotic-expansion} to show the result in
Appendix~\ref{appendix:el-smooth-proof}.



\subsection{Extensions to Dependent Sequences}
\label{section:ergodic}

In this subsection, we show an extension of the empirical likelihood theory
for smooth functionals (Theorem~\ref{theorem:el-smooth}) to $\beta$-mixing
sequence of random variables.
%
%
Let $\{ \statrv \}_{i \in \Z}$ be a sequence of strictly stationary random
variables taking values in the Polish space $\statdomain$.  We follow the
approach of \citet{DoukhanMaRi95} to prove our results, giving bracketing
number conditions sufficient for our convergence guarantees (alternative
approaches are possible~\cite{NobelDe93, ArconesYu94, Yu94, Rio17}).


We first define bracketing numbers.
\begin{definition}
  \label{def:bracketing}
  Let $\norm{\cdot}$ be a (semi)norm on $\hclass$.  For functions
  $l, u : \statdomain \to \R$ with $l \le u$, the \emph{bracket} $[l, u]$ is
  the set of functions $h : \statdomain \to \R$ such that $l \le h \le u$, and
  $[l, u]$ is an \emph{$\epsilon$-bracket} if $\norm{l - u} \le \epsilon$.
  Brackets $\{[l_i, u_i]\}_{i = 1}^m$ \emph{cover} $\hclass$ if for all
  $h \in \hclass$, there exists $i$ such that $h \in [l_i, u_i]$. The
  \emph{bracketing number}
  $N_{[\hspace{1pt}]}(\epsilon, \hclass, \norm{\cdot})$ is the minimum number
  of $\epsilon$-brackets needed to cover $\hclass$.
\end{definition}

For i.i.d.\ sequences, if the bracketing integral is
finite,
\begin{equation*}
  \int_{0}^{\infty} \sqrt{\log N_{[\hspace{1pt}]}
    (\epsilon, \hclass, \ltwop{\cdot}{P_0})} \hspace{2pt} d\epsilon < \infty,
\end{equation*}
then $\hclass$ is $P_0$-Donsker~\cite[Theorem 2.5.6]{VanDerVaartWe96}. For
$\beta$-mixing sequences, a modification of the
$L^2(P_0)$-norm yields similar result.
To state the required bracketing condition in full, we first provide
requisite notation. For any $h \in L^1(P_0)$, we let
\begin{equation*}
  Q_h(u) = \inf \{t: \P(|h(\statrv_0)| > t) \le u\}.
\end{equation*}
be the quantile function of $|h(\statval_0)|$.
Define $\beta(t) \defeq \beta_{\floor{t}}$ where $\beta_n$ are the mixing
coefficients~\eqref{eqn:beta-mixing-coefficient}, and
define the norm
\begin{equation}
  \label{eqn:beta-norm}
  \norm{h}_{L^{2, \beta}(P_0)} = \sqrt{\int_0^1 \beta^{-1}(u) Q_h(u)^2 du},
\end{equation}
where $\beta^{-1}(u) = \inf\{t : \beta(t) \le u\}$.  When $\{ \statval_i
\}_{i \in \Z}$ are i.i.d., the $(2, \beta)$-norm $\norm{\cdot}_{L^{2,
    \beta}(P_0)}$ is the $L^2(P_0)$-norm as $\beta^{-1}(u) = 1$ for $u >
0$. Lastly, we let $\Gamma$ be the covariance function
\begin{equation}
  \label{eqn:covariance-ergodic}
  \Gamma (h_1, h_2) \defeq
  \sum_{i \in \Z} \cov (h_1(\statval_0), h_2(\statval_i)).
\end{equation}
We then have the following result, which extends bracketing
entropy conditions to $\beta$-mixing sequences. 
\begin{lemma}[{\citet[Theorem 1]{DoukhanMaRi95}}]
  \label{lemma:bracketing-for-beta-mixing}
  Let $\{ \statrv\}_{i \in \Z}$ be a strictly stationary sequence of random
  vectors taking values in the Polish space $\statdomain$
  with common distribution $P_0$ satisfying
  $\sum_{n=1}^{\infty} \beta_n < \infty$. Let $\mathcal{H}$ be a class of
  functions $h: \statdomain \to \R$ with envelope $M(\cdot)$ such
  that $\norm{M}_{L^{2, \beta}(P_0)} < \infty$.
  If
  \begin{equation*}
    \int_0^1 \sqrt{ \log N_{[\hspace{1pt}]}
      (\epsilon, \hclass, \norm{\cdot}_{L^{2, \beta}(P_0)})} \hspace{2pt} d\epsilon
    < \infty,
  \end{equation*}
  then the series $\sum_i \cov (h(\statrv_0), h(\statrv_i))$ is
  absolutely convergent to $\Gamma(h, h) < \infty$ uniformly in
  $h$, and
  \begin{equation*}
    \sqrt{n} (\emp - P_0) \cd G ~~~\mbox{in}~~~\dualbdd
  \end{equation*}
  where $G$ is a Gaussian process with covariance function $\Gamma$ and
  almost surely uniformly continuous sample paths.
\end{lemma}
\noindent The discussion following~\cite[Theorem 1]{DoukhanMaRi95} provides
connections between $\norm{\cdot}_{L^{2, \beta}(P_0)}$ and other norms, as
well as sufficient conditions for
Lemma~\ref{lemma:bracketing-for-beta-mixing} to hold.  For example, if the
bracketing integral with respect to the norm $\norm{\cdot}_{L^{2r}(P_0)}$ is
finite with $\sum_{n \ge 1} n^{\frac{1}{r-1}} \beta_n < \infty$, the
conditions of Lemma~\ref{lemma:bracketing-for-beta-mixing} are satisfied.

We now give an extension of Theorem~\ref{theorem:el-smooth} for dependent
sequences. Recall that $\mathcal{M}$ is the (vector) space of signed measures
in $\dualbdd$. Let $B \subset \mathcal{M}$ be such that $G$ takes values in
$B$.
\begin{theorem}
  \label{theorem:el-smooth-ergodic}
  Let Assumption~\ref{assumption:fdiv} and the hypotheses of
  Lemma~\ref{lemma:bracketing-for-beta-mixing} hold. Let $B \subset
  \mathcal{M}$ be such that $G$ takes values in $B$, where $\sqrt{n}(\emp -
  P_0) \cd G$ in $\dualbdd$ as in
  Lemma~\ref{lemma:bracketing-for-beta-mixing}. Assume that $T: \mathcal{P}
  \subset \mathcal{M} \to \R$ is Hadamard differentiable at $P_0$
  tangentially to $B$ with influence function $T^{(1)}(\cdot; P_0)$ as
  (Eq.~\eqref{eqn:grad}) and that $dT_P$ is defined and continuous on the
  whole of $\mathcal{M}$. If $0 < \var(T^{(1)}(\statrv; P_0)) < \infty$,
  then
  \begin{equation}
    \label{eqn:el-ergodic}
    \lim_{n \to \infty}
    \P\left(T(P_0) \in \left\{T(P) : \fdiv{P}{P_n} \le \frac{\rho}{n}
      \right\}\right)
    = \P\left( \chi^2_1 \le
      \frac{\tol \var_{P_{\statval}}
          T^{(1)}(\statval; P_0)}{\Gamma(T^{(1)}, T^{(1)})} \right).
  \end{equation}
\end{theorem}
\noindent See Section~\ref{sec:proof-el-smooth-ergodic} for the proof. We show
in Section~\ref{sec:proof-el-stoch-opt-ergodic} that
Theorem~\ref{theorem:el-stoch-opt-ergodic} follows from
Theorem~\ref{theorem:el-smooth-ergodic}.


%% file: conclusion.tex
\section{Conclusion}
\label{section:conc}

We have extended generalized empirical likelihood theory in a number of
directions, showing how it provides inferential guarantees for stochastic
optimization problems. The upper confidence bound~\eqref{eqn:upper} is a
natural robust optimization problem~\cite{Ben-TalGhNe09,Ben-TalHeWaMeRe13},
and our results show that this robust formulation gives exact asymptotic
coverage. The robust formulation implements a type of regularization by
variance, while maintaining convexity and risk coherence
(Theorem~\ref{theorem:real-asymptotic-expansion}). This
variance expansion explains the coverage properties of (generalized) empirical
likelihood, and we believe it is likely to be effective in a number of
optimization problems~\cite{DuchiNa16}.

There are a number of interesting topics for further research, and we list a
few of them. On the statistical and inferential side, the uniqueness
conditions imposed in Theorem~\ref{theorem:el-stoch-opt} are stringent, so it
is of interest to develop procedures that are (asymptotically) adaptive to the
size of the solution set $S_{P_0}\opt$ without being too conservative; this is
likely to be challenging, as we no longer have normality of the asymptotic
distribution of solutions.  On the computational side, interior point
algorithms are often too expensive for large scale optimization problems
(i.e.\ when $n$ is very large)---just evaluating the objective or its gradient
requires time at least linear in the sample size. While there is a substantial
and developed literature on efficient methods for sample average approximation
and stochastic gradient methods~\cite{PolyakJu92, NemirovskiJuLaSh09,
  DuchiHaSi11, DefazioBaLa14, JohnsonZh13, Hazan16}, there are fewer
established and computationally efficient solution methods for minimax
problems of the form~\eqref{eqn:upper} (though see the papers
\cite{NemirovskiJuLaSh09, ClarksonHaWo12, Ben-TalHaKoMa15, ShalevWe16,
  NamkoongDu16} for work in this direction). Efficient solution methods need
to be developed to scale up robust optimization.

There are two ways of injecting robustness in the
formulation~\eqref{eqn:upper}: increasing $\tol$ and choosing a function $f$
defining the $f$-divergence $\fdiv{\cdot}{\cdot}$ that grows slowly in a
neighborhood of $1$ (recall the Cressie-Read family~\eqref{eqn:cressie-read}
and associated dual problems). We characterize a statistically principled way
of choosing $\tol$ to obtain calibrated confidence bounds, and we show that
all smooth $f$-divergences have the same asymptotic ($n \to \infty$) behavior
to first-order. We do not know, however, the extent to which different choices
of the divergence measure $f$ impact higher order or finite-sample behavior of
the estimators we study. While the literature on higher order corrections for
empirical likelihood offers some answers for inference problems regarding the
mean of a distribution~\cite{DiCiccioHaRo91, Baggerly98, Corcoran98, Bravo03,
  Bravo06}, the more complex settings arising in large-scale optimization
problems leave a number of open questions.





%% file: pointwise-expansion-proof-short.tex

\newcommand{\Usmall}{\mathcal{U}_{\rm sm}}
\newcommand{\Ubig}{\mathcal{U}_{\rm big}}

\section{Proof of Lemma~\ref{lemma:key-expansion}}
\label{appendix:expansion-proof}

We assume without loss of generality for both that $Z$ is mean-zero and that
$\var(Z) > 0$, as if $\var(Z) = 0$ then $s_n = 0$ and the lemma is trivial.
We prove the result by approximating the function $f$ with simpler
functions, which allows a fairly immediate proof.

The starting point of the proof of each of Lemma~\ref{lemma:key-expansion}
is the following lemma, which gives sufficient conditions for the robust
expectation to be well approximated by the variance.
\begin{lemma}
  \label{lemma:sufficient-event-for-variance}
  Let $0 \le \epsilon < 1$ and define the event
  \begin{equation*}
    \mathcal{E}_n
    \defeq \left\{
    \max_{i \le n} \frac{|Z_i - \wb{Z}_n|}{\sqrt{n}}
    \le \epsilon s_n \sqrt{\frac{1 - C \epsilon}{\tol}} \right\}.
  \end{equation*}
  Then on $\mathcal{E}_n$,
  \begin{equation*}
    \E_{\emp}[Z]
    + \sqrt{\frac{\tol}{n} s_n^2}
    \frac{1}{\sqrt{1 + C \epsilon}}
    \le \sup_{P : \fdivs{P}{\emp} \le \frac{\tol}{n}}
    \E_P[Z]
    \le 
    \E_{\emp}[Z]
    + \sqrt{\frac{\tol}{n} s_n^2}
    \frac{1}{\sqrt{1 - C \epsilon}}.
  \end{equation*}
\end{lemma}
\noindent
This result gives a nearly immediate proof of Lemma~\ref{lemma:key-expansion},
as we require showing only that $\mathcal{E}_n$ holds eventually (i.e.\ for
all large enough $n$). We defer its proof to
Section~\ref{sec:proof-sufficient-event-for-variance}.

To that end, we state the following result, due essentially to
Owen~\cite[Lemma 3]{Owen90}.
\begin{lemma}
  \label{lemma:max-goes-to-zero}
  Let $Z_i$ be (potentially dependent) identically distributed
  random variables with $\E[|Z_1|^k] < \infty$ for some $k > 0$. Then
  for all $\epsilon > 0$,
  $\P(|Z_n| \ge \epsilon n^{1/k} ~ \mbox{i.o.}) = 0$ and
  $\max_{i \le n} |Z_i| / n^{1/k} \cas 0$.
\end{lemma}
\ifdefined\usemorstyle
\proof{Proof.}
\else
\begin{proof}
  \fi
  A standard change of variables gives
  $\E[|Z_1|^k] = \int_0^\infty \P(|Z_1|^k \ge t) dt
  \gtrsim \sum_{n = 1}^\infty \P(|Z_n|^k \ge n)$. Thus for any
  $\epsilon > 0$ we obtain
  \begin{equation*}
    \sum_{n = 1}^\infty \P(|Z_n|^k \ge \epsilon n)
    \lesssim \frac{1}{\epsilon} \E[|Z_1|^k] < \infty,
  \end{equation*}
  and the Borel-Cantelli lemma gives the result.
\ifdefined\usemorstyle
\endproof
\else
\end{proof}
\fi

Birkhoff's Ergodic Theorem and
that the sequence $\{Z_n\}$ is strictly stationary and ergodic
with $\E[Z_1^2] < \infty$
implies that
\begin{equation*}
  \wb{Z}_n = \frac{1}{n} \sum_{i=1}^n Z_i \cas \E[Z_1]
  ~~ \mbox{and} ~~
  \wb{Z^2_n} = \frac{1}{n} \sum_{i = 1}^n Z_i^2 \cas \E[Z_1^2].
\end{equation*}
Thus we have that $s_n^2 = \wb{Z^2_n} - \wb{Z}_n^2
\cas \var(Z) > 0$ and by Lemma~\ref{lemma:max-goes-to-zero},
the event $\mathcal{E}_n$ holds eventually.
Lemma~\ref{lemma:sufficient-event-for-variance} thus gives
Lemma~\ref{lemma:key-expansion}.

\subsection{Proof of Lemma~\ref{lemma:sufficient-event-for-variance}}
\label{sec:proof-sufficient-event-for-variance}

We require a few auxiliary functions before continuing with the arguments.
First, for $\epsilon > 0$ define the Huber function
\begin{equation*}
  h_\epsilon(t) = \inf_y \left\{
  \half (t - y)^2 + \epsilon |y| \right\}
  = \begin{cases} \half t^2 & \mbox{if~} |t| \le \epsilon \\
    \epsilon |t| - \half \epsilon^2 & \mbox{if~} |t| > \epsilon.
  \end{cases}
\end{equation*}
Now, because the function $f$ is $\mathcal{C}^3$ in a neighborhood of $1$ with
$f''(1) = 2$, there exist constants $0 < c, C < \infty$, depending only on
$f$, such that
\begin{equation}
  \label{eqn:simple-f-inequalities}
  2 (1 - C \epsilon) h_\epsilon(t) \le
  f(t + 1)
  \le (1 + C \epsilon) t^2 + \bigindic{[-\epsilon, \epsilon]}(t)
  ~~ \mbox{for~all~} t \in \R,
  ~ \mbox{and} ~ |\epsilon| \le c,
\end{equation}
where the first inequality follows because $t \mapsto f'(t)$ is
non-decreasing and $f'(1) = 0$ and $f''(1) = 2$, and the second similarly
because $f$ is $\mathcal{C}^3$ near $1$.

With the upper and lower bounds~\eqref{eqn:simple-f-inequalities} in place,
let us rewrite the supremum problem slightly. For $\epsilon < 1$, define the
sets
\begin{align}
  \nonumber
  \Usmall & \defeq \left\{u \in \R^n
  \mid \ones^T u = 0, \linf{n u} \le \epsilon,
  (1 + C \epsilon) \ltwo{n u}^2 \le \tol \right\} \subset \ldots \\
  \mathcal{U} & \defeq \left\{u \in \R^n \mid
  \ones^T u = 0, u \ge -1/n,
  \sum_{i = 1}^n f(nu_i + 1) \le \tol \right\} \subset \ldots
  \label{eqn:def-u-sets}
  \\
  \Ubig & \defeq \left\{u \in \R^n
  \mid \ones^T u = 0, \sum_{i = 1}^n h_\epsilon(n u_i) \le \frac{\tol}{
    2(1 - C \epsilon)} \right\}.
  \nonumber
\end{align}
Then for any vector $z \in \R^n$, by inspection (replacing
$p \in \R^n_+$ with $\ones^T p = 1$ with $u \in \R^n$ with $\ones^T u = 0$
and $u \ge -(1/n)$), we have
\begin{equation}
  \sup_{u \in \Usmall} u^T z
  \le \sup_p \left\{p^T z \mid \fdivs{p}{(1/n) \ones} \le \tol/n\right\}
  - \frac{1}{n} \ones^T z
  = \sup_{u \in \mathcal{U}} u^T z
  \le \sup_{u \in \Ubig} u^T z.
  \label{eqn:upper-lower-p-stuff}
\end{equation}
To show the lemma, then, it suffices to lower bound
$\sup_{u \in \Usmall} u^T z$ and upper bound $\sup_{u \in \Ubig} u^T z$.

To that end, the next two lemmas control both the upper and lower bounds in
expression~\eqref{eqn:upper-lower-p-stuff}. In the lemmas, we let
$\wb{z^2_n} = \frac{1}{n} \sum_{i = 1}^n z_i^2$ and $\wb{z}_n = \frac{1}{n}
\sum_{i = 1}^n z_i$.
\begin{lemma}
  \label{lemma:get-lower-bound-correct}
  Let $s_n(z)^2 = \wb{z^2_n} - \wb{z}_n^2$.
  If $\linf{z - \wb{z}_n} / \sqrt{n} \le \epsilon s_n(z)
  \sqrt{(1 + C \epsilon) / \tol}$, then
  \begin{equation*}
    \sup_{u \in \Usmall} u^T z
    = \sqrt{\frac{\tol}{n} s_n(z)^2} \frac{1}{\sqrt{1 + C \epsilon}}.
  \end{equation*}
\end{lemma}
\ifdefined\usemorstyle
\proof{Proof.}
\else
\begin{proof}
  \fi
  We can without loss of generality replace $z$ with $z - \wb{z}_n \ones$ in
  the supremum, as $\ones^T u = 0$ so $u^T z = u^T(z - \wb{z}_n \ones)$, and
  so we simply assume that $\ones^T z = 0$ and thus
  $\ltwo{z} = \sqrt{n} s_n(z)$. By the Cauchy-Schwarz
  inequality, $\sup_{u \in \Usmall} u^T z
  \le \sqrt{\tol} \ltwo{z} / (n \sqrt{1 + C \epsilon})$.
  We claim that under the conditions of the lemma, it is achieved. Indeed,
  let
  \begin{equation*}
    u = \frac{\sqrt{\tol}}{n \sqrt{1 + C \epsilon} \ltwo{z}} z
    = \frac{\sqrt{\tol}}{n^{3/2} \sqrt{(1 + C \epsilon) \wb{z^2_n}}} z,
  \end{equation*}
  so $\ltwo{nu} = \tol$ and $u^T z = \tol \ltwo{z} / (n \sqrt{1 + C
    \epsilon})$.  Then $u$ satisfies $u^T \ones = 0$, $\ltwo{u}^2 \le \tol /
  (n^2 (1 + C \epsilon))$, and because $\linf{z} / \sqrt{n} \le \epsilon
  \sqrt{1 + C \epsilon} s_n(z) / \sqrt{\tol}$ by assumption, we have
  $u \in \Usmall$, which gives the result.
\ifdefined\usemorstyle
\endproof
\else
\end{proof}
\fi

\begin{lemma}
  \label{lemma:get-upper-bound-correct}
  Let $s_n(z)^2 = \wb{z^2_n} - \wb{z}_n^2$. If $\linf{z - \wb{z}_n} /
  \sqrt{n} \le \epsilon s_n(z) \sqrt{(1 - C \epsilon) / \tol}$, then
  \begin{equation*}
    \sup_{u \in \Ubig} u^T z
    \le \sqrt{\frac{\tol}{n} s_n(z)^2}
    \frac{1}{\sqrt{1 - C \epsilon}}.
  \end{equation*}
\end{lemma}
\ifdefined\usemorstyle
\proof{Proof.}
\else
\begin{proof}
  \fi
  We have $u^T z = u^T(z - \ones \wb{z}_n)$ for $u^T \ones = 0$.
  Thus, we always have upper bound that
  \begin{equation*}
    \sup_{u \in \Ubig} u^T z
    \le \sup_{u \in \R^n}
    \left\{u^T (z - \ones \wb{z}_n)
    \mid \sum_{i = 1}^n h_\epsilon(nu_i) \le \frac{\tol}{
      2(1 - C \epsilon)} \right\}.
  \end{equation*}
  Let us assume w.l.o.g.\ (as in the proof of
  Lemma~\ref{lemma:get-lower-bound-correct}) that $\wb{z}_n = 0$ so that
  $\ltwo{z} = \sqrt{n} s_n(z)$. Introducing multiplier $\lambda \ge 0$, the
  Lagrangian for the above maximization problem is
  \begin{equation*}
    L(u, \lambda)
    = u^T z - \lambda \sum_{i = 1}^n h_\epsilon(n u_i)
    + \lambda \frac{\tol}{2 (1 - C \epsilon)}.
  \end{equation*}
  Let us supremize over $u$.  In one dimension, we have
  \begin{align*}
    \sup_{u_i} \{u_i z_i - \lambda h_\epsilon(n u_i)\}
    & = \lambda \sup_{v_i} \sup_y \left\{v_i \frac{z_i}{\lambda n}
    - \half (v_i - y)^2 - \epsilon |y| \right\} \\
    & = \lambda \sup_y \left\{\frac{z_i^2}{2 \lambda^2 n^2}
    + y \frac{z_i}{\lambda n} - \epsilon |y|\right\}
    = \frac{z_i^2}{2 \lambda n^2}
    + \bigindic{[-\epsilon,\epsilon]}\left(\frac{z_i}{\lambda n}\right). 
  \end{align*}
  We obtain
  \begin{equation}
    \label{eqn:Ubig-lagrangian}
    \sup_{u \in \Ubig} u^T z
    \le \inf_{\lambda \ge 0} \sup_u L(u, \lambda)
    = \inf_\lambda
    \left\{\frac{\ltwo{z}^2}{2 \lambda n^2}
    + \frac{\tol}{2(1 - C \epsilon)} \lambda
    \mid \lambda \ge \frac{\linf{z}}{\epsilon n}\right\}.
  \end{equation}
  Now, by taking
  \begin{equation*}
    \lambda = \sqrt{\frac{1 - C \epsilon}{\tol}} \frac{\ltwo{z}}{n},
  \end{equation*}
  we see that under the conditions of the lemma, we have
  $\lambda \ge \linf{z} / (\epsilon n)$ and substituting into
  the Lagrangian dual~\eqref{eqn:Ubig-lagrangian}, the lemma
  follows.
\ifdefined\usemorstyle
\endproof
\else
\end{proof}
\fi

Combining Lemmas~\ref{lemma:get-lower-bound-correct}
and~\ref{lemma:get-upper-bound-correct} gives
Lemma~\ref{lemma:sufficient-event-for-variance}.

%% file: uniform-expansion-proof.tex
\newcommand{\dualbddz}{\mathcal{L}^\infty(\mathcal{H})}

\section{Uniform convergence results}
\label{section:proof-of-uniform-convergence}

In this section, we give the proofs of theorems related to uniform convergence
guarantees and the uniform variance expansions. The order of proofs is not
completely reflective of that we present in the main body of the paper, but we
order the proofs so that the dependencies among the results are linear. We
begin by collecting important technical definitions, results, and a few
preliminary lemmas. In Section~\ref{appendix:uniform-expansion-proof}, we then
provide the proof of Theorem~\ref{theorem:real-asymptotic-expansion}, after
which we prove Theorem~\ref{theorem:el-smooth} in
Section~\ref{appendix:el-smooth-proof}. Based on
Theorem~\ref{theorem:el-smooth}, we are then able to give a nearly immediate
proof (in Section~\ref{subsection:el-proof}) we of
Proposition~\ref{proposition:conf}.

\subsection{Preliminary results and definitions}
\label{sec:prelim-def}

We begin with several definitions and assorted standard lemmas important for
our results, focusing on results on convergence in distribution in general
metric spaces. See, for example, the first section of the book by
\citet{VanDerVaartWe96} for an overview.

\begin{definition}[Tightness]
  A random variable $X$ on a metric space $(\mathcal{X}, \mathsf{d})$ is \emph{tight} if
  for all $\epsilon > 0$, there exists a compact set $K_\epsilon$ such that
  $\P(X \in K_\epsilon) \ge 1 - \epsilon$. A sequence of random variables
  $X_n \in \mathcal{X}$ is \emph{asymptotically tight} if for every $\epsilon >
  0$ there exists a compact set $K$ such that
  \begin{equation*}
    \liminf_n P_*(X_n \in K^\delta) \ge 1 - \epsilon
    ~~ \mbox{for~all~} \delta > 0,
  \end{equation*}
  where $K^\delta = \{x \in \mathcal{X} : \dist(x, K) < \delta\}$ is the
  $\delta$-enlargement of $K$
  and $P_*$ denotes inner measure.
\end{definition}
\begin{lemma}[Prohorov's theorem~\cite{VanDerVaartWe96}, Theorem~1.3.9]
  \label{lemma:prohorov}
  Let $X_n \in \mathcal{X}$ be a sequence of random variables in
  the metric space $\mathcal{X}$. Then
  \begin{enumerate}
  \item If $X_n \cd X$ for some random variable $X$ where $X$ is tight,
    then $X_n$ is asymptotically tight and measurable.
  \item If $X_n$ is asymptotically tight, then there is a subsequence
    $n(m)$ such that $X_{n(m)} \cd X$ for some tight random variable $X$.
  \end{enumerate}
\end{lemma}
\noindent
Thus, to show that a sequence of random vectors converges in distribution,
one necessary step is to show that the sequence is tight. We now present two
technical lemmas on this for random vectors in $\dualbdd$. In each,
$\mathcal{H}$ is some set (generally a collection of functions in our
applications), and $\Omega_n$ is a sample space defined for each $n$. (In
our applications, we take $\Omega_n = \statdomain^n$.) We let $X_n(h) \in
\R$ denote the random realization of $X_n$ evaluated at $h \in \mathcal{H}$.
\begin{lemma}[Van der Vaart and Wellner~\cite{VanDerVaartWe96}, Theorem 1.5.4]
  \label{lemma:asymptotic-tight-marginal}
  Let $X_n: \Omega_n \to \dualbdd$. Then $X_n$ converges weakly to a tight
  limit if and only if $X_n$ is asymptotically tight and the marginals
  $(X_n(h_1), \ldots, X_n(h_k))$ converge weakly to a limit for every
  finite subset $\left\{ h_1,\ldots, h_k\right\}$ of $\mathcal{H}$. If $X_n$ is
  asymptotically tight and its marginals converge weakly to the marginals of
  $(X(h_1), \ldots, X(h_k))$ of $X$, then there is a version of $X$ with
  uniformly bounded sample paths and $X_n \cd X$.
\end{lemma}
Although the convergence in
distribution in outer probability does not require measurability of the
pre-limit quantities, the above lemma guarantees it.

\begin{lemma}[Van der Vaart and Wellner~\cite{VanDerVaartWe96},
    Theorem 1.5.7]
  \label{lemma:equi-continuity}
  A sequence of mappings
  $X_n: \Omega_n \to \dualbdd$ is asymptotically tight if and only if
  (i) $X_n(h)$ is asymptotically tight in $\R$ for all $h \in \mathcal{H}$,
  (ii)
  there exists a semi-metric $\norm{\cdot}$ on $\mathcal{H}$ such that
  $(\mathcal{H}, \norm{\cdot})$ is totally bounded, and (iii) $X_n$ is
  asymptotically uniformly equicontinuous in probability, \ie,~for
  every $\epsilon, \eta > 0$, there exists $\delta > 0$ such that
  $\limsup_{n\to\infty} 
  \P\left( \sup_{\norm{h- h'} < \delta} 
  \left| X_n(h) - X_n(h') \right| > \epsilon \right) < \eta$.
\end{lemma}

\subsection{Technical lemmas}

With these preliminary results stated, we provide two technical lemmas.
Recall the definition
\begin{equation}
  \label{eqn:robust-set-redefined}
  \mathcal{P}_{n, \tol} = \{P : \fdivs{P}{\emp} \le \frac{\tol}{n}\}
\end{equation}
The first, Lemma~\ref{lemma:shrink-to-emp}, shows that the vector $np$ is
close to the all-ones vector for all vectors $p \in \mathcal{P}_{n,
  \tol}$. The second (Lemma~\ref{lemma:tight}) gives conditions for tightness
of classes of functions $h : \statdomain \to \R$.
\begin{lemma}
  \label{lemma:shrink-to-emp}
  Let Assumption~\ref{assumption:fdiv} hold. Then
  \begin{equation*}
    \sqrt{\rho c_f} \le
    \sup_{n \in \N} \sup_{p\in\R^n} \Big\{ \ltwo{np - \onevec} :
    p^\top \onevec = 1,~p \ge 0, ~\sum_{i=1}^n f(np_i) \leq \tol \Big\}
    \le \sqrt{\rho C_f }
  \end{equation*}
  for some $C_f \ge c_f > 0$ depending only on $f$.
\end{lemma}
\ifdefined\usemorstyle
\proof{Proof.}
\else
\begin{proof}
  \fi
  By performing a Taylor expansion of $f$ around 1 for the point $np_i$ and
  using $f(1) = f'(1) = 0$, we obtain
  \begin{equation*}
    f(np_i) = \frac{1}{2} f^{''}(s_i) (np_i-1)^2
  \end{equation*}
  for some $s_i$ between $np_i$ and 1. As $f$ is convex with $f''(1) > 0$, it
  is strictly increasing on $\openright{1}{\infty}$. Thus there exists a
  unique $M > 1$ such that $f(M) = \tol$. If $f(0) = \infty$, there is
  similarly a unique $m \in (0, 1)$ such that $f(m) = \tol$ (if no such $m$
  exists, because $f(0) < \tol$, define $m = 0$).  Any
  $p \in \{p \in \R^n: p^\top \onevec = 1,~p\ge 0,~\sum_{i=1}^n f(np_i) \leq
  \tol \}$ must thus satisfy $np_i \in [m, M]$. Because $f$ is $C^2$ and
  strictly convex, $C_f^{-1} \defeq \inf_{s \in [m, M]} f^{''}(s)$ and
  $c_f^{-1} \defeq \sup_{s \in [m, M]} f^{''}(s)$ exists, are attained, and are
  strictly positive. Using the Taylor expansion of the $f$-divergence, we have
  $(np_i - 1)^2 = 2 f(np_i) / f''(s_i)$ for each $i$, and thus
  \begin{align*}
    \sum_{i=1}^n (np_i - 1)^2
    & = \sum_{i=1}^n \frac{2 f(np_i)}{f^{''}(s_i)}
      \le 2 C_f \sum_{i=1}^n f(np_i)
      \le 2 C_f \tol
  \end{align*}
  and similarly $\sum_{i=1}^n (np_i - 1)^2 \ge 2c_f \tol$.
  Taking the square root of each sides gives the lemma.
\ifdefined\usemorstyle
\endproof
\else
\end{proof}
\fi

\begin{lemma}
  \label{lemma:tight}
  Let $\mathcal{H}$ be $P_0$-Donsker with $L^2$-integrable envelope $M_2$,
  i.e.\ $|h(\statval)| \le M_2(\statval)$ for all $h \in \mathcal{H}$ with
  $\E_{P_0}[M_2^2(\statrv)] < \infty$. Then for any sequence
  $Q_n \in \mathcal{P}_{n, \tol}$, the mapping
  $\sqrt{n}(Q_n - P_0): \dualbdd \to \R$ is asymptotically tight.
\end{lemma}
\ifdefined\usemorstyle
\proof{Proof.}
\else
\begin{proof}
  \fi
  We use the charactization of asymptotic tightness in
  Lemma~\ref{lemma:equi-continuity}. With that in mind,
  consider an arbitrary sequence $Q_n \in \mathcal{P}_{n, \tol}$. We
  have
  \begin{align*}
    \P\left( \sup_{\norm{h - h'} < \delta} 
    \left| \sqrt{n} (Q_n - \emp) (h-h')\right| \ge \epsilon \right)
    & \stackrel{(a)}{\le} \P\left( \sup_{\norm{h -h'} < \delta} 
    \ltwo{nq - \onevec} \ltwop{h - h'}{\emp}
    \ge \epsilon \right) \\
    & \stackrel{(b)}{\le} \P\left( \sqrt{\frac{\tol}{\gamma_f}} 
    \sup_{\norm{h - h'} < \delta}  \ltwop{h - h'}{\emp}
    \ge \epsilon \right)
  \end{align*}
  where inequality $(a)$ follows from the Cauchy-Schwarz inequality
  and inequality $(b)$ follows from
  Lemma~\ref{lemma:shrink-to-emp}. Since $\mathcal{H}$ is $P_0$-Donsker, the
  last term goes to $0$ as $n \to \infty$ and $\delta \to 0$. Note that
  \begin{align*}
    \lefteqn{
      \limsup_{\delta \to 0} \limsup_{n\to\infty} \P\left( \sup_{\norm{h - h'} < \delta} 
      \left| \sqrt{n} (Q_n - P) (h-h')\right| \ge \epsilon \right)} \\
    & \le \limsup_{\delta, n}
    \left\{\P\left( \sup_{\norm{h - h'} < \delta} 
    \left| \sqrt{n} (Q_n - \emp) (h-h')\right| \ge \frac{\epsilon}{2} \right) 
    +
    \P\left( \sup_{\norm{h - h'} < \delta} 
    \left| \sqrt{n} (\emp - P) (h-h')\right| \ge \frac{\epsilon}{2} \right)
    \right\}.
  \end{align*}
  As $\sqrt{n}(\emp - P_0)$ is asymptotically tight in $\dualbdd$
  \cite[Theorem 1.5.4]{VanDerVaartWe96}, the second term vanishes by
  Lemma~\ref{lemma:equi-continuity}. Applying
  Lemma~\ref{lemma:equi-continuity} again, we conclude that
  $\sqrt{n}(Q_n - P_0)$ is asymptotically tight.
\ifdefined\usemorstyle
\endproof
\else
\end{proof}
\fi

\subsection{Proof of Theorem~\ref{theorem:real-asymptotic-expansion}}
\label{appendix:uniform-expansion-proof}


The proof of the theorem uses Lemma~\ref{lemma:key-expansion} and standard
tools of empirical process theory to make the expansion uniform.  Without
loss of generality, we assume that each $h \in \mathcal{H}$ mean-zero (as we may
replace $h(\statrv)$ with $h(\statrv) - \E[h(\statrv)]$).  We use the
standard characterization of asymptotic tightness given by
Lemma~\ref{lemma:asymptotic-tight-marginal}, so we show the finite
dimensional convergence to zero of our process. It is clear that there is
\emph{some} random function $\varepsilon_n$ such that
\begin{equation*}
  \sup_{P \in \mathcal{P}_{\tol, n}} \E_P[h(\statrv)]
  = \E_{\emp}[h(\statrv)]
  + \sqrt{\frac{\tol}{n} \var_{P_0}(Zh(\statrv))}
  + \varepsilon_n(h),
\end{equation*}
but we must establish its uniform convergence to zero at a rate
$o(n^{-\half})$.

To establish asymptotic tightness of the collection
$\{\sup_{P \in \mathcal{P}_{\tol,n}}
\E_P[h(\statrv)]\}_{h \in \mathcal{H}}$, first note that we have finite dimensional
marginal convergence. Indeed, we have $\sqrt{n} \varepsilon_n(h) \cp 0$
for all $h \in \mathcal{H}$ by Lemma~\ref{lemma:key-expansion},
and so for any finite $k$ and any $h_1,
\ldots, h_k \in \mathcal{H}$, $\sqrt{n}\left(\varepsilon_n(h_1), \ldots,
\varepsilon_n(h_k) \right) \cp 0$.  Further, by our Donsker assumption
on $\mathcal{H}$ we have that $\{h(\cdot)^2, h \in \mathcal{H}\}$ is a
Glivenko-Cantelli class~\cite[Lemma 2.10.14]{VanDerVaartWe96}, and
\begin{equation}
  \label{eqn:glivenko-cantelli-second-moment}
  \sup_{h \in \mathcal{H}} \left|\var_{\emp}(h(\statrv))
  - \var_{P_0}(h(\statrv)) \right| \cpstar 0.
\end{equation}
Now, we write the error term $\varepsilon_n$ as
\begin{align*}
  \sqrt{n} \varepsilon_n(h)
  & = \underbrace{\sqrt{n} \sup
    \left\{\E_P[h(\statrv)] - \E_{P_0}[h(\statrv)]
    \mid \fdivs{P}{\emp} \le \tol / n\right\}
  }_{(a)}  \\
  & \qquad
    - \underbrace{\sqrt{n} \, \left( \E_{\emp}[h(\statrv)]
    - \E_{P_0}[h(\statrv)]\right)}_{(b)}
    - \underbrace{\sqrt{\tol \var_{\emp} ( h(\statrv))}}_{(c)}.
\end{align*}
Then term $(a)$ is asymptotically tight (as a process on $h \in \mathcal {H}$) in
$\dualbddz$ by Lemma~\ref{lemma:tight}. The term $(b)$ is similarly tight
because $\mathcal{H}$ is $P_0$-Donsker by assumption, and term $(c)$ is
tight by the uniform Glivenko-Cantelli
result~\eqref{eqn:glivenko-cantelli-second-moment}. In particular,
$\sqrt{n}\varepsilon_n(\cdot)$ is an asymptotically tight sequence in
$\dualbddz$. As the finite dimensional distributions all converge to 0 in
probability, Lemma~\ref{lemma:asymptotic-tight-marginal} implies that
$\sqrt{n} \varepsilon_n \cd 0$ in $\dualbddz$ as desired. Of course,
convergence in distribution to a constant implies convergence in probability
to the constant.

\subsection{Proof of Theorem~\ref{theorem:el-smooth}}
\label{appendix:el-smooth-proof}
We first state a standard result that the delta method applies for
Hadamard differentiable functionals, as given by~\citet[Section
3.9]{VanDerVaartWe96}. In the lemma, the sets $\Omega_n$ denote the
implicit sample spaces defined for each $n$. For a proof,
see~\cite[Theorem 3.9.4]{VanDerVaartWe96}.
\begin{lemma}[Delta method]
  \label{lemma:delta}
  Let $T: \mathcal{P} \subset \mathcal{M} \to \R$ be Hadamard
  differentiable at $W$ tangentially to $B$ with $dT_Q$ linear and
  continuous on the whole of $\mathcal{M}$. Let $Q_n : \Omega_n \to \R$ be
  maps (treated as random elements of $\mathcal{M} \subset \dualbdd$) with
  $r_n(Q_n - Q) \cd Z$ in $\dualbdd$, where $r_n \to \infty$ and $Z$
  is a separable, Borel-measurable map. Then
  $r_n(T(Q_n) - T(Q)) - dT_Q\left(r_n(Q_n - Q)\right) \cpstar 0$.
\end{lemma}

For a probability measure $P$, define
$\kappa(P) \defeq T(P) - T(P_0) - \E_P[T^{(1)}(\statrv; P_0)]$. Since
$\mathcal{H}$ was assumed to be $P_0$-Donsker, we have
$\sqrt{n}(\emp - P_0) \cd G$ in $\dualbddz$. 
Recalling the canonical derivative $T^{(1)}$, we have from
Lemma~\ref{lemma:delta} that
\begin{equation}
  \label{eqn:delta-method}
  T(\emp) = T(P_0) + \E_{\emp}[T^{(1)}(\statval, P_0)] + \kappa(\emp)
\end{equation}
where $\kappa(\emp) = o_P(n^{-\half})$. Next, we show that this is true
uniformly over $\{ P: \fdivs{P}{\emp} \le \frac{\tol}{n} \}$.  We return to
prove the lemma in Section~\ref{subsection:uniform-delta-proof} for the proof.
\begin{lemma}
  \label{lemma:uniform-delta}
  Under the assumptions of Theorem~\ref{theorem:el-smooth}, for any
  $\epsilon > 0$
  \begin{equation}
    \label{eqn:little-o-remainder}
    \limsup_n \P\left(\sup_P \left\{|\kappa(P)| : \fdivs{P}{\emp} \le
        \frac{\tol}{n} \right\} \ge \frac{\epsilon}{\sqrt{n}}\right) = 0
  \end{equation}
  where
  $\kappa(P) \defeq T(P) - T(P_0) - \E_P[T^{(1)}(\statrv; P_0)]$.
\end{lemma}
We now see how the theorem is a direct consequence of
Theorem~\ref{theorem:real-asymptotic-expansion} and
Lemma~\ref{lemma:uniform-delta}.  Taking $\sup$ over
$\{P: \fdivs{P}{\emp} \le \frac{\tol}{n}\}$ in the definition of
$\kappa(\cdot)$, we have
\begin{align*}
  \left| \sup_{P: \fdivs{P}{\emp} \le \frac{\tol}{n}} T(P) - T(P_0) 
  - \sup_{P: \fdivs{P}{\emp} \le \frac{\tol}{n}} 
  \E_P[T^{(1)}(\statval; P_0)] \right|
  \le \sup_{P: \fdiv{P}{\emp} \le \frac{\tol}{n}} \left| \kappa(P) \right|.
\end{align*}
Now, multiply both sides by $\sqrt{n}$ and apply
Theorem~\ref{theorem:real-asymptotic-expansion} and
Lemma~\ref{lemma:uniform-delta} to obtain
\begin{align*}
  \left| \sqrt{n} \left(
  \sup_{P: \fdiv{P}{\emp} \le \frac{\tol}{n}} T(P) - T(P_0) \right)
  - \sqrt{n} \E_{\emp}[T^{(1)}(\statval;P_0)] 
  - \sqrt{\tol \var_{\emp}\hspace{2pt} T^{(1)}(\statval;P_0)} \right|
  = o_p(1).
\end{align*}
Since $\E_{P_0}[T^{(1)}(\statrv; P_0)] = 0$ by assumption, the central limit
theorem then implies that
\begin{equation*}
  \sqrt{n} \left(
  \sup_{P: \fdiv{P}{\emp} \le \frac{\tol}{n}} T(P) - T(P_0) \right)
  \cd \sqrt{\tol\var \hspace{2pt} T^{(1)}(\statval;P_0)} 
  + N\left(0, \var \hspace{2pt} T^{(1)}(\statval;P_0)\right).
\end{equation*}
Hence, we have
$\P\left( T(P_0) \le \sup_{P: \fdiv{P}{\emp} \le \frac{\tol}{n}}
  T(P)\right) \to P(N(0,1) \ge -\sqrt{\tol})$.
By an exactly symmetric argument on $-T(P_0)$, we similarly have
$\P\left(T(P_0) \ge \inf_{P: \fdiv{P}{\emp} \le \frac{\tol}{n}} T(P)\right) 
\to P(N(0,1) \le \sqrt{\tol})$. We conclude that
\begin{equation*}
  \P\left( T(P_0)
    \in \left\{ T(P): \fdiv{P}{\emp} \le \frac{\tol}{n}\right\} \right)
  \to P(\chi_1^2 \le \tol).
\end{equation*}

\subsubsection{Proof of Lemma~\ref{lemma:uniform-delta}}
\label{subsection:uniform-delta-proof}
Let
$\mathcal{P}_{n, \tol} \defeq \{ P: \fdivs{P}{\emp} \le \frac{\tol}{n}
\}$.
Recall that $\left\{X_n\right\} \subset \dualbdd$ is asymptotically
tight if for every $\epsilon > 0$, there exists a compact $K$ such
that
$\liminf_{n\to\infty} \P\left( X_n \in K^\delta \right) \ge
1-\epsilon$
for all $\delta > 0$ where
$K^\delta \defeq \left\{y \in \dualbdd: d(y, K) < \delta \right\}$
(\eg, \cite[Def 1.3.7]{VanDerVaartWe96}). 
Now, for an arbitrary $\delta > 0$, let $Q_n \in \mathcal{P}_{n, \tol}$
such that
$|\kappa(Q_n)| \ge (1-\delta) \sup_{Q\in\mathcal{P}_{n,\tol}} |\kappa(Q)|$.
Since the sequence $\sqrt{n}(Q_n - P_0)$ is asymptotically tight by
Lemma~\ref{lemma:tight}, every subsequence has a further subsequence
$n(m)$ such that $\sqrt{n(m)} (Q_{n(m)} - P_0) \cd X$ for some tight
and Borel-measurable map $X$. It then follows from
Lemma~\ref{lemma:delta} that $\sqrt{n(m)} \kappa_{n(m)}(Q_{n(m)}) \to 0$ as
$m \to \infty$. The desired result follows since
\begin{equation*}
  \P\left( (1-\epsilon) \sqrt{n} \sup_{Q\in\mathcal{P}_{n,\tol}} |\kappa_{n(m)}(Q)|
    \ge \epsilon \right) 
  \le \P\left( \sqrt{n} |\kappa_{n(m)}(Q_n)| \ge \epsilon \right) \to 0.
\end{equation*}

\subsection{Proof of Proposition~\ref{proposition:conf}}
\label{subsection:el-proof}


Let $Z \in \R^d$ be random vectors with covariance $\Sigma$, where
$\rank(\Sigma) = d_0$. From Theorem~\ref{theorem:el-smooth}, we have that
if we define
\begin{equation*}
  T_{s,n}(\lambda)  \defeq 
  s \sup_{P : \fdiv{P}{P_n} \le \tol/n}
  \left\{s \E_P[Z^T \lambda] \right\},
  ~~~ s \in \{-1, 1\},
\end{equation*}
then
\begin{equation*}
  \left[\begin{matrix}
      \sqrt{n} (T_{1,n}(\lambda) - \E_{P_0}[Z^T\lambda])
      \\
      \sqrt{n} (T_{-1,n}(\lambda) - \E_{P_0}[Z^T\lambda])
    \end{matrix}\right]
  = \left[\begin{matrix} \sqrt{n}
      (\E_{P_n}[Z]^T\lambda - \E_{P_0}[Z]^T \lambda)
      + \sqrt{\tol \lambda^T \Sigma \lambda} \\
      \sqrt{n}
      (\E_{P_n}[Z]^T\lambda - \E_{P_0}[Z]^T \lambda)
      - \sqrt{\tol \lambda^T \Sigma \lambda} 
    \end{matrix}\right]
  + o_P(1)
\end{equation*}
uniformly in $\lambda$ such that $\ltwo{\lambda} = 1$.
(This class of functions is trivially $P_0$-Donsker.)
The latter quantity converges (uniformly in $\lambda$) to
\begin{equation*}
  \left[\begin{matrix} \lambda^T W + \sqrt{\tol \lambda^T \Sigma \lambda} \\
      \lambda^T W - \sqrt{\tol \lambda^T \Sigma \lambda}
    \end{matrix}\right]
\end{equation*}
for $W \sim \normal(0, \Sigma)$ by the central limit theorem.
Now, we have that
\begin{equation*}
  \E_{P_0}[Z] \in \underbrace{\{\E_P[Z] : \fdiv{P}{P_n} \le \tol/n\}}_{
    \eqdef C_{\tol,n}}
\end{equation*}
if and only if
\begin{equation*}
  \inf_{\lambda : \ltwo{\lambda} \le 1}
  \{T_{1,n}(\lambda) - \E_{P_0}[Z^T\lambda]\} \le 0 ~ \mbox{and} ~
  \sup_{\lambda : \ltwo{\lambda} \le 1}
  \{T_{-1,n}(\lambda) - \E_{P_0}[Z^T\lambda]\} \ge 0
\end{equation*}
by convexity of the set $C_{\tol,n}$.
But of course, by convergence in distribution and the homogeneity of
$\lambda \mapsto \lambda^T W + \sqrt{\tol \lambda^T \Sigma \lambda}$, the
probabilities of this event converge to
\begin{equation*}
  \P\left(\inf_{\lambda} \{\lambda^T W + \sqrt{\tol \lambda^T \Sigma \lambda}
  \} \ge 0,
  \sup_{\lambda} \{\lambda^T W - \sqrt{\tol \lambda^T \Sigma \lambda}
  \} \le 0
  \right)
  = \P\left(\norm{W}_{\Sigma^\dag} \ge \sqrt{\tol}\right)
  = \P(\chi_{d_0}^2 \ge \tol)
\end{equation*}
by the continuous mapping theorem.


%% file: danskin-proof.tex
\section{Proofs of Statistical Inference for Stochastic Optimization}
\label{appendix:danskin-proof}

In this appendix, we collect the proofs of the results in
Sections~\ref{section:el-stoch-opt} and \ref{section:robust} on statistical
inference for the stochastic optimization problem~\eqref{eqn:pop}. We first
give a result explicitly guaranteeing smoothness of
$T_{\rm opt}(P) = \inf_{x\in\mathcal{X}} \E_{P}[\obj]$. The following variant
of Danskin's theorem~\cite{Danskin67} gives Hadamard differentiability of
$T_{\rm opt}$ tangentially to the space $B(\mathcal{H}, P_0) \subset \dualbdd$
of bounded linear functionals continuous w.r.t.\ $L^2(P_0)$ (which we may
identify with measures, following the discussion after
Definition~\ref{definition:hadamard-directional}). The proof of
Lemma~\ref{lemma:danskin}---which we include in
Appendix~\ref{section:proof-danskin-lemma} for completeness---essentially
follows that of~\citet{Romisch05}.
\begin{lemma}
  \label{lemma:danskin}
  Let Assumption~\ref{assumption:lsc} hold and assume
  $x \mapsto \loss(x; \statval)$ is continuous for $P_0$-almost all
  $\statval \in \statdomain$. Then the functional
  $T_{\rm opt} : \mathcal{P} \to \R$ defined by
  $T_{\rm opt}(P) = \inf_{x\in \xdomain} \E_{P}[\obj]$ is Hadamard
  directionally differentiable on $\mathcal{P}$ tangentially to
  $B(\mathcal{H}, P_0)$ with derivative
  \begin{equation*}
    dT_P(H) \defeq \inf_{x \in S_P\opt} \int \obj dH(\statrv)
  \end{equation*}
  where $S_P\opt = \argmin_{x \in \fr} \E_{P}[\obj]$.
\end{lemma}



\subsection{Proof of Theorem~\ref{theorem:el-stoch-opt}}
\label{appendix:el-stoch-opt-proof}

By Example~\ref{example:lipschitz}, we have that
$\{\loss(x; \cdot: x\in \mathcal{X} \}$ is $P_0$-Donsker with envelope
function
$M_2(\statval) = |\loss(x_0; \statval)| + M(\statval)
\diam(\xdomain)$. Further, Lemma~\ref{lemma:danskin} implies that the
hypotheses of Theorem~\ref{theorem:el-smooth} are satisfied. Indeed, when the
set of $P$-optima $S_P\opt$ is a singleton, Lemma~\ref{lemma:danskin} gives
that $dT_P$ is a linear functional on the space of bounded measures
$\measures$,
\begin{equation*}
  dT_{P_0}(H) = \int \loss(x\opt; \statval) dH(\statval)
\end{equation*}
where $x\opt = \argmin_{x \in \xdomain} \E_{P_0}[\loss(x; \statrv)]$ and the
canonical gradient of $T_{\rm opt}$ is given by
$T^{(1)}(\statval; P_0) = \loss(x\opt; \statval) - \E_{P_0}[\loss(x\opt;
\statrv)]$.

\subsection{Proof of Lemma~\ref{lemma:danskin}}
\label{section:proof-danskin-lemma}

For notational convenience, we identify the set
$\mathcal{H} \defeq \{\loss(x; \cdot) : x \in \xdomain\}$ as a subset of
$L^2(P_0)$, viewed as functions mapping $\statdomain \to \R$ indexed by
$x$. Let $H \in B(\mathcal{H}, P_0)$, where for convenience we use the notational
shorthand
\begin{equation*}
  H(x) \defeq H(\loss(x; \cdot)) = \int \loss(x; \statval) dH(\statval),
\end{equation*}
where we have identified $H$ with a measure in $\measures$, as in the
discussion following Definition~\ref{definition:hadamard-directional}.
We  have the norm $\norm{H} \defeq \sup_{x \in \xdomain} |H(x)|$,
where $\norm{H} < \infty$ for $H \in B(\mathcal{H}, P_0)$.
In addition, we denote the set of $\epsilon$-minimal points for
problem~\eqref{eqn:pop} with distribution $P$ by
\begin{equation*}
  S_P\opt(\epsilon) \defeq \left\{ x \in \mathcal{X}: \E_P[\obj] \le \inf_{x
      \in \mathcal{X}} \E_P[\obj] + \epsilon \right\},
\end{equation*}
where we let $S_P\opt = S_P\opt(0)$.

We first show that for $H_n \in B(\mathcal{H}, P_0)$ with $\norm{H - H_n} \to 0$,
we have for any sequence $t_n \to 0$ that
\begin{equation}
  \label{eqn:danskin-simple-upper}
  \limsup_n \frac{1}{t_n} \left(T_{\rm opt}(P_0 + t_n H_n) - T_{\rm opt}(P_0)\right)
  \le \inf_{x\opt \in S_{P_0}\opt} H(x\opt).
\end{equation}
Indeed, let $x\opt \in S_{P_0}\opt$. Then
\begin{equation*}
  T_{\rm opt}(P_0 + t_n H_n) - T_{\rm opt}(P_0) \le
  \E_{P_0}[\loss(x\opt; \statrv)]
  + t_n H_n(x\opt) - \E_{P_0}[\loss(x\opt; \statrv)]
  \le t_n H_n(x\opt).
\end{equation*}
By definition, we have $|H_n(x\opt) - H(x\opt)| \le \norm{H_n - H}
\to 0$ as $n \to \infty$, whence
\begin{equation*}
  \limsup_n \frac{1}{t_n} \left(T_{\rm opt}(P_0 + t_n H_n) - T_{\rm opt}(P_0)\right)
  \le \limsup_n \frac{1}{t_n} t_n H_n(x\opt) = H(x\opt).
\end{equation*}
As $x\opt \in S_{P_0}\opt$ is otherwise arbitrary, this yields
expression~\eqref{eqn:danskin-simple-upper}.

We now turn to the corresponding lower bound that
\begin{equation}
  \label{eqn:danskin-simple-lower}
  \liminf_n \frac{1}{t_n} \left(T_{\rm opt}(P_0 + t_n H_n) - T_{\rm opt}(P_0)\right)
  \ge \inf_{x\opt \in S_{P_0}\opt} H(x\opt).
\end{equation}
Because $\norm{H} < \infty$ and $\norm{H_n - H} \to 0$, we see that
\begin{align*}
  T_{\rm opt}(P_0 + t_n H_n) = \inf_{x \in \xdomain}
  \left\{\E_{P_0}[\loss(x; \statrv)] + t_n H_n(x)\right\}
  & \le \inf_{x \in \xdomain}
  \left\{\E_{P_0}[\loss(x; \statrv)] + t_n \norm{H_n - H} + t_n
    \norm{H}\right\} \\
  & \le \inf_{x \in \xdomain} \E_{P_0}[\loss(x; \statrv)]
  + O(1) \cdot t_n.
\end{align*}
Thus, for any $y_n \in S_{P_0 + t_n H_n}\opt$ we have $y_n \in S_{P_0}(c t_n)$
for a constant $c < \infty$. Thus each subsequence of $y_n$ has a further
subsequence converging to some $x\opt \in S_{P_0}\opt$ by the assumed
compactness of $S_{P_0}(\epsilon)$, and the dominated convergence theorem
implies that $\norm{\loss(y_n; \cdot) - \loss(x\opt; \cdot)}_{L^2(P_0)} \to 0$
if $y_n \to x\opt$. In particular, we find that
\begin{equation*}
  \liminf_n \E_{P_0}[\loss(y_n; \statrv)] = \E_{P_0}[\loss(x\opt; \statrv)]
\end{equation*}
for any $x\opt \in S_{P_0}\opt$.
Letting $y_n \in S_{P_0 + t_n H_n}\opt$, then, we have
\begin{align*}
  T_{\rm opt}(P_0 + t_n H_n) - T_{\rm opt}(P_0)
  & \ge \E_{P_0}[\loss(y_n; \statrv)] + t_n H_n(y_n)
    - \E_{P_0}[\loss(y_n; \statrv)]
    = t_n H_n(y_n).
\end{align*}
Moving to a subsequence if necessary along which $y_n \to x\opt$, we have
$H_n(y_n) - H(x\opt) \le \norm{H_n - H} + |H(y_n) - H(x\opt)| \to 0$, where we
have used that $H$ is continuous with respect to $L^2(P_0)$.  Thus
$t_n H(y_n) \ge t_n H(x\opt) - o(t_n)$, which gives the lower
bound~\eqref{eqn:danskin-simple-lower}.

\subsection{Proof of Theorem~\ref{theorem:ucb}}
\label{subsection:ucb-proof}

We prove only the asymptotic result for the upper confidence bound $u_n$, as
the proof of the lower bound is completely parallel.  By
Theorem~\ref{theorem:real-asymptotic-expansion}, we have that
\begin{equation*}
  \sqrt{n} \left( \sup_{P: \fdiv{P}{\emp} \le \frac{\tol}{n}} \E_P[\loss(\cdot; \statval)]
    - \E_{P_0}[\loss(\cdot; \statval)] \right)
  \cd H_+(\cdot) ~~\text{in}~~ \dualbdd,
\end{equation*}
where we recall the definition~\eqref{eqn:upper-lower-GPs} of the Gaussian
processes $H_+$ and $H_-$.  Applying the delta method as in the proof of
Theorem~\ref{theorem:el-stoch-opt} (see
Section~\ref{appendix:el-stoch-opt-proof} and Lemma~\ref{lemma:danskin},
noting that this is essentially equivalent to the continuity of the infimum
operator in the sup-norm topology) we obtain
\begin{equation*}
  \sqrt{n} \left(u_n - \inf_{x\in\mathcal{X}} \E_{P_0}[\obj]\right)
  \cd \inf_{x \in S_{P_0}\opt} H_+(x)
\end{equation*}
where $S_{P_0}\opt = \argmin_{x\in\mathcal{X}} \E_{P_0}[\obj]$.
This is equivalent to the first claim of the theorem, and the result when
$S_{P_0}\opt$ is a singleton is immediate.

\subsection{Proof of Lemma~\ref{lemma:cressie-read-risk}}
\label{sec:proof-cressie-read-risk}

We first show the calculation to derive expression~\eqref{eqn:cressie-read}
of the conjugate $f_k^*$. For $k > 1$, we when $t \ge 0$ we have
\begin{equation*}
  \frac{\partial}{\partial t} \left[st - f_k(t)\right] = s -
  \frac{1}{2(k-1)} (t^{k-1}-1).
\end{equation*}
If $s < 0$, then the supremum is attained at $t = 0$, as the
derivative above is $< 0$ at $t = 0$. If $s \ge -\frac{1}{2(k-1)}$,
then we solve $\frac{\partial}{\partial t} \left[st - f_k(t)\right]
= 0$ to find $t = ((k-1)s / 2 + 1)^{1/(k-1)}$, and substituting gives
\begin{equation*}
  st - f(t) = \frac{2}{k}\left(\frac{k-1}{2}s + 1\right)^{\frac{k}{k-1}} - \frac{2}{k}
\end{equation*}
which is our desired result
as $1 - 1/k = 1/k_*$.
When $k < 1$, a completely similar proof gives the result.

We now turn to computing the supremum in the lemma. For shorthand,
let $Z = \loss(x; \statrv)$. By the duality
result of Lemma~\ref{lemma:dual-problem}, for any $P_0$ and $\tol \ge 0$
we have
\begin{align*}
  \sup_{P \in \fdivs{P}{P_0} \le \tol} \E_P[Z]
  & = \inf_{\lambda \ge 0, \eta}
    \left\{ \lambda\E_{P_0} \left[f_k^*\left(\frac{Z - \eta}{\lambda}\right)\right]
    + \lambda \tol + \eta \right\} \\
  & = \inf_{\lambda \ge 0, \eta}
    \left\{2^{1 - k_*} \frac{(k-1)^{k_*}}{k} \lambda^{1-k_*}
    \E_{P_0} \left[\hinge{Z - \eta + \frac{2\lambda}{k-1}}^{k_*}\right]
    + \lambda \left(\tol - \frac{2}{k}\right) + \eta \right\} \\
  & = \inf_{\lambda \ge 0, \tilde{\eta}}
    \left\{
    \frac{2^{1 - k_*} (k-1)^{k_*}}{k} \lambda^{1-k_*} 
    \E_{P_0}\left[\hinge{Z - \wt{\eta}}^{k_*}\right]
    + \lambda \left(\tol + \frac{2}{k(k-1)}\right)
    + \wt{\eta} \right\}
\end{align*}
where in the final equality we set
$\wt{\eta} = \eta - \frac{2 \lambda}{k - 1}$, because $\eta$ is
unconstrained.
Taking derivatives with respect to $\lambda$ to infimize the preceding
expression, we have (noting that $\frac{k_* - 1}{k_*} = \frac{1}{k}$)
\begin{align*}
  & 2 \left(\frac{k-1}{2 \lambda} \right)^{k_*} \frac{1 - k_*}{k}
    \E_{P_0}\left[\hinge{Z - \wt{\eta}}^{k_*}\right]
    + \left(\tol + \frac{2}{k(k-1)}\right)
    = 0 \\
  & ~~ \mbox{or} ~~
    \lambda = 2^{\frac{1}{k}}
    (k - 1) (2 + \rho k (k - 1))^{-\frac{1}{k_*}}
    \E_{P_0}\left[\hinge{Z - \wt{\eta}}^{k_*}\right]^\frac{1}{k_*}.
\end{align*}
Substituting $\lambda$ into the preceding display and mapping
$\tol \mapsto \tol/n$ gives the claim of the lemma.


%% file: ergodic.tex
\section{Proofs for Dependent Sequences}
\label{appendix:ergodic-expansion-proof}

In this section, we present proofs of our results on dependent sequences
(Example~\ref{example:lyapunov}, Theorem~\ref{theorem:el-stoch-opt-ergodic},
Proposition~\ref{proposition:sectioning}, and
Theorem~\ref{theorem:el-smooth-ergodic}). We begin by giving a proof of claims
in Example~\ref{example:lyapunov} in
Section~\ref{section:example-lyapunov}. Then, for logical consistency, we
first present the proof of the general result
Theorem~\ref{theorem:el-smooth-ergodic} in
Section~\ref{sec:proof-el-smooth-ergodic}, which is an extension of
Theorem~\ref{theorem:el-smooth} to $\beta$-mixing sequences. We apply this
general result for Hadamard differentiable functionals to stochastic
optimization problems
$T_{\rm opt}(P) = \inf_{x\in \mathcal{X}} \E_{P}[\loss(x; \statrv)]$ and prove
Theorem~\ref{theorem:el-stoch-opt-ergodic} in
Section~\ref{sec:proof-el-smooth-ergodic}. Finally, we prove
Proposition~\ref{proposition:sectioning} in
Section~\ref{section:proof-sectioning}, a sectioning result that provides
exact coverages even for dependent sequences.

\subsection{Proof of Example~\ref{example:lyapunov}}
\label{section:example-lyapunov}

First, we note from~\citet[Theorem 15.0.1]{MeynTw09} that
$\{\statval_n\}_{n \ge 0}$ is aperiodic, positive Harris recurrent and
geometrically ergodic. Letting $\pi$ be the stationary distribution of
$\{\statrv_n\}$, it follows that for some $s \in (0, 1)$ and
$R \in (0, \infty)$, we have
\begin{equation}
  \label{eqn:geom-ergodic}
  \sum_{n=1}^\infty
  s^n \norm{P^n(z, \cdot) - \pi(\cdot)}_{w} \le R w(z)
  ~~\mbox{for all}~~z \in \statdomain
\end{equation}
where the distance $\norm{P - Q}_{w}$ betwen two probabilities $P$ and $Q$ is
given by
\begin{equation*}
  \norm{P(\cdot) - Q(\cdot)}_w
  \defeq 
  \sup \left\{ \left| \int_{\statdomain}
    f(y) P(dy)
    - \int_{\statdomain} f(y) Q(dy) \right|:
  f \mbox{ measurable},~|f| \le w 
  \right\}.
\end{equation*}

Now, let $\{ A_i \}_{i \in \mathcal{I}}$ and $\{ B_j \}_{j \in \mathcal{J}}$
be finite partitions of $\statdomain$ such that $A_i, B_j \in \sigalg$ for all
$i \in \mathcal{I}$ and $j \in \mathcal{J}$. By definition, the $\beta$-mixing
coefficient can be written as
\begin{align}
  \beta_n
  & = \half \sup \sum_{i \in \mathcal{I}, j \in \mathcal{J}}
    \left| \P_{\pi} \left(X_0 \in A_i, \statrv_n \in B_j\right)
    - \pi \left(A_i\right) \pi \left(B_j\right) \right|
  \nonumber \\
  & = \half \sup \sum_{i \in \mathcal{I}, j \in \mathcal{J}}
    \left|
    \int_{A_i} 
    \left( \P_z(\statrv_n \in B_j) - \pi(B_j) \right) \pi( dz)
    \right|
    = \half \sup \sum_{i \in \mathcal{I}, j \in \mathcal{J}}
    \left|
    \int_{A_i} 
    \nu_n(z, B_j) \pi( dz)
    \right|
    \label{eqn:def-beta-mixing}
\end{align}
where $\nu_n(z, \cdot)$ is the signed measure on $(\statdomain, \sigalg)$
given by $\nu_n(z, B) \defeq \P_z(\statrv_n \in B) - \pi(B)$. From the
Hahn-Jordan decomposition theorem, there exists a positive and negative set,
$P_{n, z}$ and $N_{n, z}$, for the signed measure $\nu_n(z, \cdot)$ so that we
can write
\begin{equation*}
  \nu_n(z, B) = \nu_n(z, B \cap P_{n, z}) + \nu_n(z, B \cap N_{n, z})
  ~~\mbox{for all}~~B \in \sigalg.
\end{equation*}
Now, note that
\begin{align*}
  \sum_{j \in \mathcal{J}} |\nu_n(z, B_j)|
  & =  \sum_{j \in \mathcal{J}} \nu_n(z, B_j \cap P_{n, z})
    - \sum_{j \in \mathcal{J}} \nu_n(z, B_j \cap N_{n, z}) \\
  & = \nu_n(z,  P_{n, z}) - \nu_n(z,  N_{n, z}) 
  = 2 \nu_n(z, P_{n, z})
\end{align*}
where second equality follows since $\{B_j\}_{j \in \mathcal{J}}$ is a
partition of $\statdomain$ and the last inequality follows from definition of
$\nu_n(z, \cdot)$.

Since $w \ge 1$, we further have that
  $2 \nu_n(z, P_{n, z})
  \le 2 \norm{P^n(x, \cdot) - \pi(\cdot)}_{w}$.
Collecting these bounds, we have from
inequality~\eqref{eqn:geom-ergodic} that
$\sum_{j \in \mathcal{J}} |\nu_n(z, B_j)| \le s^{n} R w(z)$.  From the
representation~\eqref{eqn:def-beta-mixing}, we then obtain
  $\beta_n
  \le s^{-n} R \E_{\pi} w(\statrv_0)$.
Now, from the Lyapunov conditions
\begin{equation*}
  \E_z w(\statrv_1) \le \gamma w(z) + b~~\mbox{for all}~~ z \in \statdomain
\end{equation*}
where we let $ b \defeq \sup_{z' \in C} \E_{z'} w(\statrv_1)$. By taking
expectations over $\statrv_0 \sim \pi$, note that
$\E_{\pi}w(\statrv_0) \le b / (1-\gamma) < \infty$ (see, for
example,~\citet{GlynnZe08}). This yields our final claim $\beta_n = O(s^n)$.

\subsection{Proof of Theorem~\ref{theorem:el-smooth-ergodic}}
\label{sec:proof-el-smooth-ergodic}

Armed with Lemma~\ref{lemma:key-expansion} and its uniform counterpart given
in Theorem~\ref{theorem:real-asymptotic-expansion} (the proof goes through,
\emph{mutatis mutandis}, under the hypotheses of
Theorem~\ref{theorem:el-smooth-ergodic}), we proceed similarly as in the
proof of Theorem~\ref{theorem:el-smooth}. Only now,
Lemma~\ref{lemma:bracketing-for-beta-mixing} implies that
\begin{equation*}
  \sqrt{n} \left(\E_{\emp}[T^{(1)}(\statval; P_0)]
    - \E_{P_{\statval}}[T^{(1)}(\statval; P_0)] \right)
  \cd N\left(0, \Gamma(T^{(1)}, T^{(1)})\right),
\end{equation*}
we have
\begin{equation}
  \label{eqn:used-later}
  \sqrt{n} \left(
  \sup_{P: \fdivs{P}{\emp} \le \tol/n} T(P)
  - T(P_0) \right)
  \cd \sqrt{\tol\var \left( T^{(1)}(\statval;P_0) \right) } 
  + N\left(0, \Gamma(T^{(1)}, T^{(1)})\right)
\end{equation}
and
\begin{equation*}
  \P\left( T(P_0) \le \sup_{P} \{ T(P)
  \mid \fdivs{P}{\emp} \le \tol / n \} \right)
  \to \P\bigg(W
  \ge - \sqrt{\frac{\tol \var_{P_{\statval}}
      \left( T^{(1)}(\statval; P_0) \right)}{\Gamma(T^{(1)}, T^{(1)})}} \bigg),
\end{equation*}
where $W \sim \normal(0, 1)$.
From a symmetric argument for $\inf$, we obtain the desired result.

\subsection{Proof of Theorem~\ref{theorem:el-stoch-opt-ergodic}}
\label{sec:proof-el-stoch-opt-ergodic}

\paragraph{Case 1: $\statval_0 \sim \pi$}

We first show the result for $\statval_0 \sim \pi$. Since
$P \mapsto T_{\rm opt}(P) = \inf_{x \in \mathcal{X}} \E_P[\loss(x; \statrv)]$
is Hadamard differentiable by Lemma~\ref{lemma:danskin}, it suffices to verify
the hypothesis of Theorem~\ref{theorem:el-smooth-ergodic}. Note from the
discussion following Theorem 1 in~\citet{DoukhanMaRi95} that if
$\hclass \subset L^{2r}(\pi)$ satisfies
$\sum_{n \ge 1} n^{\frac{1}{r-1}} \beta_n < \infty$ and
\begin{equation*}
   \int_0^1 \sqrt{\log N_{[\hspace{1pt}]} (\epsilon, \hclass,
    \norm{\cdot}_{L^{2r}(\pi)})}
  \hspace{2pt} d\epsilon < \infty,
\end{equation*}
then the hypotheses of Lemma~\ref{lemma:bracketing-for-beta-mixing}
holds. Since the other assumptions of Theorem~\ref{theorem:el-smooth-ergodic}
hold from Lemma~\ref{lemma:danskin}, we need only show that this bracketing
integral is finite.  Conveniently, that $\loss(\cdot; \statval)$ is
$M(\statval)$-Lipschitz by Assumption~\ref{assumption:lipschitz}
implies~\cite[Chs.~2.7.4 \& 3.2]{VanDerVaartWe96}
\begin{equation*}
  \int_0^1 \sqrt{
  \log N_{[\hspace{1pt}]} (\epsilon, \hclass, \norm{\cdot}_{L^{2r}(\pi)})}
  \le C \norm{M}_{L^{2r}(\pi)} \int_0^1 \sqrt{ d \log \epsilon^{-1} }
  \hspace{2pt} d\epsilon
  < \infty
\end{equation*}
for a compact set $\mathcal{X} \subset
\R^d$.

\paragraph{Case 2:  $\statval_0 \sim \nu$ for general measures $\nu$.}

As $\loss(\cdot; \statval)$ is continuous, we can ignore issues of outer
measure and treat convergence in the space $\mathcal{C}(\mathcal{X})$ of
continuous functions on $\mathcal{X}$.
We will show
\begin{equation}
  \label{eqn:robust-asymptotics-restated}
  \sqrt{n} \bigg(
  \sup_{P: \fdivs{P}{\emp} \le \frac{\tol}{n}} T(P) - T(P_0) \bigg)
  \cd
  W \defeq
  \normal\left(\sqrt{\tol\var_{\pi} \hspace{2pt} T^{(1)}(\statval;P_0)},
  \Gamma(T^{(1)}, T^{(1)})\right)
\end{equation}
under any initial distribution $\statval_0 \sim \nu$. The result
for infima of $T(P)$ over $\{P : \fdivs{P}{\emp} \le \tol/n\}$ is analogous,
so that this implies the theorem.

We abuse notation and let $W_n(\statval_j^{j+n}) = \sqrt{n} (\sup_{P:
  \fdivs{P}{\emp} \le \frac{\tol}{n}} T(P) - T(P_0))$, except that we
replace the empirical $\emp$ with the empirical distribution over
$\statrv_{j}, \ldots, \statrv_{j+n}$.  To show the
limit~\eqref{eqn:robust-asymptotics-restated}, it suffices to show
$W_n(\statrv_{m_n}^{n+m_n}) \cd W$ for appropriate increasing sequences
$m_n$:
\begin{lemma}
  \label{lemma:harris}
  For any initial distribution $\statrv_0 \sim \nu$,
  $W_n(\statrv_{m_n}^{n+m_n}) - W_n(\statrv_0^n) \cas 0$
  whenever $m_n \to \infty$ and $m_n / \sqrt{n} \to 0$.
\end{lemma}
\ifdefined\usemorstyle
\proof{Proof.}
\else
\begin{proof}
  \fi
  By Lemma~\ref{lemma:shrink-to-emp}, there exists $C >0$ depending
  only on the choice of $f$ and $\rho$ such that
  \begin{align*}
    \left| W_n(\statrv_{m_n}^{n+m_n}) - W_n(\statrv_0^n) \right|
    & \le \sqrt{n} \frac{C}{n} \sup_{x \in \mathcal{X}}
    \sum_{i=0}^{m_n} \left| \loss(x; \statrv_i) - \loss(x; \statrv_{n+i}) \right|  \\
    & \le
    \frac{Cm_n}{\sqrt{n}} \frac{1}{m_n}\sum_{i=0}^{m_n}
    \left( \sup_{x \in \mathcal{X}}
    \left| \loss(x; \statrv_i) \right|
    + \sup_{x \in \mathcal{X}}\left| \loss(x; \statrv_{n+i}) \right| \right).
  \end{align*}
  By hypothesis, we have
  $\sup_{x \in \mathcal{X}} |\loss(x; \statval)| \le |\loss(x_0; \statval)| +
  M(\statval) {\rm diam}(\mathcal{X})$ where
  $\E|\loss(x_0; \statval)| + \E[M(\statrv)] < \infty$, and as
  $\{M(\statrv_i)\}_{i=1}^\infty$ are $\beta$-mixing, the law of large numbers
  holds for any initial distribution~\cite[Proposition 17.1.6]{MeynTw09}.
  Then
  $\frac{1}{\sqrt{n}}\sum_{i=0}^{m_n} \sup_{x \in \mathcal{X}} \left| \loss(x;
    \statrv_i) \right| \cas 0$ so long as $m_n / \sqrt{n} \to 0$.
  \ifdefined\usemorstyle
  \endproof
  \else
\end{proof}
\fi

Fix an arbitrary initial distribution $\statrv_0 \sim
\nu$. Letting $m_n = n^{1/4}$, Case 1 and Lemma~\ref{lemma:harris} yield
\begin{equation*}
  W_n(\statrv_{m_n}^{n+m_n}) \cd W
  ~~\mbox{when}~~\statrv_0 \sim \pi.
\end{equation*}
Let $\mathcal{L}_{\nu}(\statrv_{m_n}^{n+m_n})$ denote the law of
$(\statrv_{m_n}, \ldots, \statrv_{n+m_n})$ when
$\statrv_0 \sim \nu$, and let $Q^n(\statval, \cdot)$ be
the distribution of $\statrv_n$ conditional on $\statrv_0 = \statval$
and $\nu \circ Q^m = \int Q^m(\statval, \cdot) d\nu(\cdot)$.
The Markov property then implies
\begin{align*}
  \tvnorm{\mathcal{L}_{\nu}(\statrv_{m_n}^{n+m_n})
  - \mathcal{L}_{\pi}(\statrv_{0}^{n})}
  = \tvnorm{\nu \circ Q^{m_n} - \pi \circ Q^{m_n}}.
\end{align*}
By positive Harris recurrence
and aperiodicity~\cite[Theorem 13.0.1]{MeynTw09},
$\tvnorm{\nu \circ Q^{m_n} - \pi \circ Q^{m_n}} \to 0$ for any
$m_n \to \infty$. We conclude that
$W_n(\statrv_{m_n}^{n+m_n}) \cd W$ for any $\nu$;
Lemma~\ref{lemma:harris} gives the final result.

\subsection{Proof of Proposition~\ref{proposition:sectioning}}
\label{section:proof-sectioning}

We first show the result for $\nu = \pi$. The general result follows by a
similar argument as in the second part of the proof of
Theorem~\ref{theorem:el-stoch-opt-ergodic}, which we omit for conciseness. To
ease notation, define $N^j_b = \sqrt{b}(U^j_b - T(\pi))$. From the proof of
Theorem~\ref{theorem:el-smooth-ergodic}, we have the asymptotic expansion
\begin{align*}
  N^j_b
  = \sqrt{b} \left( \frac{1}{b} \sum_{k=1}^b \loss(x\opt; \statrv_{(j-1)b + k})
  - \E_{\pi}[\loss(x\opt; \statrv)]
  \right)
  + \sqrt{\tol \var_{\pi} \loss(x\opt; \statrv)}
  + \epsilon_{b, j}
\end{align*}
where $\epsilon_{b, j}$ is a remainder term that satisfies
$\epsilon_{b, j} \cp 0$ as $b \to \infty$. From Cramer's
device~\cite{Billingsley86}, we have that $(N^j_b)_{j=1}^m$ jointly converges
in distribution to a normal distribution with marginals given by
\begin{equation*}
  N^j_b \cd \sqrt{\tol\var_{\pi} \hspace{2pt} \loss(x^*; \statrv)} 
  + N\left(0, \sigma^2_{\pi}\right)
\end{equation*}
for all $j = 1, \ldots, m$. If we can show that $N^j_b$ have asymptotic
covariance equal to $0$, then we have
\begin{equation*}
  \frac{\frac{1}{m} \sum_{j=1}^m  N^j_b - \sqrt{\tol \var_{\pi} \loss(x^*; \statrv)}}{
    \sqrt{b} s_m^2(U_b)}
  \cd T_{m-1}
\end{equation*}
by the continuous mapping theorem.  Since
$\var_{\emp} \loss(x_n^*; \statrv) \cp \var_{\pi} \loss(x^*; \statrv)$ from
Corollary~\ref{corollary:uniform-consistency}, this gives our desired
result. We now show that $N^j_b$ have asymptotic covariance equal to $0$.

Since $\beta$-mixing coefficients upper bound their strongly mixing
counterparts, we have from~\citet[Corollary 2.5.5]{EthierKu09}
\begin{equation*}
  \cov_{\pi} (N^1_b, N^j_b)
  \le 2^{2r+1} \beta_b^{1-1/r} \left( \E_{\pi} |N_b^j|^{2r} \right)^{\frac{1}{2r}}
\end{equation*}
for $j \ge 3$ (we deal with $j = 2$ case separately below). The below lemma
controls moments of $N^j_b$.
\begin{lemma}
  \label{lemma:moment-bound}
  Let Assumption~\ref{assumption:lipschitz} hold with
  $\E_{\pi}[M(\statrv)^{2r}] < \infty$. Then, for all $j = 1, \ldots, m$,
  \begin{equation*}
    \E_{\pi} [ |N^j_b|^{2r} ] \le C_{f, \tol, r, \mathcal{X}}
    \left( \E_{\pi} \left[ M(\statrv)^{2r} \right]
        + \E_{\pi}\left[ |\loss(x_0; \statrv)|^{2r}\right]
    \right)
  \end{equation*}
  where
  $\diam (\mathcal{X}) = \sup_{x, x'} \norm{x-x'}$ and
  $C_{f, \tol, r, \mathcal{X}}$ is a constant that only depends on $f$,
  $\tol$, $r$ and $\diam(\mathcal{X})$.
\end{lemma}
\noindent We defer the proof of the lemma to
Section~\ref{section:proof-moment-bound}. We conclude that
$ \cov_{\pi} (N^1_b, N^i_b) \to 0$ for $i \ge 3$.

To show that $\cov_{\pi} (N^1_b, N^2_b) \to 0$, define $N^2_{b, \epsilon_b}$
identically as $N^2_b$, except now we leave $\floor{\epsilon_b b}$ number
of samples in the beginning. Letting $\epsilon_b = 1/\sqrt{b}$, we still
obtain $\cov_{\pi}(N^1_b, N^2_{b, \epsilon_b}) \to 0$ from an identical
argument as above. Since
$\cov_{\pi} (N^1_b, N^2_{b, \epsilon_b}) - \cov_{\pi} (N^1_b, N^2_b) \to 0$ by
dominated convergence theorem, we obtain the result.

\subsubsection{Proof of Lemma~\ref{lemma:moment-bound}}
\label{section:proof-moment-bound}

First, we consider the following decomposition
\begin{align}
  \label{eqn:nb-decomposition}
  N_b^j = \sqrt{b} \left(
  \sup_{P \in \mathcal{P}_{n, \rho, j}} T(P) - T(\what{P}_b^j) \right)
  + \sqrt{b} \left( T(\what{P}_b^j) - T(\pi) \right).
\end{align}
To bound the moments of the first term, note that
\begin{align}
  0 \le
  \sqrt{b} \left(
  \sup_{P \in \mathcal{P}_{n, \rho, j}} T(P) - T(\what{P}_b^j) \right)
  & = \sqrt{b} \left(
      \inf_{x \in \mathcal{X}} \sup_{P \in \mathcal{P}_{n, \rho, j}} \E_{P}[\obj]
      - \inf_{x \in \mathcal{X}} \E_{\what{P}_b^j} [\obj]
      \right) \nonumber \\
    & \le \sqrt{b} \sup_{x \in \mathcal{X}}
      \left(
       \sup_{P \in \mathcal{P}_{n, \rho, j}} \E_{P}[\obj]
      - \E_{\what{P}_b^j} [\obj] \right)
      \label{eqn:nb-first-term}
\end{align}
From Lemma~\ref{lemma:shrink-to-emp}, for some constant $C_f$ depending only
on $f$, we have
\begin{equation*}
  \mathcal{P}_{n, \tol, j} \subseteq
  \left\{
    P \ll \what{P}^j_b:  \dchi{P}{\what{P}^j_b} \le \frac{C_f \tol}{n}
  \right\}
  \eqdef \mathcal{P}_{2, n, \tol, j}.
\end{equation*}
The right hand side of the bound~\eqref{eqn:nb-first-term} is then bounded by
\begin{align*}
  \sqrt{b} \sup_{x \in \mathcal{X}}
      \left(
       \sup_{P \in \mathcal{P}_{2, n, \rho, j}} \E_{P}[\obj]
      - \E_{\what{P}_b^j} [\obj] \right).
\end{align*}
Now, the following lemma bounds the error term in the variance
expansion~\eqref{eqn:simple-expansion}.
\begin{lemma}[{\cite[Theorem 1]{DuchiNa16,NamkoongDu17}}]
  \label{lemma:error-bound}
  Let $f(t) = \half (t - 1)^2$. Then
  \begin{equation*}
    \sup_{P} \left\{ \E_P[Z] :
    \fdivs{P}{\emp}
    \le \frac{\tol}{n} \right\}
    \le 
    \E_{\emp}[Z] + \sqrt{\frac{2 \tol}{n} \var_{\emp}(Z)}.
  \end{equation*}
\end{lemma}
We conclude that
\begin{align*}
  0 \le
  \sqrt{b} \left(
  \sup_{P \in \mathcal{P}_{n, \rho, j}} T(P) - T(\what{P}_b^j) \right)
  & \le \sup_{x \in \mathcal{X}} \sqrt{2 C_f \tol
    \var_{\what{P}_b^j} \left( \obj \right)} \\
  & \le 2 \sqrt{C_f \tol}
  \left( \E_{\what{P}_b^j} |\loss(x_0; \statrv)|
  + \diam(\mathcal{X}) \E_{\what{P}_b^j}[M(\statrv)]
  \right)
\end{align*}
and hence
\begin{equation}
  b^r \E \left|
  \sup_{P \in \mathcal{P}_{n, \rho, j}} T(P) - T(\what{P}_b^j) \right|^{2r}
  \le 2^{4r-1} (C_f \tol)^r
    \left( \E_{\pi} |\loss(x_0; \statrv)|^{2r}
    + \diam(\mathcal{X})^{2r} \E_{\pi}[M(\statrv)]^{2r}
    \right).
  \label{eqn:nb-first-term-bound}
\end{equation}

To bound the second term in the decomposition~\eqref{eqn:nb-decomposition},
note that
\begin{align*}
  \sqrt{b} \left| T(\what{P}_b^j) - T(\pi) \right|
  & = \sqrt{b}
   \left| \inf_{x \in \mathcal{X}} \E_{\what{P}_b^j}[\obj]
    - \inf_{x \in \mathcal{X}} \E_{\pi} [\obj]
    \right| \\
  & \le
    \sqrt{b} \sup_{x \in \mathcal{X}} 
    \left| \E_{\what{P}_b^j}[\obj] - \E_{\pi} [\obj]
    \right| 
\end{align*}
Now, from a standard symmetrization argument~\cite[Section
2.3]{VanDerVaartWe96}, we have
\begin{align*}
  b^r \E \left[ \sup_{x \in \mathcal{X}} 
    \left| \E_{\what{P}_b^j}[\obj] - \E_{\pi} [\obj]
  \right|^{2r} \right]
  \le \E\left[ \sup_{x \in \mathcal{X}} \left|
        \frac{1}{\sqrt{b}} \sum_{i=1}^b \epsilon_i \loss(x; \statrv_i)
      \right|^{2r} \right]
\end{align*}
where $\epsilon_i$'s are i.i.d. random signs so that
$\P(\epsilon_i = +1) = \P(\epsilon_i = -1) = \half$.  The following standard
chaining bound controls the right hand side~\cite{VanDerVaartWe96}.
\begin{lemma}
  \label{lemma:chaining-bound}
  Under the conditions of Proposition~\ref{proposition:sectioning}, for
  $j = 1, \ldots, m,$
  \begin{align*}
    & \E_{\epsilon}\left[ \sup_{x \in \mathcal{X}} \left|
        \frac{1}{\sqrt{b}} \sum_{i=1}^b \epsilon_i \loss(x; \statrv_i)
      \right|^{2r} \right] \\
    &  \le C_r \left( d^r \diam\left(\mathcal{X}\right)
      \norm{M(\statrv)}_{L^2(\what{P}_b^j)}^{2r}
      + \left( \sqrt{d \diam (\mathcal{X})} + 1 \right)^{2r}
      \norm{\loss(x_0; \statrv)}_{L^2(\what{P}_b^j)}^{2r}
    \right)
  \end{align*}
  for a constant $C_r > 0$ depending only on $r \ge 1$.
\end{lemma}
Taking expectations with respect to $\statrv_i$'s in the preceeding display,
we obtain
\begin{align*}
  & \E \left[ \sup_{x \in \mathcal{X}} \left|
    \frac{1}{\sqrt{b}} \sum_{i=1}^b \epsilon_i \loss(x; \statrv_i)
    \right|^{2r} \right] \\
  &  \le C_r \left( d^r \diam\left(\mathcal{X}\right)
    \E[M(\statrv)^{2r}]
    + \left( \sqrt{d \diam (\mathcal{X})} + 1 \right)^{2r}
    \E[|\loss(x_0; \statrv)|^{2r}] \right).
\end{align*}

Using the bound~\eqref{eqn:nb-first-term-bound} to bound the first term in the
decomposition~\eqref{eqn:nb-decomposition}, and using the preceeding display
to bound the second term, we obtain the final result.

%% file: consistency-proof.tex
\section{Proofs of Consistency Results}

In this appendix, we collect the proofs of the major theorems in
Section~\ref{section:consistency} on consistency of minimizers of the robust
objective~\eqref{eqn:upper}.

\subsection{Proof of Theorem~\ref{theorem:uniform-convergence}}
\label{appendix:uniform-convergence-proof}

Let $\mathcal{P}_{n, \tol} \defeq \{ P: \fdivs{P}{\emp} \le \frac{\tol}{n} \}$
be the collection of distributions near $\emp$.  We use
Lemma~\ref{lemma:shrink-to-emp} to prove the theorem.  Let $\epsilon > 0$ be
as in Assumption~\ref{assumption:moment}, and define $p$ and $q$ by
$q = \min\{2, 1+\epsilon\}$, $p = \max\{2, 1 + \frac{1}{\epsilon}\}$. Then
defining the likelihood ratio
$\likerat(\statval) \defeq \frac{dP}{d\emp}(\statval)$ and $\mathcal{L}_{n,\tol}$
the likelihood ratio set corresponding to $\mathcal{P}_{n,\tol}$, we have
\begin{align}
  \lefteqn{\sup_{P \in \mathcal{P}_{n,\tol}}
  \left| \E_{P}[\obj] - \E_{P_0}[\obj] \right|} \nonumber \\
& \le \sup_{\likerat \in \mathcal{L}_{n,\tol}}
  \E_{\emp} \left[\left|\likerat (\statrv) - 1\right| \obj  \right]
  + \left|\E_{\emp}[\obj] - \E_{P_0}[\obj] \right|
  \nonumber \\
& \le \sup_{\likerat \in \mathcal{L}_{n,\tol}}
  \E_{\emp} \left[\left|\likerat (\statrv)-1\right|^p \right]^{1/p}
  \cdot \E_{\emp}[|\obj|^q]^{1/q} 
  + \left|\E_{\emp}[\obj] - \E_{P_0}[\obj] \right|
  \label{eqn:consistency}
\end{align}
where inequality~\eqref{eqn:consistency} is a consequence of H\"older's inequality.
Applying
Lemma~\ref{lemma:shrink-to-emp}, we have that
\begin{equation*}
  \E_{\emp}[|\likerat(\statval) - 1|^p]^{1/p} 
  = n^{-1/p} \norm{np-\onevec}_p
  \le n^{-1/p} \norm{np - \onevec}_2
  \le n^{-1/p}\sqrt{\frac{\tol}{\gamma_f}}
\end{equation*}
where $\gamma_f$ is as in the lemma. Combining this inequality with
Assumption~\ref{assumption:moment}, the first term in the upper
bound~\eqref{eqn:consistency} goes to $0$. Since the second term converges
uniformly to $0$ (in outer probability) by the Glivenko-Cantelli property,
the desired result follows.

\subsection{Proof of Theorem~\ref{theorem:epi-consistency}}
\label{appendix:consistency-proof}

\newcommand{\epsargmin}{\mbox{$\epsilon$-}\!\argmin}
\newcommand{\epsnargmin}{\mbox{$\epsilon_n$-}\!\argmin}

Before giving the proof proper, we provide a few standard definitions
that are useful.
\begin{definition}
  \label{definition:set-convergence}
  Let $\{A_n\}$ be a sequence of sets in $\R^d$. The \emph{limit supremum}
  (or \emph{limit exterior} or \emph{outer limit}) and \emph{limit infimum}
  (\emph{limit interior} or \emph{inner limit}) of the sequence $\{A_n\}$
  are
  \begin{align*}
    \limsup_n A_n & \defeq \left\{x \in \R^d
    \mid \liminf_{n \to \infty} \dist(x, A_n) = 0 \right\} \\
    \liminf_n A_n & \defeq \left\{x \in \R^d
    \mid \limsup_{n \to \infty} \dist(x, A_n) = 0 \right\}.
  \end{align*}
  Moreover, we write $A_n \to A$ if
  $\limsup_n A_n = \liminf_n A_n = A$.
\end{definition}
\noindent
The last definition of convergence of $A_n \to A$ is
\emph{Painlev\'{e}-Kuratowski convergence}. With this definition, we may
define epigraphical convergence of functions.
\begin{definition}
  \label{definition:epi-conv}
  Let $g_n : \R^d \to \R$ be a sequence of functions, and $g : \R^d \to \R$.
  Then \emph{$g_n$ epi-converges to a function $g$} if
  \begin{equation}
    \label{eqn:epi-convergence}
    \epi g_n \to \epi g
  \end{equation}
  in the sense of Painlev\'{e}-Kuratowski convergence,
  where $\epi g = \{(x, r) \in \R^d \times \R : g(x) \le r \}$.
\end{definition}
We use $g_n \cepi g$ to denote the epi-convergence of $g_n$ to
$g$. If $g$ is proper (meaning that $\dom g \neq \varnothing$), the following
lemma characterizes epi-convergence for closed convex functions.
\begin{lemma}[Rockafellar and Wets~\cite{RockafellarWe98},
  Theorem~7.17]
  \label{lemma:epi}
  Let $g_n, g$ be convex, proper, and lower semi-continuous. The
  following are equivalent.
  \begin{enumerate}[(i)]
  \item $g_n \cepi g$
  \item There exists a dense set $A \subset \R^d$ such that
    $g_n(x) \to g(x)$ for all $x \in A$.
  \item For all compact $C \subset \dom g$ not containing a boundary point
    of $\dom g$,
    \begin{equation*}
      \lim_{n \to \infty} \sup_{x \in C} |g_n(x) - g(x)| = 0.
    \end{equation*}
  \end{enumerate}
\end{lemma}
\noindent The last characterization says that epi-convergence is
equivalent to uniform convergence on compacta. Before moving to the proof of
the theorem, we give one more useful result. 
\begin{lemma}[Rockafellar and Wets~\cite{RockafellarWe98}, Theorem 7.31]
  \label{lemma:epi-and-argmins}
  Let $g_n \cepi g$, where $g_n$ and $g$ are extended real-valued functions
  and $\inf_x g(x) \in (-\infty, \infty)$. Then
  $\inf_x g_n(x) \to \inf_x g(x)$ if and only if for all $\epsilon > 0$,
  there exists a compact set $C$ such that
  \begin{equation*}
    \inf_{x \in C} g_n(x) \le \inf_x g_n(x) + \epsilon
    ~~ \mbox{eventually}.
  \end{equation*}
\end{lemma}

We now show that the sample-based robust upper bound converges
to the population risk. For notational convenience, based on a sample
$\statrv_1, \ldots, \statrv_n$ (represented by the empirical distribution
$\emp$), define the functions
\begin{equation*}
  F(x) \defeq \E_{P_0}[\obj]
  ~~ \mbox{and} ~~
  \what{F}_n(x)
  \defeq \sup_{P \ll \emp} \left\{
    \E_{P} [\loss(x; \statrv)] : \fdivs{P}{\emp} \leq \frac{\tol}{n}
  \right\}.
\end{equation*}
These are both closed convex: $F$ by~\cite{Bertsekas73} and $\what{F}_n$ as
it is the supremum of closed convex functions.
We now show condition (ii) of Lemma~\ref{lemma:epi} holds. Indeed, let
$\epsilon > 0$ be such that
$\E_{P_0}[|\loss(x; \statrv)|^{1 + \epsilon}] < \infty$ for all
$x \in \xdomain$, and define $q = \min\{2, 1 + \epsilon\}$ and
$p = \max\{2, 1 + \epsilon^{-1}\}$ to be its conjugate. Then the
bound~\eqref{eqn:consistency} in the proof of
Theorem~\ref{theorem:uniform-convergence}
implies that for any $x \in \xdomain$ we have
\begin{align*}
  |F(x) - \what{F}_n(x)| \le n^{-1/p}\sqrt{\frac{\tol}{\gamma_f}}
  \E_{\emp}[|\loss(x; \statrv)|^q]^\frac{1}{q}
  + \left|\E_{\emp}[\obj] - \E_{P_0}[\obj] \right|.
\end{align*}
The strong law of large numbers and continuous mapping theorem imply that
$\E_{\emp}[|\loss(x; \statrv)|^q]^{1/q} \cas \E_{P_0}[|\loss(x;
\statrv)|^q]^{1/q}$ for each $x$, and thus for each $x \in \xdomain$, we have
$\what{F}_n(x) \cas F(x)$.  Letting $\what{\xdomain}$ denote any dense but
countable subset of $\xdomain$, we then have
\begin{equation*}
  \what{F}_n(x) \to F(x)
  ~~\mbox{for~all}~ x \in \what{\xdomain}
\end{equation*}
except on a set of $P_0$-probability 0. This is condition (ii) of
Lemma~\ref{lemma:epi}, whence we see that
\begin{equation*}
  \what{F}_n \cepi F
  ~~ \mbox{with~probability}~ 1.
\end{equation*}

With these convergence guarantees, we prove the claims of the theorem. Let us
assume that we are on the event that $\what{F}_n \cepi F$, which occurs with
probability 1. For simplicity of notation and with no loss of generality, we
assume that $F(x) = \what{F}_n(x) = \infty$ for $x \not \in \xdomain$. By
Assumption~\ref{assumption:lsc}, the sub-level sets
$\{x \in \xdomain : \E_{P_0}[\loss(x;\statrv)] \le \alpha\}$
are compact, and $S_{P_0}\opt = \argmin_{x \in \xdomain} \E_{P_0}[\obj]$ is
non-empty ($F$ is closed), convex, and compact. Let $C \subset \R^d$ be a
compact set containing $S_{P_0}\opt$ in its interior. We may then apply
Lemma~\ref{lemma:epi}(iii) to see that
\begin{equation*}
  \sup_{x \in C} |\what{F}_n(x) - F(x)| \to 0.
\end{equation*}
Now, we claim that $S_{\emp}\opt \subset \interior C$ eventually. Indeed, because
$F$ is closed and $S_{P_0}\opt \subset \interior C$, we know that on the
compact set $\bd C$, we have $\inf_{x \in \bd C} F(x) > \inf_x F(x)$. The
uniform convergence of $\what{F}_n$ to $F$ on $C$ then implies that eventually
$\inf_{x \in \bd C} \what{F}_n(x) > \inf_{x \in C} \what{F}_n(x)$, and thus
$S_{\emp}\opt \subset \interior C$. This shows that for any sequence
$x_n \in S_{\emp}\opt$, the points $x_n$ are eventually in the interior of any
compact set $C \supset S_{P_0}\opt$ and thus $\sup_{x_n \in S_{\emp}\opt}
\dist(x_n, S_{P_0}\opt) \to 0$.

The argument of the preceding paragraph shows that any compact set $C$
containing $S_{P_0}\opt$ in its interior guarantees that, on the event
$\what{F}_n \cepi F$, we have
$\inf_{x \in C} \what{F}_n(x) \le \inf_x \what{F}_n(x) + \epsilon$ and
$S_{\emp}\opt \subset {\rm int}~ C$ eventually.  Applying
Lemma~\ref{lemma:epi-and-argmins} gives that
$\inf_x \what{F}_n(x) \cpstar \inf_x F(x)$ as desired. To show the second
result, we note that from the continuous mapping theorem~\cite[Theorem
1.3.6]{VanDerVaartWe96} and $\what{F}_n \cpstar F$ uniformly on $C$
\begin{align*}
  \limsup_{n \to \infty}
  \P^*\left(\dinclude(S_{\emp}\opt, S_{P_0}\opt) \ge \epsilon\right)
  & \le \limsup_{n \to \infty}
    \P^*\left( \inf_{x \in S_{P_0}^{\star \epsilon}} \what{F}_n(x)
    > \inf_{x \in \mathcal{X}} \what{F}_n(x)\right) \\
  & = \P^*\left( \inf_{x \in S_{P_0}^{\star \epsilon}} F(x)
    > \inf_{x \in \mathcal{X}} F(x)\right) = 0
\end{align*}
where $A^{\epsilon} = \{ x : \dist(x, A) \le \epsilon \}$ denotes the
$\epsilon$-enlargement of $A$.
